\documentclass{article}

\usepackage[utf8]{inputenc} % allow utf-8 input
\usepackage[T1]{fontenc}    % use 8-bit T1 fonts
\usepackage{hyperref}       % hyperlinks
\usepackage{url}            % simple URL typesetting
\usepackage{booktabs}       % professional-quality tables
\usepackage{amsfonts, amsmath}       % blackboard math symbols
\usepackage{nicefrac}       % compact symbols for 1/2, etc.
\usepackage{lscape}

\usepackage{microtype}
\usepackage{graphicx}

\usepackage{hyperref}

\usepackage[accepted]{icml2019}

\usepackage[dvipsnames]{xcolor}
\definecolor{myAcolor}{rgb}{0.1215,0.366, 0.8058}%{0.0039, 0.0976, 0.4414}
\newcommand{\myAcolorname}{\textbf{blue}}

\newcommand{\myAcolormarker}{$^*$}
\newcommand{\myAcolorfull}{{\color{myAcolor}\myAcolorname\myAcolormarker}\xspace}

\definecolor{myBcolor}{rgb}{1.0, 0.4980392156862745, 0.054901960784313725}
\newcommand{\myBcolorname}{\textbf{orange}}

\newcommand{\myBcolormarker}{$^\dagger$}
\newcommand{\myBcolorfull}{{\color{myBcolor}\myBcolorname\myBcolormarker}\xspace}

\definecolor{myCcolor}{rgb}{0.17254901960784313, 0.6274509803921569, 0.17254901960784313}
\newcommand{\myCcolorname}{\textbf{green}}

\newcommand{\myCcolormarker}{$^\ddagger$}
\newcommand{\myCcolorfull}{{\color{myCcolor}\myCcolorname\myCcolormarker}\xspace}

\usepackage{todonotes}
\usepackage{subfig}
\usepackage{xspace}

\newcommand{\G}{GP\xspace}
\newcommand{\GG}{GP-GP\xspace}
\newcommand{\GGG}{GP-GP-GP\xspace}
\newcommand{\LG}{LV-GP\xspace}
\newcommand{\LGG}{LV-GP-GP\xspace}
\newcommand{\GLG}{GP-LV-GP\xspace}
\newcommand{\LGGG}{LV-GP-GP-GP\xspace}
\newcommand{\GLGG}{GP-LG-GP-GP\xspace}
\newcommand{\GGLG}{GP-GP-LV-GP\xspace}

\newcommand{\Supmat}{Supplementary material\xspace}
\newcommand{\supmat}{supplementary material\xspace}

\newcommand{\ffn}{f}
\newcommand{\ggn}{g}
\newcommand{\hhn}{h}

\newcommand{\loglikelihood}{log-likelihood\xspace}

\newcommand{\KL}{\textsc{kl}}
\def\Figref#1{Figure~\ref{#1}}
\def\secref#1{section~\ref{#1}}

\icmltitlerunning{DGPs with Importance-Weighted Variational Inference}

\begin{document}

\icmlsetsymbol{equal}{*}
\twocolumn[

\icmltitle{Deep Gaussian Processes with Importance-Weighted Variational Inference}
\begin{icmlauthorlist}
\icmlauthor{Hugh Salimbeni}{ic,prowler}
\icmlauthor{Vincent Dutordoir}{prowler}
\icmlauthor{James Hensman}{prowler}
\icmlauthor{Marc Peter Deisenroth}{ic,prowler}
\end{icmlauthorlist}

\icmlaffiliation{ic}{Imperial College London}
\icmlaffiliation{prowler}{PROWLER.io}

\icmlcorrespondingauthor{Hugh Salimbeni}{hrs13@ic.ac.uk}

\icmlkeywords{Gaussian processes, variational inference, deep learning}

\vskip 0.3in
]

\printAffiliationsAndNotice{}

\begin{abstract}
Deep Gaussian processes (DGPs) can model complex marginal densities as well as complex mappings. Non-Gaussian marginals are essential for modelling real-world data, and can be generated from the DGP by incorporating uncorrelated variables to the model. Previous work on DGP models has introduced noise additively and used variational inference with a combination of sparse Gaussian processes and mean-field Gaussians for the approximate posterior. Additive noise attenuates the signal, and the Gaussian form of variational distribution may lead to an inaccurate posterior. We instead incorporate noisy variables as latent covariates, and propose a novel importance-weighted objective, which leverages analytic results and provides a mechanism to trade off computation for improved accuracy. Our results demonstrate that the importance-weighted objective works well in practice and consistently outperforms classical variational inference, especially for deeper models.
\end{abstract}

\section{Introduction}

Gaussian processes are powerful and popular models with widespread use, but the joint Gaussian assumption of the latent function values can be limiting in many situations. This can be for at least two reasons: firstly, not all prior knowledge is possible to express solely in terms of mean and covariance, and secondly Gaussian marginals are not sufficient for many applications. The deep Gaussian process (DGP) \citep{damianou2013deep} can potentially overcome both these limitations.

We consider the very general problem of conditional density estimation. A Gaussian process model with a Gaussian likelihood is a poor model unless the true data marginals are Gaussian. Even if the marginals are Gaussian, a Gaussian process can still be a poor model if the mapping is not well-captured by a known mean and covariance. For example, changes of lengthscale at unknown locations cannot be captured with a fixed covariance function \citep{neil1998regression}. 

\begin{figure}[t]
    \centering
    \includegraphics[width=0.9\linewidth]{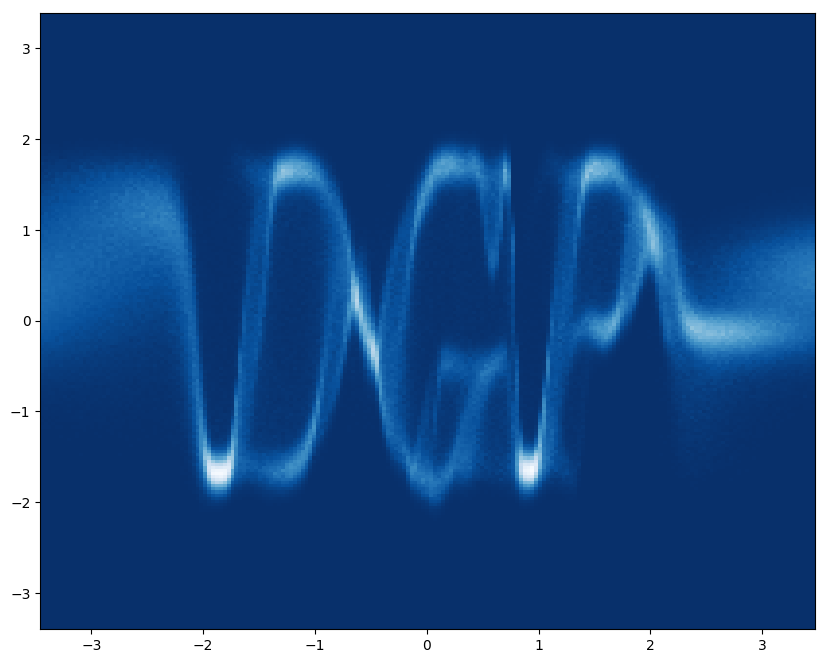}
    % \missingfigure[]{first page fig}
    \caption{Posterior density from a two-layer model, illustrating non-Gaussian marginals and non-smooth input dependence.
    }
    \label{fig:first_page_fig}
\end{figure}

As an example, consider a 1D density estimation task formed from the letters `\textit{DGP}', where the inputs are the horizontal coordinates and the outputs are the vertical coordinates of points uniformly in the glyphs. The marginals are multimodal and piece-wise constant, and also change non-smoothly in some regions of the space (for example changing discontinuously from a bi- to tri-modal density mid-way through the `G'). \Figref{fig:first_page_fig} shows the posterior of a two-layer DGP with this data. % Posteriors from other models (together with the training data) are shown in \figref{fig:posteriors}.  

In this work, we develop the DGP model by revisiting the original construction of \citet{damianou2013deep}, which includes variables that are a priori independent for each data point. Independent variables are important for modelling non-Gaussian marginals, as otherwise function values with the same input will be perfectly correlated. Differently from \citet{damianou2013deep}, we introduce the independent variables to the model as latent covariates (that is, as additional inputs to the GP) \citep{wang2012gaussian} rather than additively as process noise. As we have a clean separation between variables that are correlated and variables that are independent, we can apply appropriate inference for each. For the correlated variables we use a sparse variational GP to represent the posterior \citep{titsias2009, matthews16}. For the uncorrelated variables we use a mean-field approach. We first derive a straightforward combination of the stochastic variational approach of \citet{salimbeni2017doubly} with a mean-field Gaussian approximation for the latent variables. We then propose a novel importance-weighted scheme that improves upon the Gaussian approach and trades additional computation for accuracy while retaining analytic results for the GP parts. 
%
% Our contributions are as follows:
% %
% \begin{itemize}
% \item We derive a new approximate inference scheme using a novel combination of partially collapsed variational inference \citep{salimbeni2017doubly} and importance-weighted sampling \citep{burda2015importance, domke2018importance}.
% \item We develop the DGP model in two ways: we introduce Gaussian noise as a latent covariate rather than process noise, and we use linear projections between layers.
% \end{itemize}

Our results show that the deep Gaussian process with latent variables is an effective model for real data. We investigate a large number of datasets and demonstrate that highly non-Gaussian marginals occur in practice, and that they are not well modelled by the noise-free approach of \citet{salimbeni2017doubly}. We also show that our importance-weighted scheme is always an improvement over variational inference, especially for the deeper models. 

\section{Model}

Our DGP model is built from two components (or \emph{layers}) that can be freely combined to create a family of models. The two types of layers have orthogonal uses: one is for modelling \emph{epistemic} uncertainty (also known as \emph{model} or \emph{reducible} uncertainty, when the output depends deterministically on the input, but there is uncertainty due to lack of observations) and the other is for modelling \emph{aleatoric} uncertainty (also known as \emph{irreducible} uncertainty, when the output is inherently random). Both layers are Gaussian processes, but we use the term \emph{latent variable} when the process has the white noise covariance, and we use \emph{Gaussian process} when the covariance has no noise component. Each layer takes an input and returns an output. The model is constructed by applying the layers sequentially to the input $x_n$ to get a density over an output $y_n$.

\subsection{Gaussian process (\G) layer}
The Gaussian process layer defines a set of continuously indexed random variables, which are a priori jointly Gaussian, with mean and covariance that depend on the indices. We write $f$ for the full set of variables (or `function') and $f(x)$ the particular variable with index $x$. The notation $f\sim\mathcal{GP}(\mu, k)$ states that for any finite set $\{x_i\}$, the variables $\{f(x_i)\}$ are jointly Gaussian, with $\mathbb{E}\left((f(x_i)\right)= \mu(x_i)$ and $\text{cov}\left(f(x_i), f(x_j)\right) = k(x_i, x_j)$, where $\mu$ is a mean function and $k$ is a positive semi-definite covariance function. In this work, we always use `noise free' kernels, which satisfy $k(x, x) = \lim_{x'\to x}k(x, x')$.  

The GP prior is defined for all inputs, not just the ones associated with data. When we notate the GP function in a graphical model (see Figure \ref{fig:graphical_model}), all the variables appear together as $f$. We include an infinity symbol in the graphical model to indicate that the GP node represents an infinite collection of random variables, one for every possible input. %It is important to consider the entire function when the inputs are themselves random variables.

% Note that the prior does not depend \emph{statistically} on the inputs, but the variables are \emph{indexed} by the inputs. This distinction is important when the inputs are themselves random variables, as we discuss in Section \ref{sec:ex_model}. For an input $x$, the output of a Gaussian processes layer is the variable $f(x)$.

\subsection{Latent-variable (LV) layer}
The latent-variable layer introduces variables to the model that are independent for each data point. As we wish to interpret these variables as unobserved covariates we introduce them through concatenation rather than through addition, as in previous work \citep[see, for example ][]{dai2015variational, bui2016deep, damianou2013deep}. We use the notation $[x_n, w_n]$ to denote the concatenation of $x_n$ with $w_n$. Throughout this work each component of $w_n$ will be distributed as $\mathcal N(0, 1)$. For input $x_n$, the output of the latent variable layer is $[x_n, w_n]$. 

\subsection{Example model: \LGG}
\label{sec:ex_model}

\begin{figure}
\centering
\includegraphics[width=0.6\linewidth]{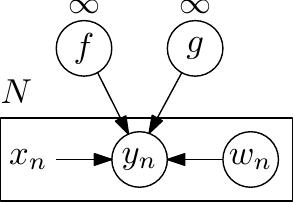}
\caption{Graphical model for \LGG} \label{fig:graphical_model}
\end{figure}

For the purpose of demonstrating the details of inference in the next section, we focus on a particular model. The model has an initial layer of latent variables followed by two Gaussian process layers. We refer to this model as `\LGG'. 
The graphical model indicating the a priori statistical relationship between the variables is shown in Figure \ref{fig:graphical_model}. 
Writing $y$ and $w$ for $\{y_n\}$ and $\{w_n\}$, the likelihood of this model is $p(y|\ffn, \ggn, w) = \prod_n p(y_n|\ffn, \ggn, w_n)$, with
\begin{align}%\label{eq:model_likelihood_nth_term}
p(y_n|\ffn, \ggn, w_n)  = \mathcal N(y_n|f(g([x_n, w_n])), \sigma^2) \,.
\label{eq:likelihood_nth_term}
\end{align}
The priors are
\begin{align*}
w_n \sim  \mathcal N(0, 1)\,,~  \ffn \sim \mathcal{GP}(\mu_1, k_1)\,,  ~ \ggn \sim \mathcal{GP}(\mu_2, k_2)\,. 
\end{align*}
The model is appropriate for continuously valued data, and can model heavy and asymmetric tails, as well as multimodality, as we later demonstrate. We discuss choices for the mean and covariance functions in \secref{sec:mean_cov_functions}. 

% We emphasize that the GP priors are defined over the entire function, not just the $N$ data points. While this distinction can be safely ignored in the single-layer case, it important in the multilayer model to consider the posterior over the whole function. A problem arises if we try to reason about a prior (or posterior) over the variable defined as, for example, $g_n=g([x_n, w_n])$. This variable is coupled to $w_n$ yet the function $g$ itself is independent of $w_n$,. This issue is resolved by considering the prior (and posterior) over the entire function $g$. 

\subsection{Model variants}
The \LGG model can be extended by adding more layers, and by adding more latent variables. For example, we could insert an additional GP layer,
\begin{align*}
p(y|\ffn, \ggn, \hhn, w) = \prod_n \mathcal N\left(y_n|h(f\left(g([x_n, w_n]))\right), \sigma^2\right)\,,
\end{align*}
where $\hhn \sim \mathcal{GP}(\mu_3, k_3)$. We refer to this model as \LGGG. Alternatively we could place the latent variables between the GPs, and have instead $f([g(x_n), w_n])$ for the conditional mean.
% \begin{align*}
% p(y|\ffn, \ggn, w) = \prod_n \mathcal N(y_n|f([g(x_n), w_n]), \sigma^2)\,.
% \end{align*}
%
We refer to this model as \GLG. Other models are analogously defined. For example, \GG is as above but with no latent variable. We refer to models without latent variables as `noise-free'. %They have been used by \citet{cutajar2016random} and \citet{salimbeni2017doubly}.

% \subsection{Model interpretation}
\begin{figure*}
\centering
\subfloat[][\G]{
  \includegraphics[width=0.24\linewidth]{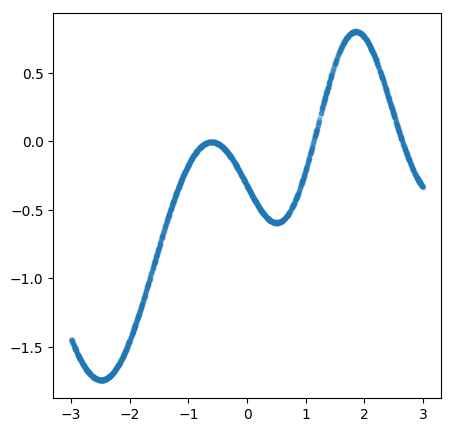}
 }
 \subfloat[][\GG]{
  \includegraphics[width=0.24\linewidth]{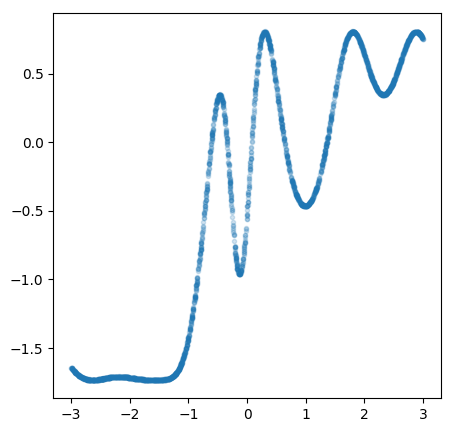}
 }
\subfloat[][\LG]{
  \includegraphics[width=0.24\linewidth]{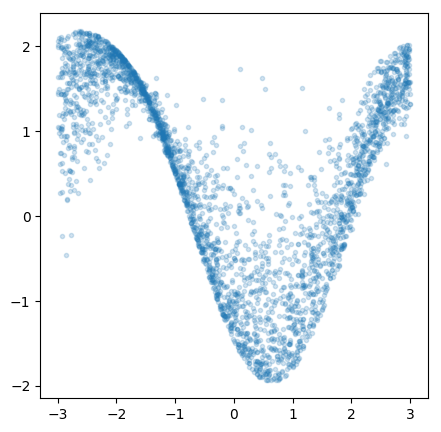}
 }
 \subfloat[][\LGG]{
  \includegraphics[width=0.24\linewidth]{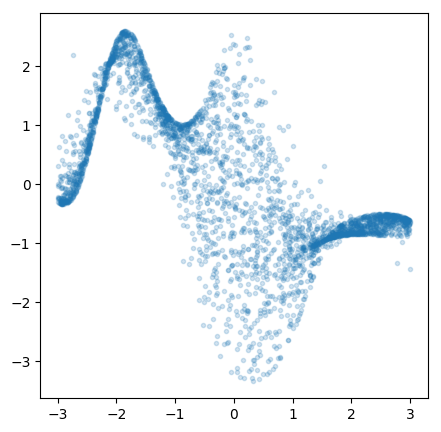}
 }
\caption{A single sample from the prior for 4 illustrative models. The \G and \GG model have no latent variables. Their samples appear as deterministic functions. See the \supmat for more examples.}
\label{fig:prior_samples}
\end{figure*}

Prior samples from models with one and two GP layers are shown in Figure \ref{fig:prior_samples} to illustrate their properties. All the GP layers have 1D outputs and use the RBF kernel with unit lengthscale. Further illustrations with additional samples and additional models are shown in the \supmat, Figures \ref{fig:without_latent_vars}-\ref{fig:three_layer_latent_vars}. For the latent variable models, the number of GPs above the latent variable layer determines the complexity of the marginals, but the number of GPs in total determines the complexity of the functional mapping. %We refer to figures in the \supmat for prior samples from models with latent variables at different depths. 
%For example, the \GGLG (see Figure \ref{fig:three_layer_latent_vars} in the \supmat) and \LG models have similar marginals, but the deep model has a more complicated dependency on input. The \LGGG model (also in Figure \ref{fig:three_layer_latent_vars} in the \supmat), in contrast has a more complicated structure in the marginals as well as the mapping, with multimodal marginals apparent in some regions of the space. 

% \textbf{Interpretation of models}

% Models can be flexibly constructed to suit data. Additional layers increase the complexity of the relationship between input and output, and latent variables deeper in the hierarchy lead to more complex marginals. For example, if the data has Gaussian marginals but a complex dependency on input then the \GG might be a good model. If the data has a simple dependency on input but a non-Gaussian conditional density then \LG is likely to be suitable. If the data has a non-Gaussian conditional density and also has a complex dependency on input then \GLG would be appropriate, or \LGG if the conditional density is highly non-Gaussian.

\subsection{Multiple outputs}\label{sec:multioutput}
All variables in our model (including the model inputs and outputs) may be vector valued. The `multiple output GP' is a GP with a covariance function defined  between outputs $k_{dd'}(x, x')$, where $d$ indexes output dimension and $x$ indexes the input. In this notation, the independent output approach of \citet{damianou2013deep} can be written as $k_{dd'}(x, x')=\delta_{dd'}k(x, x')$. For all our models, we consider a linear model of covariance between outputs: $k_{dd'}(x, x')= \sum_e P_{de}k_e(x, x')P_{ed'}$ \citep{alvarez2012kernels}. For $k_e(x, x')=k(x, x')$ this is equivalent to assuming a Kronecker structured covariance between the outputs stacked into a single vector (i.e. with covariance $(P^{\top}P)\otimes K$), and is also equivalent to multiplying independent GPs stacked into a vector by the matrix $P$. %Note that we are free to take $e<d$. %When there are multiple observed outputs we assume the likelihood factorizes over the outputs, and in this work we assume independent GPs in the final layer.% A priori correlations between outputs are straightforward to include using the linear model, following \citet{dai2017efficient} and \citet{dutordoir2018}.

% \textbf{Mean functions}
\subsection{Mean and covariance functions}
\label{sec:mean_cov_functions}
The DGP model suffers from a pathology if used with zero mean GPs at each layer \citep{duvenaud2014avoiding, dunlop2017deep}. To remedy this problem, \citet{duvenaud2014avoiding} propose concatenating the inputs with the outputs at each GP layer, whereas \citet{salimbeni2017doubly} use a linear mean function to address this issue. We follow the latter approach. %Combining a mean function with the linear multi-output covariance described in \ref{sec:multioutput} is straightforward: it is equivalent to augmenting the output of independent GPs with their inputs (that is, the input propagation approach of \citet{duvenaud2014avoiding}), and then multiply the augmented output by a matrix $V$, where $V=[P, I]$ where.... %We follow this approach, and further optimize $P$, allowing % This is computationally advantageous as we can reduce the number of GPs in the model without lowering the dimensionality of inner layers.
%
% a multi-output kernel via linearly projecting independent GPs, but additionally concatenate the inputs to the GP values before the linear projection. This is equivalent to extending the GP with additional deterministic outputs (a mean function with $m_d(x)=x$ zero otherwise, and covariance $k_d(x, x')=0$). With this choice we can initialize the projection matrix as $P=[V, I]$ where $V$ is a matrix of principle components of the input data, and $I$ is the identity matrix. In this way we achieve the linear mean function of \citet{salimbeni2017doubly}, but if we optimize $P$ we can achieve arbitrary linear transformations of the inputs and outputs. %This has the additional advantage of allowing us to set the number of independent GPs (or equivalently, the rank of the covariance) in advance. 
%
The inference we present in the next section is agnostic to covariance function, so we are free to make any choice. In our experiments, we use an RBF kernel for each layer, sharing the kernel over the different outputs.% (that is, $k_e(x, x')=k(x, x')$). % the same RBF kernel for each output.% $k_d(x, x')=k(x, x')=\sigma_\ell^2\exp\left(-(x-x')^2/l_{\ell}\right)$ for the $\ell$th layer, where $l$ is a lengthscale vector of the same dimension as $x$.

\section{Inference in the \LGG model}
In this section, we present two approaches for approximate inference: the first with variational inference and a second with importance-weighted variational inference. Both schemes can optionally amortize the optimization of the local variational parameters (often referred to as `auto-encoding'), are scalable through data sub-sampling, and can exploit natural gradients of the variational parameters for the final layer. While the variational approach is a straightforward extension of the doubly-stochastic method by \citet{salimbeni2017doubly}, the importance-weighted approach is more subtle and requires a careful choice of variational distribution to retain the analytic results for the final layer.

\subsection{Variational inference}
\label{sec:VI}
Variational inference seeks an approximate posterior that is close to the true posterior in terms of $\KL$ divergence. The posterior is typically restricted to some tractable family of distributions, and an optimization problem is formed by  minimizing the $\KL$ divergence from an approximate posterior to the true posterior. Equivalently, the same objective can be obtained by applying Jensen's inequality to an importance-weighted expression for the marginal likelihood \citep{domke2018importance}. For an approximate posterior, we follow \citet{damianou2013deep} and use a mean-field Gaussian distribution for the latent variables $q(w)=\prod_n q(w_n)$ with $q(w_n) = \mathcal N(a_n, b_n)$, and independent processes for the functions. The posterior density\footnote{We abuse notation and write a process as if it has a density. The derivation can be made rigorous as the densities only appear as expectations of ratios. See \citet{matthews16} for details.} then has the same structure as the prior: $q(w, f, g)=q(w)q(f)q(g)$.
%
% : $, $q(\ffn) \sim \mathcal{GP}(m_1, \Sigma_1)$ and $q(\ggn) \sim \mathcal{GP}(m_2, \Sigma_2)$. %To see the connection to importance sampling
We begin by writing the (exact) marginal likelihood as
\begin{align}\label{eq:exact_loglikelihood}
    p(y) = \mathop{\mathbb{E}}_{\ffn, \ggn, w} \left[ p(y|\ffn, \ggn, w) \frac{p(\ffn)p(\ggn)p(w)}{q(\ffn)q(\ggn)q(w)}\right]\,,
\end{align}
where the expectations are taken with respect to the variational distributions.   %The expectation \eqref{eq:exact_loglikelihood} always has the same value regardless of $q(f, g, w)$.
Applying Jensen's inequality to the logarithm of both sides of \eqref{eq:exact_loglikelihood}, we obtain
\begin{align} \label{eq:ordinary_VI_bound}
    \log p(y) \ge \sum_n\left(A_n- \textsc{kl}_{w_n}\right) - \textsc{kl}_f - \textsc{kl}_g\,,
\end{align}
where we have used the short-hand $\textsc{kl}_h$ for $\textsc{kl}(q(h)||p(h))$%=:-\mathbb E_{h}\log p(h)/q(h)$
, and $A_n$ is given by
\begin{align}
    A_n = \mathbb{E}_{\ffn, \ggn, w_n}\log p(y_n|f, g, w_n) \,.
    % A_n = \mathop{\mathbb{E}}_{\ffn, \ggn, w_n}\log p(y|f(g([x_n, w_n]))) .
\end{align}
By considering a variational distribution over the entire function for $f$ and $g$ we avoid the difficulty of representing the indeterminately indexed inner-layer variables. We require only that $\KL_f$ and $\KL_g$ are finite, which is possible if we construct $q(f)$ and $q(g)$ as measures with respect to their priors. We follow \citet{hensman2013} and form these posteriors by conditioning the prior on some finite set of `inducing' points with a parameterized Gaussian distribution. We defer the details to the \supmat (Section \ref{sec:var_posterior_derivation}) as the results are widely known \citep[see, for example][]{matthews16, cheng2016incremental, dutordoir2018}. The priors and variational posteriors over $f$ and $g$ are independent (the coupling is only in the likelihood), so the results from the single layer GP apply to each layer.

To evaluate $A_n$ exactly is intractable except in the single-layer case (and then only with a certain kernel; see \citet{titsias2010bayesian} for details), so we rely on a Monte Carlo estimate. To obtain unbiased samples we first reparameterize the expectations over $w_n$ and $g$, where
$w_n = a_n + \epsilon_1 \sqrt{b_n}$ 
and 
$g([x_n, w_n]) = \mu_2([x_n, w_n]) + \epsilon_2 \sqrt{k_2([x_n, w_n], [x_n, w_n])}$
where $\epsilon_1, \epsilon_2\sim \mathcal N(0, 1)$.
We then obtain an estimate of $A_n$ by sampling from $\epsilon_1, \epsilon_2$, which is the `reparameterization trick' popularized by \citet{rezende2014stochastic, kingma2013auto}. After these two expectations have been approximated with a Monte Carlo estimate we can take the expectation over $f$ analytically as the likelihood is Gaussian.

\subsection{Importance-weighted variational inference}

Jensen's inequality is tighter if the quantity inside the expectation is concentrated about its mean \citep{domke2018importance}. To decrease the variance inside \eqref{eq:exact_loglikelihood} while preserving the value, we can replace the $w$ term with a sample average of $K$ terms,
\newcommand{\Efgw}{\!\!\!\mathop{\mathbb{E}}_{\ffn, \ggn, w}\!}
\begin{align}
    p(y) = \Efgw \frac{1}{K}\sum_{k=1}^K p(y|\ffn, \ggn, w^{(k)})\frac{p(w^{(k)})}{q(w^{(k)})} \frac{p(\ffn)p(\ggn)}{q(\ffn)q(\ggn)}\,.
\label{eq:exact_likelihood_K_sample}
\end{align}

The expression inside the expectation in \eqref{eq:exact_likelihood_K_sample} is known as the importance sampled estimate (with respect to $w$), motivating the term `importance-weighted variational inference' when we take the logarithm of both sides and form a lower bound using Jensen's inequality. %Note that while in the previous section the slack in Jensen's inequality is exactly the $\KL$ divergence between the approximate posterior and the true posterior, this is not true here. We are not generally interested in the posterior over $w$, however, as it tells us nothing about test points. 
%
% the variance of the term inside the expectation will be zero when the denominator equals the numerator (that is, the unnormalized posterior). Taking Monte Carlo samples for the expectation \eqref{eq:exact_loglikelihood} is known as importance sampling. Instead of taking samples, 
%
%
% This approach is equivalent to minimizing a $\KL$ divergence between an approximating distribution and the true posterior, but now the approximate posterior is defined on an augmented space. For prediction we are not interested in the posterior over $w$ as it is always the prior at test points, but we are interested in the posterior over the Gaussian process parts. Therefore the combination of IWVI for $w$ with VI for $f$ and $g$ is a 
%
Applying Jensen's inequality to \eqref{eq:exact_likelihood_K_sample}, we have the lower bound
\begin{align}
    \log p(y) \ge \sum_{n=1}^N B_n - \textsc{kl}_f - \textsc{kl}_g \,,
\label{eq:pure_iwae}
\end{align}
where the data-fit term $B_n$ is given by
\newcommand{\Efgwn}{\!\!\!\!\mathop{\mathbb{E}}_{\ffn, \ggn, w_n}\!\!\!\!}
\begin{align}
%  \!B_n = \Efgwn \log \frac{1}{K}\sum_k p(y_n|f(g([x_n, w^{(k)}_n]))) \frac{p(w_n^{(k)})}{q(w_n^{(k)})}\,.
  B_n = \Efgwn \log \frac{1}{K}\sum_k p(y_n| f, g, w^{(k)}_n)\frac{p(w_n^{(k)})}{q(w_n^{(k)})}\,.
\label{eq:pure_iwae_nth_term}
\end{align}

This approach provides a strictly tighter bound \citep{burda2015importance, domke2018importance}, but the expression for $B_n$ is less convenient to estimate than $A_n$ as we cannot apply the analytic result for the $f$ expectation due to the non-linearity of the logarithm. We must therefore resort to sampling for the expectation over $f$ as well as $g$ and $w_n$, incurring potentially high variance. This seems like an unacceptable price to pay as we have not gained any additional flexibility over $f$. In the next section, we show how we can retain the analytic expectation for $f$ while exploiting importance-weighted sampling for $w$.  

% The approach we have just described is a hybrid of importance-weighted Variational Inference \citep{burda2015importance, domke2018importance} with standard variational inference. It provides a tighter bound than variational inference, at the expense of extra variance due to the sampling of $\tfrac{p(w_n^{(k)})}{q(w_n^{(k)})}$. We can compute an implicit approximate posterior over $w$ \citep{domke2018importance}, but we are not primarily interested in the posterior over $w$, but rather the posterior over $f$ and $g$. Note that we cannot use importance sampling for $f$ or $g$ as Gaussian processes do not admit a density function \footnote{We have notated a Gaussian process density, but we only evaluate it inside a $\textsc{kl}$ divergence, which is well defined. For a rigorous treatement see \citep{matthews16}}. 

\subsection{Analytic final layer}
\label{sec:IW_inference}
The expression in \eqref{eq:pure_iwae_nth_term} does not exploit the conjugacy of the Gaussian likelihood with the final layer.
In this section, we present a novel two-stage approach to obtain a bound that has all the analytic properties of the variational bound \eqref{eq:ordinary_VI_bound}, but with improved accuracy. 
Our aim is to obtain a modified version of~\eqref{eq:pure_iwae_nth_term} but with the $f$ expectation taken over the logarithm of the likelihood, since this expectation is tractable. We will show that 
% by carefully choosing the approximations,
we can find a bound that replaces the term  $p(y_n|f, g, w_n)$ in \eqref{eq:pure_iwae_nth_term} with $\exp\mathbb{E}_f \log p(y_n|f, g, w_n)$, which we can compute analytically. 
It is worth noting that since the likelihood is Gaussian we could analytically integrate out $\ffn$ to obtain $p(y|\ggn, w)$, though doing so precludes the use of minibatches and incurs $\mathcal O (N^3)$ complexity. It is not surprising, then, that it is possible to `collapse' the bound over $f$ approximately. %The bound we derive is always tighter than that in \eqref{eq:pure_iwae} as it is optimal conditioned on the assumptions for $g$ and $w$, whereas \eqref{eq:pure_iwae} makes an additional independence assumption for $f$ \todo{discuss this in appendix?}. 
Our approach is inspired by the partially collapsed approaches in \citet{hensman2012fast}.
We begin by applying Jensen's inequality to the $f$ expectation in \eqref{eq:exact_loglikelihood}:
\begin{align*}
    \log p(y|\ggn, w) 
    \ge \sum_n L_n(g, w_n) - \textsc{kl}_f,
\end{align*}
where the data term is given by
\begin{align}\label{eq:L_g_w}
L_n(g, w_n) = \mathbb{E}_{\ffn} \log p(y_n|f, g, w_n)\,.
\end{align}
The expression for $L_n(g, w_n)$ is available in closed form as the conditional likelihood is Gaussian  \citep[see, for example, ][]{titsias2009} . 
Applying the exponential function to both sides of \eqref{eq:L_g_w} gives the bound
\begin{align}
    p(y|\ggn, w)\geq \exp \left[ \sum_n L_n(g, w_n) - \textsc{kl}_f \right] \,.
    \label{eq:conditional_marginal_bound}
    % p(y|\ggn, w)\geq e^{\sum_n L_n(g, w_n) - \textsc{kl}_f}
\end{align}
Returning again to \eqref{eq:exact_loglikelihood}, the exact marginal likelihood can be expressed equivalently with $f$ marginalized as
\begin{align}
    p(y) = \mathop{\mathbb{E}}_{\ggn, w} p(y|\ggn,w)
    \frac{p(w)p(\ggn)}{q(w)q(\ggn)}\,.
    \label{eq:marginal_no_g}
\end{align}
We can now use the bound on $p(y|\ggn,w)$ from \eqref{eq:conditional_marginal_bound} and substitute into \eqref{eq:marginal_no_g} to obtain
\begin{align*}
    p(y) \ge \mathop{\mathbb{E}}_{g, w}\exp\left[\sum_n L_n(g, w_n) - \textsc{kl}_f
    \right]\frac{p(w)p(\ggn)}{q(w)q(\ggn)}\,.
\end{align*}

Next we use Jensen's inequality for the $g$ expectation. After some rearranging the bound is given by
\begin{align*}
    \log p(y) 
    \geq \sum_n \mathbb{E}_{\ggn} \underbrace{\log \mathbb{E}_w \frac{e^{L_n(g, w_n)}p(w_n)}{q(w_n)}}_{T_n(\ggn)}
    - \textsc{kl}_f - \textsc{kl}_g  \,.
\end{align*}
To bound $T_n(\ggn)$, we first reduce the variance of the quantity inside the expectation using the sample average as before to tighten the subsequent use of Jensen's inequality. For any $K$, $T_n(\ggn)$ is (exactly) equal to
\begin{align*}
    {T_n(\ggn)} = \log \mathbb{E}_{w_n} \frac{1}{K}\sum_k \frac{e^{L_n(\ggn, w^{(k)}_n)}p(w_n^{(k)})}{q(w^{(k)}_n)}\,,
\end{align*}
where $w^{(k)}_n$ are independent samples from $q(w_n)$. We now make a final use of Jensen for the $w_n$ expectation,
% \begin{align*}
%     {T_n(\ggn)} \ge \mathbb{E}_{w_n} \log \frac{1}{K}\sum_k \frac{e^{L_n(\ggn, w^{(k)}_n)}p(w^{(k)}_n)}{q(w^{(k)}_n)}\,.
% \end{align*}
and the final objective is then given by
% \begin{figure*}
% \centering
% \subfloat[][solar GP]{
%   \includegraphics[width=0.25\linewidth]{figs/plots/solar_GP.png}
%  }
%  \subfloat[][solar GP-GP]{
%   \includegraphics[width=0.25\linewidth]{figs/plots/solar_GP_GP.png}
%  }
% \subfloat[][solar LV-GP]{
%   \includegraphics[width=0.25\linewidth]{figs/plots/solar_LV_GP.png}
%  }
%  \subfloat[][solar LV-GP-GP]{
%   \includegraphics[width=0.25\linewidth]{figs/plots/solar_LV_GP_GP.png}
%  }\\
%  \subfloat[][bike GP]{
%   \includegraphics[width=0.25\linewidth]{figs/plots/bike_GP.png}
%  }
%  \subfloat[][bike GP-GP]{
%   \includegraphics[width=0.25\linewidth]{figs/plots/bike_GP_GP.png}
%  }
% \subfloat[][bike LV-GP]{
%   \includegraphics[width=0.25\linewidth]{figs/plots/bike_LV_GP.png}
%  }
%  \subfloat[][bike LV-GP-GP]{
%   \includegraphics[width=0.25\linewidth]{figs/plots/bike_LV_GP_GP.png}
%  }
% \caption{%Samples from the predictive distribution at test points (orange) together with the actual value (blue) in the PCA projection to 2D.
% }
% \label{fig:posteriors}
% \end{figure*}
\newcommand{\Egwn}{\!\mathop{\mathbb{E}}_{\ggn, w_n}\!}
\begin{align}
    % \mathcal L = 
    \sum_n \Egwn\log \frac{1}{K}\sum_k \frac{e^{L_n(\ggn, w^{(k)}_n)}p(w^{(k)}_n)}{q(w^{(k)}_n)}
    - \textsc{kl}_f - \textsc{kl}_g  \,.
    \label{eq:iwae_final_objective}
\end{align}
%
% This bound has a strong similarity to the more straightforward variational inference bound ~\eqref{eq:ordinary_VI_bound}, but the
This bound can be evaluated using Monte Carlo sampling for $g$ and $w_n$, both with the reparameterization described in Section \ref{sec:VI}. %We refer to \eqref{eq:iwae_final_objective} as the importance-weighted evidence lower bound (IW-ELBO) and \eqref{eq:ordinary_VI_bound} as the variational inference evidence lower bound (VI-ELBO).

The $K$ terms inside the sum in \eqref{eq:iwae_final_objective} must be sampled with a single draw from $q(g)$, and not independently. This is not an insurmountable problem, however, as we can draw $K$ samples using the full covariance and reparameterize using the Cholesky factor at $\mathcal O( K^3 )$ cost. Note that the decomposition over the $N$ data points is a consequence of our choice of proposal distributions and the factorization of the likelihood, so we do not incur $\mathcal{O}(N^3)$ complexity and can sample each term in the sum over $N$ independently.

\subsection{The posterior over the latent variables}
Variational inference simultaneously finds a lower bound and an approximate posterior. The importance weighted approach does also minimize the $\textsc{kl}$ divergence posterior, but the posterior is an implicit one over an augmented space. Samples from this posterior are obtained by first sampling $w_n^k \sim q(w_n)$ and then selecting one of the $w_n^k$ with probabilities proportional to $\tfrac{e^{L_n(\ggn, w^{(k)}_n)}p(w^{(k)}_n)}{q(w^{(k)}_n)}$. We refer to \citet{domke2018importance} for details. In this work, we are not concerned with the posterior over the latent variables themselves, but rather with prediction at test points where we sample from the prior. 

\subsection{Further inference details}
We can amortize the optimization of the $a_n, b_n$ parameters by making them parameterized functions of $(x_n, y_n)$. As the bound is a sum over the data we can use data subsampling. We can also use natural gradients for the variational parameters of the final layer, following \citet{salimbeni2018natural, dutordoir2018}. Our bound is modular in both the GP and LV layers so that it extends straightforwardly to the other model variants. %In the noise-free case, both approaches reduce to the bound of \citet{salimbeni2017doubly}. 

\section{Results}

 \begin{figure*}
	\centering
	\subfloat[][\G ]{
		\includegraphics[width=0.24\linewidth]{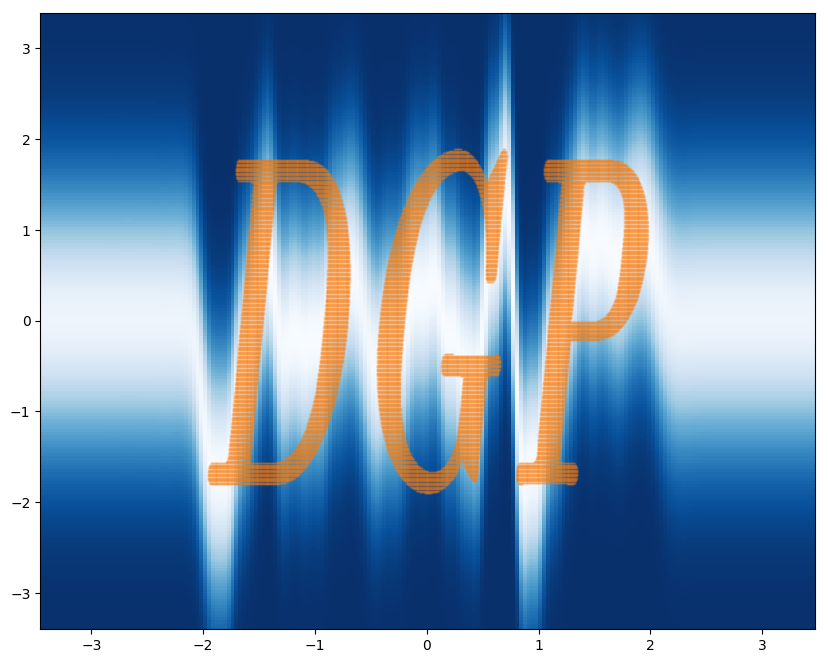}
	}
	\subfloat[][\GG ]{
		\includegraphics[width=0.24\linewidth]{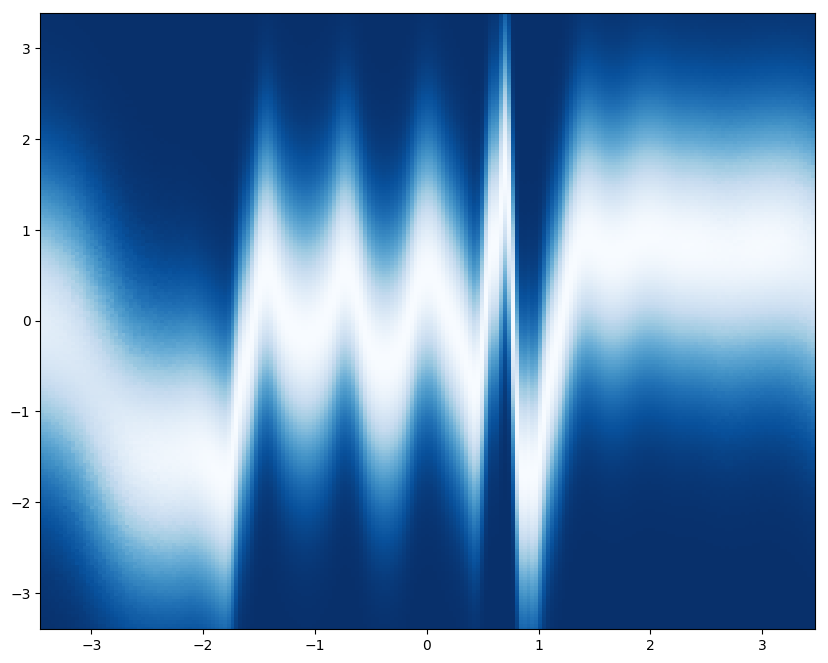}
	}
	\subfloat[][\LG ]{
		\includegraphics[width=0.24\linewidth]{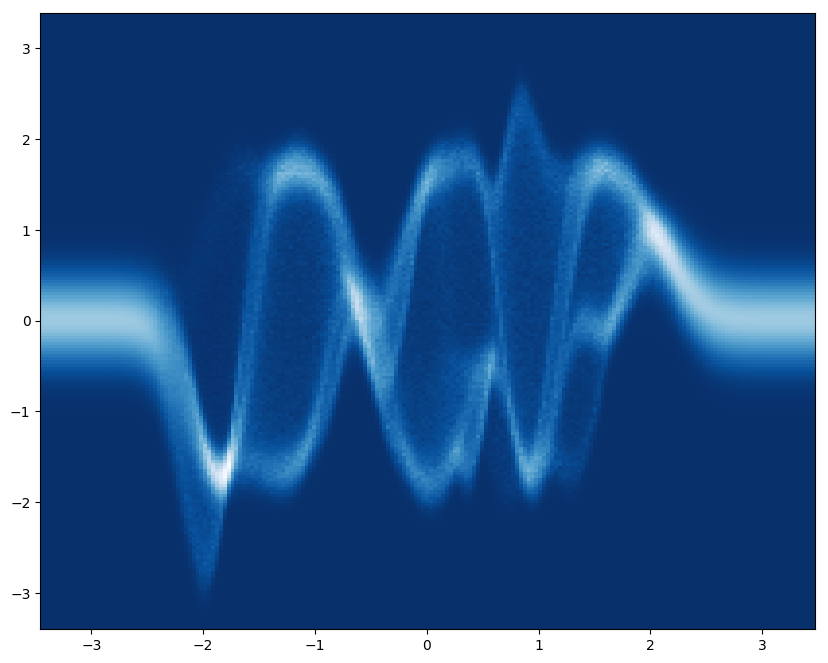}
	}
	\subfloat[][\LGG ]{
		\includegraphics[width=0.24\linewidth]{figs/posteriors/posterior_italics_H20_DGP_L1_G1_IWAE.png}
	}
	\caption{Posteriors for 1D synthetic data. The models without latent variables cannot capture the bimodality or heteroscedasticity. See Figure \ref{fig:first_page_fig} for the \LGGG model.}
	\label{fig:1D_posteriors}
\end{figure*}

We use a density estimation task to establish the utility of the DGP models with and without latent variables. For the latent variable models we also compare our importance-weighted (IW) inference \eqref{eq:iwae_final_objective} against the variational inference (VI) approach \eqref{eq:ordinary_VI_bound}. Our central interest is how well the models balance the complexity of the marginals with the complexity of the mapping from inputs to outputs. 

\subsection{1D example}

To illustrate the inductive bias of our models we show a 1D example with a conditional density that is non-Gaussian, heteroscedastic and changes discontinuously in some regions of the space. We form a dataset from the letters `\textit{DGP}' by taking a fine grid of points inside the glyphs of a cropped image and using the horizontal coordinates as the input and the vertical coordinates as the output ($73,419$ points in total). Posteriors are shown in Figure \ref{fig:1D_posteriors} for four models (the same models as in Figure \ref{fig:prior_samples}), with the data plotted in the first figure. The \G model can obviously not model this data, but neither can the \GG 
% (VI-ELBO$=-1.0\times10^{5}$)
%
% (VI-ELBO$=-1.0\times10^{5}$) 
as the approximate posterior has no way to separate out the points with the same input but different outputs. The \LG model can fit the data to some extent, and the \LGG model more closely captures the shapes of the letters. 
% (VI-ELBO$=-7.9\times10^4$, IW-ELBO$=-6.8\times10^4$)
% , and the \LGG performs the best.
% (VI-ELBO$=-7.5\times10^4$, IW-ELBO$-5.9\times10^4$).

We see that the \LG has a tendency to smoother densities than \LGG, and also a smoother response as a function of the input. Between the `\textit{G}' and `\textit{P}' the \LGG model extrapolates with high confidence, whereas the \LG model maintains a bi-modal distribution connecting the two letters. Posteriors from further models are in the \supmat.

\subsection{UCI datasets}

\begin{table*}[h]
\centering
    \caption{Average test \loglikelihood results for five splits (standard error). \textbf{Bold} font indicates overlapping error bars with the best performing method. \textit{Italic} indicates a significantly non-Gaussian posterior, computed by the Shapiro–Wilk test (see Table \ref{tab:shapiro_wilk}). Colours indicate datasets demonstrating prominent examples of complex {\color{myAcolor} \textbf{marginals}\myAcolormarker}, {\color{myBcolor} \textbf{mappings}\myBcolormarker},~or {\color{myCcolor} \textbf{both}\myCcolormarker}.
    %\color{red}Red$^*$\color{black}~ datasets have complex marginals, in \color{blue}blue$^\dagger$\color{black}~ have complex input dependence, and in \color{Brown}brown^\ddagger$\color{black}~ have both complex input dependence and complex marginals. See table \ref{tab:all_methods} for a comparison with C-VAE and other approaches
    }
    \scalebox{0.69}{
    \begin{tabular}{llllllllllll}
\toprule
% \multicolumn{3}{l}{Model architecture}   
\multicolumn{3}{l}{}   
& \multicolumn{1}{c}{GP} &  \multicolumn{1}{c}{GP-GP}& \multicolumn{1}{c}{GP-GP-GP} &   \multicolumn{2}{c}{LV-GP} &   \multicolumn{2}{c}{LV-GP-GP} & 
 \multicolumn{2}{c}{LV-GP-GP-GP}  
\\
% \multicolumn{3}{l}{Importance weighted?} 
Dataset & N & D & 
\multicolumn{1}{c}{VI} &
\multicolumn{1}{c}{VI} &
\multicolumn{1}{c}{VI} &
\multicolumn{1}{c}{VI} &
\multicolumn{1}{c}{IWVI} &
\multicolumn{1}{c}{VI} &
\multicolumn{1}{c}{IWVI} &
\multicolumn{1}{c}{VI} &
\multicolumn{1}{c}{IWVI} \\
% & \multicolumn{1}{c}{-} & \multicolumn{1}{c}{-}  & \multicolumn{1}{c}{-}  &
% \multicolumn{1}{c}{$\times$} &  \multicolumn{1}{c}{$\checkmark$} &
% \multicolumn{1}{c}{$\times$} &  \multicolumn{1}{c}{$\checkmark$} &
% \multicolumn{1}{c}{$\times$} &  \multicolumn{1}{c}{$\checkmark$} \\
\midrule
challenger                                                &       23 &    4 &          -3.22 (0.07) &  \textit{-2.04 (0.02)} &  \textit{-2.10 (0.03)} &  \textit{\textbf{-0.67 (0.01)}} &  \textit{\textbf{-0.68 (0.00)}} &           \textit{-9.99 (3.60)} &           \textit{-9.47 (2.19)} &                    -1.21 (0.01) &           \textit{-2.45 (0.04)} \\
fertility                                                 &      100 &    9 &          -2.51 (0.13) &           -1.43 (0.01) &           -1.40 (0.01) &                    -1.41 (0.01) &                    -1.40 (0.01) &                    -1.42 (0.01) &                    -1.69 (0.01) &           \textbf{-1.32 (0.00)} &           \textbf{-1.31 (0.01)} \\
concreteslump                                             &      103 &    7 &           1.91 (0.01) &            1.55 (0.00) &            1.45 (0.00) &                     1.93 (0.00) &            \textbf{1.94 (0.00)} &                     1.53 (0.00) &                     1.75 (0.00) &                     1.43 (0.00) &                     1.62 (0.00) \\
autos                                                     &      159 &   25 &          -0.36 (0.00) &           -0.14 (0.01) &           -0.14 (0.03) &                    -0.36 (0.01) &                    -0.35 (0.00) &           \textbf{-0.06 (0.04)} &                    -0.34 (0.01) &                    -0.25 (0.05) &                    -0.33 (0.02) \\
servo                                                     &      167 &    4 &          -0.17 (0.00) &           -0.19 (0.01) &           -0.17 (0.01) &                    -0.07 (0.01) &                    -0.08 (0.01) &                    -0.19 (0.01) &                    -0.07 (0.01) &                    -0.25 (0.01) &            \textbf{0.03 (0.03)} \\
breastcancer                                              &      194 &   33 &          -1.32 (0.00) &           -1.36 (0.00) &           -1.34 (0.00) &           \textbf{-1.31 (0.00)} &           \textbf{-1.31 (0.00)} &                    -1.43 (0.01) &                    -1.59 (0.10) &                    -1.38 (0.00) &                    -1.80 (0.19) \\
machine                                                   &      209 &    7 &          -0.70 (0.01) &           -0.65 (0.02) &           -0.61 (0.01) &                    -0.75 (0.02) &                    -0.71 (0.02) &                    -0.63 (0.01) &           \textbf{-0.51 (0.01)} &                    -0.59 (0.01) &           \textbf{-0.52 (0.03)} \\
yacht                                                     &      308 &    6 &           1.32 (0.02) &            1.84 (0.01) &   \textbf{2.00 (0.06)} &                     1.64 (0.00) &                     1.65 (0.00) &                     1.69 (0.09) &                     1.71 (0.01) &                     1.77 (0.07) &            \textbf{2.05 (0.02)} \\
autompg                                                   &      392 &    7 &          -0.44 (0.01) &           -0.57 (0.02) &           -0.48 (0.03) &                    -0.32 (0.01) &                    -0.33 (0.01) &                    -0.38 (0.04) &           \textbf{-0.24 (0.01)} &                    -0.65 (0.11) &                    -0.28 (0.03) \\
boston                                                    &      506 &   13 &           0.02 (0.00) &            0.07 (0.00) &            0.03 (0.01) &                    -0.04 (0.00) &                    -0.07 (0.00) &                     0.06 (0.00) &                     0.14 (0.00) &                    -0.04 (0.01) &            \textbf{0.17 (0.00)} \\
{\color{myAcolor} \textbf{forest}\myAcolormarker}         &      517 &   12 &          -1.38 (0.00) &           -1.37 (0.00) &           -1.37 (0.00) &           \textit{-0.99 (0.00)} &           \textit{-1.00 (0.00)} &  \textit{\textbf{-0.91 (0.00)}} &           \textit{-1.02 (0.02)} &  \textit{\textbf{-0.92 (0.01)}} &  \textit{\textbf{-0.91 (0.01)}} \\
stock                                                     &      536 &   11 &          -0.22 (0.00) &           -0.29 (0.00) &           -0.26 (0.00) &                    -0.22 (0.00) &                    -0.22 (0.00) &                    -0.29 (0.00) &           \textbf{-0.19 (0.00)} &                    -0.23 (0.01) &                    -0.36 (0.02) \\
pendulum                                                  &      630 &    9 &          -0.14 (0.00) &            0.22 (0.01) &            0.12 (0.08) &                    -0.14 (0.00) &                    -0.14 (0.00) &                     0.22 (0.01) &            \textbf{0.33 (0.08)} &            \textbf{0.25 (0.00)} &                     0.20 (0.03) \\
energy                                                    &      768 &    8 &           1.71 (0.00) &            1.85 (0.00) &            2.07 (0.01) &                     1.86 (0.01) &                     1.92 (0.01) &                     2.00 (0.04) &                     1.96 (0.01) &                     2.07 (0.01) &            \textbf{2.28 (0.01)} \\
concrete                                                  &     1030 &    8 &          -0.43 (0.00) &           -0.45 (0.00) &           -0.35 (0.02) &                    -0.32 (0.00) &                    -0.32 (0.00) &                    -0.35 (0.00) &                    -0.21 (0.00) &                    -0.22 (0.01) &           \textbf{-0.12 (0.00)} \\
{\color{myAcolor} \textbf{solar}\myAcolormarker}          &     1066 &   10 &          -1.75 (0.08) &           -1.21 (0.02) &           -1.20 (0.03) &            \textit{0.04 (0.07)} &            \textit{0.07 (0.01)} &   \textit{\textbf{0.54 (0.01)}} &            \textit{0.22 (0.01)} &   \textit{\textbf{0.54 (0.01)}} &            \textit{0.20 (0.02)} \\
airfoil                                                   &     1503 &    5 &          -0.79 (0.05) &            0.08 (0.02) &            0.14 (0.03) &                    -0.44 (0.03) &                    -0.36 (0.01) &                     0.07 (0.00) &                     0.30 (0.00) &                    -0.02 (0.03) &            \textbf{0.34 (0.01)} \\
winered                                                   &     1599 &   11 &          -1.09 (0.00) &           -1.11 (0.00) &           -1.08 (0.00) &                    -1.07 (0.00) &                    -1.06 (0.00) &                    -1.10 (0.00) &  \textit{\textbf{-0.84 (0.01)}} &           \textit{-1.06 (0.03)} &           \textit{-1.31 (0.40)} \\
gas                                                       &     2565 &  128 &           1.07 (0.00) &   \textbf{1.60 (0.05)} &   \textbf{1.69 (0.05)} &            \textbf{1.69 (0.08)} &                     1.56 (0.06) &                     0.70 (0.10) &            \textbf{1.57 (0.16)} &                     1.30 (0.14) &            \textbf{1.61 (0.12)} \\
skillcraft                                                &     3338 &   19 &          -0.94 (0.00) &           -0.94 (0.00) &           -0.94 (0.00) &           \textbf{-0.91 (0.00)} &           \textbf{-0.91 (0.00)} &                    -0.92 (0.00) &                    -0.93 (0.00) &                    -0.92 (0.00) &                    -0.94 (0.00) \\
{\color{myBcolor} \textbf{sml}\myBcolormarker}            &     4137 &   26 &           1.53 (0.00) &            1.72 (0.01) &            1.83 (0.01) &                     1.52 (0.00) &                     1.52 (0.00) &                     1.79 (0.00) &                     1.91 (0.00) &            \textbf{1.92 (0.01)} &            \textbf{1.97 (0.05)} \\
winewhite                                                 &     4898 &   11 &          -1.14 (0.00) &           -1.14 (0.00) &           -1.14 (0.00) &                    -1.13 (0.00) &                    -1.13 (0.00) &                    -1.13 (0.00) &           \textbf{-1.10 (0.00)} &                    -1.13 (0.00) &           \textbf{-1.09 (0.00)} \\
parkinsons                                                &     5875 &   20 &           1.99 (0.00) &            2.61 (0.01) &            2.75 (0.02) &                     1.79 (0.02) &                     1.82 (0.02) &                     2.36 (0.02) &                     2.76 (0.02) &                     2.71 (0.07) &            \textbf{3.12 (0.05)} \\
{\color{myBcolor} \textbf{kin8nm}\myBcolormarker}         &     8192 &    8 &          -0.29 (0.00) &           -0.01 (0.00) &           -0.00 (0.00) &                    -0.29 (0.00) &                    -0.29 (0.00) &                    -0.02 (0.00) &                    -0.00 (0.00) &                    -0.00 (0.00) &            \textbf{0.03 (0.00)} \\
pumadyn32nm                                               &     8192 &   32 &           0.08 (0.00) &   \textbf{0.11 (0.01)} &   \textbf{0.11 (0.00)} &                    -1.44 (0.00) &                    -1.44 (0.00) &                     0.10 (0.01) &           \textit{-0.62 (0.25)} &            \textbf{0.11 (0.00)} &            \textbf{0.11 (0.00)} \\
{\color{myAcolor} \textbf{power}\myAcolormarker}          &     9568 &    4 &          -0.65 (0.04) &           -0.75 (0.03) &           -0.80 (0.02) &                    -0.39 (0.04) &                    -0.23 (0.02) &                    -0.36 (0.04) &           \textit{-0.28 (0.05)} &                    -0.25 (0.06) &  \textit{\textbf{-0.11 (0.06)}} \\
naval                                                     &    11934 &   14 &  \textbf{4.52 (0.02)} &            4.43 (0.03) &            4.35 (0.03) &            \textit{4.19 (0.01)} &            \textit{4.24 (0.01)} &                     4.36 (0.01) &            \textbf{4.52 (0.02)} &                     4.27 (0.02) &                     4.41 (0.02) \\
{\color{myCcolor} \textbf{pol}\myCcolormarker}            &    15000 &   26 &           0.48 (0.00) &            1.51 (0.01) &            1.45 (0.01) &            \textit{0.37 (0.08)} &           \textit{-0.50 (0.00)} &            \textit{2.34 (0.02)} &   \textit{\textbf{2.63 (0.01)}} &                     1.47 (0.01) &            \textbf{2.72 (0.10)} \\
elevators                                                 &    16599 &   18 &          -0.44 (0.00) &           -0.41 (0.01) &           -0.40 (0.00) &                    -0.37 (0.00) &                    -0.36 (0.00) &                    -0.29 (0.00) &                    -0.27 (0.00) &                    -0.28 (0.00) &           \textbf{-0.27 (0.00)} \\
{\color{myCcolor} \textbf{bike}\myCcolormarker}           &    17379 &   17 &           0.82 (0.01) &            3.49 (0.01) &            3.68 (0.01) &            \textit{2.48 (0.01)} &            \textit{2.66 (0.01)} &                     3.48 (0.00) &                     3.75 (0.01) &                     3.72 (0.01) &            \textbf{3.95 (0.01)} \\
{\color{myBcolor} \textbf{kin40k}\myBcolormarker}         &    40000 &    8 &           0.02 (0.00) &            0.84 (0.00) &            1.17 (0.00) &                     0.04 (0.00) &                     0.05 (0.00) &                     0.87 (0.01) &                     0.93 (0.00) &                     1.15 (0.00) &            \textbf{1.27 (0.00)} \\
protein                                                   &    45730 &    9 &          -1.06 (0.00) &           -0.98 (0.00) &           -0.95 (0.00) &           \textit{-0.85 (0.00)} &           \textit{-0.80 (0.00)} &           \textit{-0.70 (0.00)} &           \textit{-0.61 (0.00)} &           \textit{-0.68 (0.00)} &  \textit{\textbf{-0.57 (0.00)}} \\
tamielectric                                              &    45781 &    3 &          -1.43 (0.00) &           -1.43 (0.00) &           -1.43 (0.00) &  \textit{\textbf{-1.31 (0.00)}} &           \textit{-1.31 (0.00)} &  \textit{\textbf{-1.31 (0.00)}} &           \textit{-1.31 (0.00)} &           \textit{-1.31 (0.00)} &           \textit{-1.31 (0.00)} \\
{\color{myCcolor} \textbf{keggdirected}\myCcolormarker}   &    48827 &   20 &           0.16 (0.01) &            0.21 (0.01) &            0.26 (0.02) &            \textit{1.24 (0.06)} &            \textit{2.03 (0.01)} &            \textit{1.84 (0.05)} &            \textit{2.23 (0.01)} &            \textit{1.92 (0.02)} &   \textit{\textbf{2.26 (0.01)}} \\
{\color{myBcolor} \textbf{slice}\myBcolormarker}          &    53500 &  385 &           0.86 (0.00) &            1.80 (0.00) &            1.86 (0.00) &                     0.85 (0.00) &                     0.90 (0.00) &                     1.80 (0.00) &                     1.88 (0.00) &                     1.86 (0.00) &            \textbf{2.02 (0.00)} \\
{\color{myAcolor} \textbf{keggundirected}\myAcolormarker} &    63608 &   27 &           0.06 (0.01) &            0.06 (0.01) &            0.07 (0.01) &                \textit{-75 (4)} &           \textit{-4.21 (0.33)} &                \textit{-64 (6)} &            \textit{2.37 (0.29)} &              \textit{-116 (17)} &   \textit{\textbf{2.98 (0.21)}} \\
3droad                                                    &   434874 &    3 &          -0.79 (0.00) &           -0.61 (0.01) &           -0.58 (0.01) &           \textit{-0.71 (0.00)} &           \textit{-0.65 (0.00)} &           \textit{-0.69 (0.00)} &           \textit{-0.55 (0.01)} &           \textit{-0.61 (0.00)} &  \textit{\textbf{-0.48 (0.00)}} \\
song                                                      &   515345 &   90 &          -1.19 (0.00) &           -1.18 (0.00) &           -1.18 (0.00) &                    -1.15 (0.00) &                    -1.14 (0.00) &                    -1.13 (0.00) &           \textit{-1.10 (0.00)} &                    -1.12 (0.00) &  \textit{\textbf{-1.07 (0.00)}} \\
buzz                                                      &   583250 &   77 &          -0.24 (0.00) &           -0.15 (0.00) &           -0.15 (0.00) &                    -0.22 (0.01) &                    -0.23 (0.01) &                    -0.09 (0.01) &                    -0.04 (0.03) &                    -0.09 (0.01) &   \textit{\textbf{0.04 (0.00)}} \\
nytaxi                                                    &  1420068 &    8 &          -1.42 (0.01) &           -1.44 (0.01) &           -1.42 (0.01) &           \textit{-1.09 (0.03)} &  \textit{\textbf{-0.90 (0.04)}} &           \textit{-1.57 (0.04)} &           \textit{-1.03 (0.04)} &           \textit{-1.74 (0.05)} &  \textit{\textbf{-0.95 (0.07)}} \\
{\color{myCcolor} \textbf{houseelectric}\myCcolormarker}  &  2049280 &   11 &           1.37 (0.00) &            1.41 (0.00) &            1.47 (0.01) &            \textit{1.65 (0.03)} &            \textit{1.82 (0.01)} &                     1.66 (0.05) &            \textit{2.01 (0.00)} &                     1.77 (0.06) &   \textit{\textbf{2.03 (0.00)}} \\
\midrule
\multicolumn{3}{l}{Median difference from GP baseline} &
                      0 &                   0.06 &                   0.09 &                            0.04 &                            0.05 &                            0.12 &                            0.26 &                            0.18 &                            0.32 \\
\multicolumn{3}{l}{Average ranks} &
  2.72 (0.14) &            4.01 (0.14) &            4.84 (0.14) &                     4.19 (0.17) &                     4.60 (0.16) &                     4.84 (0.14) &                     6.69 (0.15) &                     5.62 (0.16) &                     7.49 (0.16)\\
\bottomrule
\end{tabular}
    }
    \label{tab:1D_results}
\end{table*}

We use 41 publicly available datasets\footnote{The full datasets with the splits and pre-processing can be found at \url{github.com/hughsalimbeni/bayesian_benchmarks}.} with 1D targets. The datasets range in size from $23$ points to $2,049,280$. In each case we reserve 10\% of the data for evaluating a test \loglikelihood, repeating the experiment five times with different splits. 
We use five samples for the importance-weighted models, $128$ inducing points, and five GP outputs for the inner layers. Hyperparameters and initializations are the same for all models and datasets and are fully detailed in the \supmat. Results for test log-likelihood are reported in Table \ref{tab:1D_results} for the GP models. Table \ref{tab:all_methods} in the \supmat shows additional results for conditional VAE models \cite{sohn2015learning}, and results from deep Gaussian process models using stochastic gradient Hamiltonian Monte Carlo \cite{havasi2018inference}. To assess the non-Gaussianity of the predictive distribution we compute the Shapiro–Wilk test statistic on the test point marginals. The test statistics are shown in \ref{tab:shapiro_wilk} in the \supmat. Full code to reproduce our results is available online \footnote{\url{https://github.com/hughsalimbeni/DGPs_with_IWVI}}. We draw five conclusions from the results.

%, (most rows are omitted due to lack of space; the complete table in the supplementary material, Table \ref{tab:1D_results_full}).% See the supplementary material for tables for the ELBOs for all the datasets. %We also ran all the experiments with the \GLG model (reported in the supplementary material), but obtain similar results to the \LGG, averaged over datasets. 
%

\textbf{1) Latent variables improve performance}. For a given depth of Gaussian process mappings (for example \GG compared to \LGG) the latent variable model generally has equal or better performance, and often considerably better. This is true both using IW and VI. %We see also that the latent variable modes capture non-Gaussian marginals on many of the datasets, whereas the noise-free models only have non-Gaussian marginals for the smallest datasets. %This is because the non-Gaussian marginals in the noise free case can only arise from epistemic uncertainty, which decreases when there is more data. 

%This is perhaps not surprising as the latent variable model can recover the noise-free version in the limit of a large lengthscale for the appropriate dimension.

\textbf{2) IW outperforms VI}. The difference is more pronounced on the deeper models, which is to be expected as the deeper models are likely to have more complicated posteriors, so the VI approximation is likely to be more severe. There are also a few datasets (`keggundirected', `servo' and `automgp') where the VI latent variable model has very poor performance, but the IW approach performs well. This suggests that the IW approach might be less prone to local optima compared with VI.

%, perhaps due to the VI approximation being more prone to local optima
% In a few cases the IW inference is less effective for the three layer model, which may be because the reduced flexibility of the posterior has the effect of regularizing the model.

\textbf{3) Some datasets require latent variables for any improvement over the single-layer model}. For several datasets (for example `forest' and `power') the inclusion of latent variables makes a very pronounced difference, whereas depth alone cannot improve over the single layer model. These datasets are highlighed in \myAcolorfull. For all these datasets the marginals for the latent variable models are non-Gaussian (see Table \ref{tab:shapiro_wilk} in the \supmat). Conversely, for some datasets (e.g., `kin40k' and `sml') we observe much greater benefit from the deep noise free models than the single layer model with latent variables. These datasets are highlighted in \myBcolorfull. % Interestingly, while these datasets have Gaussian marginals even for the latent variable models, the best performing model is still \LGGG. One explanation for could be that the inclusion of latent variables results in a easier optimization problem, even if their effect at the end of optimization is small.

%~ and are characterized by having non-Gaussian marginals but with simple input dependence. Plots of power and keggundirected posterior marginals are shown in \ref{fig:marginals} in the \supmat. 

\textbf{4) Some datasets benefit from both depth and latent variables}. For the `bike' and `keggdirected' data, for example, we see that both the \LG and \GG model improve over the single layer model, suggesting that complex input dependence and non-Gaussian marginals are beneficial for modelling the data. Four examples of these datasets are highlighted in \myCcolorfull. For these datasets the marginals are non-Gaussian for the latent variable models, and in each case the \LGGG model is the best performing. 

\textbf{5) On average, the \LGGG model is best performing}. This indicates that the inductive bias of this model is suitable over the broad range of datasets considered. We also compare to Conditional VAE models \citep{sohn2015learning} (see Table \ref{tab:all_methods} in the \supmat) and find that these models overfit considerably for the smaller datasets and even on the larger datasets do not outperform our deep latent variable models. The marginals of the CVAE models are more Gaussian than our models (see Table \ref{tab:shapiro_wilk} in the \supmat), indicating that these models explain the data through the input mapping and not the complexity of the marginal densities. See Figures \ref{fig:non-gaussian_marginals} and \ref{fig:both} in the \supmat for the plots of posterior marginals.

\section{Related Work}

If the GP layers are replaced by neural networks our models are equivalent to (conditional) variational autoencoders (VAE) \citep{kingma2013auto, rezende2014stochastic} when used with VI \ref{sec:VI}, or importance weighted autoencoders (IWAEs) \citep{burda2015importance} when used with the IW inference \ref{sec:IW_inference}. Our model can be thought of as a VAE (or IWAE) but with a deep GP for the `encoder' mapping. Compared to the conditional VAE, the deep GP approach has several advantages: it incorporates model (epistemic) uncertainty, is automatically regularized, and allows a fine-grained control of the properties of the mapping. For example, the lengthscale corresponding to the latent variable can be tuned to favour complex marginals (a small value) or near Gaussian marginals (a large value). Note that our IW inference is not a simple adaptation of \citet{burda2015importance} as we have to perform additional inference over the GPs. 

The \LG model was proposed by \citet{wang2012gaussian}, and then developed in \cite{dutordoir2018} and used for meta learning in \citet{saemundsson2018meta}. A version with discrete latent variables was presented in \citet{bodin2017latent}. The \LG without any inputs is known as the Gaussian process latent variable model \citep{lawrence2004gaussian}, which was used in a semi-supervised setting in \citet{damianou2015semi}, which is also equivalent to \LG. 

The deep model of \citet{damianou2013deep} is closely related to our model but incorporates the latent variables additively rather than through concatenation (that is, $f(g(x_n)+w_n)$ rather than $f(g([x_n, w_n]))$). In principle, it would be possible to recover our model using the approach of \citet{damianou2013deep} in a certain setting of hyperparameters, but in all previous work the kernel hyperparameters were tied within each layer, so this limit was not achievable. \citet{bui2016deep} also used this model, with a form of expectation propagation for approximate inference. 

The variational inference we have presented (without importance weighting) is not equivalent to that of \citet{damianou2013deep}. In \citet{damianou2013deep} a mean-field variational posterior is used for the noisy corruptions% (not the noise itself)
, which may be a poor approximation as there are a priori correlations between these outputs. This mean-field assumption also forces independence between the inputs and outputs of the layers, whereas we make no such assumption.

\section{Discussion}
On a broad range of 1D density estimation tasks we find that our DGP with latent variables outperforms the single-layer and noise-free models, sometimes considerably.  Closer investigation reveals that non-Gaussian marginals are readily found by our model, and that the importance-weighted objective improves performance in practice. 

Conditional density estimation must balance the complexity of the density with the complexity of the input dependency. The inductive bias of our \LGGG model appears to be suitable for a broad range of datasets we have considered. An advantage of our approach is that the deep models all contain the shallower models as special cases. A layer can be `turned off' with a single scalar hyperparameter (the kernel variance) set to zero. This is a consequence of the ResNet-inspired \citep{he2016deep} use of mean functions. This may explain why we observe empirically that in practice adding depth rarely hurts performance. The latent variables can similarly be `turned off' if the appropriate lengthscale is large. This may explain why the latent variables models are rarely outperformed by the noise-free models, even when the marginals are Gaussian. %We caution that the lower bound is not necessarily a reliable tool for model comparison, but we have seen that in practice it can be used effectively. This observation is consistent with previous work \citep[see][for a recent example]{van2018learning}. 

There are very few hyperparameters in our model to optimize (the kernel parameters and the likelihood variance). This allows us to use the same model across a wide range of data. In the small and medium data regimes we have considered, an unregularized mapping (for example, a neural network as in the conditional VAE model \citep{sohn2015learning}) is likely to overfit, as indeed we have observed (see Table \ref{tab:all_methods} in the \supmat).

%
% An advantage of the Bayesian approach is that models can be compared using the marginal likelihood. In recent work, \citet{van2018learning} used the marginal likelihood to infer invariance proprieties of the data.  

\subsection{Limitations}

As depth increases it becomes increasingly difficult to reason about the DGP priors. %Optimizing the linear projections makes the interpretation even more difficult. 
Our approach is unlikely to recover interpretable features, such as periodicity, and the approaches of  \citet{lloyd2014automatic, sun2018differentiable} may be more appropriate if an interpretable model is required. The latent variables are also difficult to interpret as their effect is coupled with the GP mappings.

We have only considered 1D latent variables and 1D outputs, and while the extension to higher dimensions is straightforward for training, difficulties may arise in evaluating posterior expectations as our model does not provide a closed-form predictive density and Monte Carlo sampling may have unacceptably high variance. The task of estimating test likelihoods with high-dimensional latent variables is challenging, and techniques described in \citet{wu16} may be necessary in higher dimensions. %Normalizing flows \citep{rezende2015variational} avoid this difficulty by using an invertible function to map from the latent variables to the data, though it is not clear how to place a non-parameteric prior on such functions. 

% In the 1D cases we have considered we were able to use sampling and kernel smoothing to estimate likelihood, but in higher dimensions this would become infeasible. 

% The use of IW inference makes it harder to reason about the latent variables. In the VI case the variational posterior can be interpreted as the `latent covariate', but in the

The inference we have presented is limited by cubic scaling in both $K$ and the number of inducing points. The importance-weighting approach may also suffer from problems of vanishing signal for parameters of $q(w_n)$, as discussed in \citet{rainforth2018tighter}. The doubly reparameterized gradient estimator from \citet{tucker2018doubly} could be used to alleviate this problem. %With $K=5$ and 100 inducing points we were able to run 20K iterations in a few minutes on a standard 8 core CPU machine. 

% It is not possible to make our approach deterministic except in the single-layer case, so that we are restricted to using stochastic optimizers. Unlike in the single-layer case, the DGP variational optimization problem is not convex, and may be multimodal, rendering results sensitive to initialization and random seeds, even for fixed hyperparameters.

% \subsection{Further work}
% A natural extension is to use importance weighted inference for the functions as well as the latent variable layers. This is much less straightforward, and would require joint sampling over the function values at the data, resulting in an $N^3$ complexity. 

% We have not considered the case where the marginals are complex but the mapping between inputs and output density is simple. This would be possible to achieve by mapping a latent variable through several GPs and then concatenating with $x$ before a final layer. Equivalently, input propagation with the GP inputs defined on a subset of dimensions could achieve the same model within our framework. 

% The independence assumptions between the variational distributions for $f$ and $g$. This assumption could be relaxed by coupling the inducing point distributions as in \citet{havasi2018inference}. 

\section{Conclusion}
We have presented a novel inference scheme for the deep Gaussian process with latent variables, combining importance weighting with partially collapsed variational inference. We have also developed a variant of the deep Gaussian process model where uncorrelated variables are introduced as latent covariates rather than process noise. %The latent covariate approach allows us to tune the inductive bias of the model, though in practice we find that placing latent variables at the first layer is more effective for real data. 
We have shown empirically that latent variables models deep models outperform the noise-free deep GP on a range of data, and also that our importance-weighted inference delivers an advantage over variational inference in practice. 

\section*{Acknowledgements}
We gratefully acknowledge the contributions of anonymous reviewers, and thank Sanket Kamthe, Victor Picheny and Alan Saul for feedback on the manuscript. 

% \clearpage
\bibliography{bib}

\begin{thebibliography}{37}
\providecommand{\natexlab}[1]{#1}
\providecommand{\url}[1]{\texttt{#1}}
\expandafter\ifx\csname urlstyle\endcsname\relax
  \providecommand{\doi}[1]{doi: #1}\else
  \providecommand{\doi}{doi: \begingroup \urlstyle{rm}\Url}\fi

\bibitem[Alvarez et~al.(2012)Alvarez, Rosasco, Lawrence,
  et~al.]{alvarez2012kernels}
Mauricio~A Alvarez, Lorenzo Rosasco, Neil~D Lawrence, et~al.
\newblock Kernels for vector-valued functions: A review.
\newblock \emph{Foundations and Trends in Machine Learning}, 2012.

\bibitem[Bodin et~al.(2017)Bodin, Campbell, and Ek]{bodin2017latent}
Erik Bodin, Neill~D Campbell, and Carl~H Ek.
\newblock {Latent Gaussian Process Regression}.
\newblock \emph{arXiv:1707.05534}, 2017.

\bibitem[Bui et~al.(2016)Bui, Hern{\'a}ndez-Lobato, Hernandez-Lobato, Li, and
  Turner]{bui2016deep}
Thang Bui, Daniel Hern{\'a}ndez-Lobato, Jose Hernandez-Lobato, Yingzhen Li, and
  Richard Turner.
\newblock Deep gaussian processes for regression using approximate expectation
  propagation.
\newblock \emph{International Conference on Machine Learning}, 2016.

\bibitem[Burda et~al.(2016)Burda, Grosse, and
  Salakhutdinov]{burda2015importance}
Yuri Burda, Roger Grosse, and Ruslan Salakhutdinov.
\newblock Importance weighted autoencoders.
\newblock \emph{International Conference on Learning Representations}, 2016.

\bibitem[Cheng and Boots(2016)]{cheng2016incremental}
Ching-An Cheng and Byron Boots.
\newblock {Incremental variational sparse Gaussian process regression}.
\newblock \emph{Advances in Neural Information Processing Systems}, 2016.

\bibitem[Dai et~al.(2015)Dai, Damianou, Gonz{\'a}lez, and
  Lawrence]{dai2015variational}
Zhenwen Dai, Andreas Damianou, Javier Gonz{\'a}lez, and Neil Lawrence.
\newblock {Variational auto-encoded deep Gaussian processes}.
\newblock \emph{International Conference on Learning Representations}, 2015.

\bibitem[Damianou and Lawrence(2013)]{damianou2013deep}
Andreas Damianou and Neil Lawrence.
\newblock Deep gaussian processes.
\newblock \emph{Artificial Intelligence and Statistics}, 2013.

\bibitem[Damianou and Lawrence(2015)]{damianou2015semi}
Andreas Damianou and Neil~D Lawrence.
\newblock {Semi-described and Semi-supervised Learning with Gaussian
  Processes}.
\newblock \emph{arXiv:1509.01168}, 2015.

\bibitem[Domke and Sheldon(2018)]{domke2018importance}
Justin Domke and Daniel Sheldon.
\newblock Importance weighting and varational inference.
\newblock \emph{Advances in Neural Information Processing Systems}, 2018.

\bibitem[Dunlop et~al.(2018)Dunlop, Girolami, Stuart, and
  Teckentrup]{dunlop2017deep}
Matthew~M Dunlop, Mark Girolami, Andrew~M Stuart, and Aretha~L Teckentrup.
\newblock {How Deep Are Deep Gaussian Processes?}
\newblock \emph{Journal of Machine Learning Research}, 2018.

\bibitem[Dutordoir et~al.(2018)Dutordoir, Salimbeni, Deisenroth, and
  Hensman]{dutordoir2018}
Vincent Dutordoir, Hugh Salimbeni, Marc~P Deisenroth, and James Hensman.
\newblock {Gaussian Process Conditional Density Estimation}.
\newblock \emph{Advances in Neural Information Processing Systems}, 2018.

\bibitem[Duvenaud et~al.(2014)Duvenaud, Rippel, Adams, and
  Ghahramani]{duvenaud2014avoiding}
David Duvenaud, Oren Rippel, Ryan Adams, and Zoubin Ghahramani.
\newblock Avoiding pathologies in very deep networks.
\newblock \emph{Artificial Intelligence and Statistics}, 2014.

\bibitem[Glorot and Bengio(2010)]{glorot2010understanding}
Xavier Glorot and Yoshua Bengio.
\newblock Understanding the difficulty of training deep feedforward neural
  networks.
\newblock \emph{Artificial Intelligence and Statistics}, 2010.

\bibitem[Havasi et~al.(2018)Havasi, Lobato, and Fuentes]{havasi2018inference}
Marton Havasi, Jos{\'e} Miguel~Hern{\'a}ndez Lobato, and Juan Jos{\'e}~Murillo
  Fuentes.
\newblock Inference in deep gaussian processes using stochastic gradient
  hamiltonian monte carlo.
\newblock \emph{Advances in Neural Information Processing Systems}, 2018.

\bibitem[He et~al.(2016)He, Zhang, Ren, and Sun]{he2016deep}
Kaiming He, Xiangyu Zhang, Shaoqing Ren, and Jian Sun.
\newblock Deep residual learning for image recognition.
\newblock \emph{The IEEE conference on computer vision and pattern
  recognition}, 2016.

\bibitem[Hebbal et~al.(2019)Hebbal, Brevault, Balesdent, Talbi, and
  Melab]{bayesian2019}
Ali Hebbal, Loic Brevault, Mathieu Balesdent, El-Ghazali Talbi, and Nouredine
  Melab.
\newblock {Bayesian Optimization using Deep Gaussian Processes}.
\newblock \emph{arxiv/1905.03350}, 2019.

\bibitem[Hensman et~al.(2012)Hensman, Rattray, and Lawrence]{hensman2012fast}
James Hensman, Magnus Rattray, and Neil~D Lawrence.
\newblock Fast variational inference in the conjugate exponential family.
\newblock \emph{Advances in neural information processing systems}, 2012.

\bibitem[Hensman et~al.(2013)Hensman, Fusi, and Lawrence]{hensman2013}
James Hensman, Nicolo Fusi, and Neil~D Lawrence.
\newblock {Gaussian Processes for Big Data}.
\newblock \emph{Uncertainty in Artificial Intelligence}, 2013.

\bibitem[Kingma and Ba(2015)]{kingma2014adam}
Diederik~P Kingma and Jimmy Ba.
\newblock Adam: A method for stochastic optimization.
\newblock \emph{International Conference on Learning Representations}, 2015.

\bibitem[Kingma and Welling(2014)]{kingma2013auto}
Diederik~P Kingma and Max Welling.
\newblock {Auto-encoding Variational Bayes}.
\newblock \emph{International Conference for Learning Representations}, 2014.

\bibitem[Lawrence(2004)]{lawrence2004gaussian}
Neil~D Lawrence.
\newblock {Gaussian Process Latent Variable Models for Visualisation of High
  Dimensional Data}.
\newblock \emph{Advances in Neural Information Processing Systems}, 2004.

\bibitem[Lloyd et~al.(2014)Lloyd, Duvenaud, Grosse, Tenenbaum, and
  Ghahramani]{lloyd2014automatic}
James~Robert Lloyd, David~K Duvenaud, Roger~B Grosse, Joshua~B Tenenbaum, and
  Zoubin Ghahramani.
\newblock Automatic construction and natural-language description of
  nonparametric regression models.
\newblock \emph{The AAAI Conference on Artificial Intelligence}, 2014.

\bibitem[Matthews et~al.(2016)Matthews, Hensman, Richard, and
  Ghahramani]{matthews16}
Alexander Matthews, James Hensman, Turner Richard, and Zoubin Ghahramani.
\newblock {On Sparse Variational Methods and the Kullback-Leibler Divergence
  between Stochastic Processes}.
\newblock \emph{Artificial Intelligence and Statistics}, 2016.

\bibitem[Matthews et~al.(2017)Matthews, van~der Wilk, Nickson, Fujii,
  Boukouvalas, Le{\'o}n-Villagr{\'a}, Ghahramani, and
  Hensman]{matthews2016gpflow}
Alexander G de~G Matthews, Mark van~der Wilk, Tom Nickson, Keisuke Fujii,
  Alexis Boukouvalas, Pablo Le{\'o}n-Villagr{\'a}, Zoubin Ghahramani, and James
  Hensman.
\newblock {GPflow: a Gaussian process library using TensorFlow}.
\newblock \emph{The Journal of Machine Learning Research}, 2017.

\bibitem[Neil(1998)]{neil1998regression}
Radford~M Neil.
\newblock Regression and classification using gaussian process priors.
\newblock \emph{Bayesian statistics}, 1998.

\bibitem[Rainforth et~al.(2018)Rainforth, Kosiorek, Le, Maddison, Igl, Wood,
  and Teh]{rainforth2018tighter}
Tom Rainforth, Adam~R Kosiorek, Tuan~Anh Le, Chris~J Maddison, Maximilian Igl,
  Frank Wood, and Yee~Whye Teh.
\newblock Tighter variational bounds are not necessarily better.
\newblock \emph{International Conference on Machine Learning}, 2018.

\bibitem[Rezende et~al.(2014)Rezende, Mohamed, and
  Wierstra]{rezende2014stochastic}
Danilo~Jimenez Rezende, Shakir Mohamed, and Daan Wierstra.
\newblock {Stochastic Backpropagation and Approximate Inference in Deep
  Generative Models}.
\newblock \emph{International Conference on Machine Learning}, 2014.

\bibitem[S{\ae}mundsson et~al.(2018)S{\ae}mundsson, Hofmann, and
  Deisenroth]{saemundsson2018meta}
Steind{\'o}r S{\ae}mundsson, Katja Hofmann, and Marc~Peter Deisenroth.
\newblock {Meta Reinforcement Learning with Latent Variable Gaussian
  Processes}.
\newblock \emph{Uncertainty in Artificial Intelligence}, 2018.

\bibitem[Salimbeni and Deisenroth(2017)]{salimbeni2017doubly}
Hugh Salimbeni and Marc~P Deisenroth.
\newblock {Doubly Stochastic Variational Inference for Deep Gaussian
  Processes}.
\newblock \emph{Advances in Neural Information Processing Systems}, 2017.

\bibitem[Salimbeni et~al.(2018)Salimbeni, Eleftheriadis, and
  Hensman]{salimbeni2018natural}
Hugh Salimbeni, Stefanos Eleftheriadis, and James Hensman.
\newblock {Natural Gradients in Practice: Non-Conjugate Variational Inference
  in Gaussian Process Models}.
\newblock \emph{Artificial Intelligence and Statistics}, 2018.

\bibitem[Sohn et~al.(2015)Sohn, Lee, and Yan]{sohn2015learning}
Kihyuk Sohn, Honglak Lee, and Xinchen Yan.
\newblock Learning structured output representation using deep conditional
  generative models.
\newblock In \emph{Advances in Neural Information Processing Systems}, 2015.

\bibitem[Sun et~al.(2018)Sun, Zhang, Wang, Zeng, Li, and
  Grosse]{sun2018differentiable}
Shengyang Sun, Guodong Zhang, Chaoqi Wang, Wenyuan Zeng, Jiaman Li, and Roger
  Grosse.
\newblock Differentiable compositional kernel learning for gaussian processes.
\newblock \emph{International Conference on Machine Learning}, 2018.

\bibitem[Titsias(2009)]{titsias2009}
Michalis Titsias.
\newblock {Variational Learning of Inducing Variables in Sparse {G}aussian
  Processes}.
\newblock \emph{Artificial Intelligence and Statistics}, 2009.

\bibitem[Titsias and Lawrence(2010)]{titsias2010bayesian}
Michalis Titsias and Neil~D Lawrence.
\newblock {Bayesian {G}aussian Process Latent Variable Model}.
\newblock \emph{Artificial Intelligence and Statistics}, 2010.

\bibitem[Tucker et~al.(2019)Tucker, Lawson, Gu, and Maddison]{tucker2018doubly}
George Tucker, Dieterich Lawson, Shixiang Gu, and Chris~J Maddison.
\newblock {Doubly Reparameterized Gradient Estimators for Monte Carlo
  Objectives}.
\newblock \emph{International Conference on Learning Representations}, 2019.

\bibitem[Wang and Neal(2012)]{wang2012gaussian}
Chunyi Wang and Radford Neal.
\newblock {Gaussian Process Regression with Heteroscedastic or Non-Gaussian
  Residuals}.
\newblock \emph{arXiv:1212.6246}, 2012.

\bibitem[Wu et~al.(2016)Wu, Burda, Salakhutdinov, and Grosse]{wu16}
Yuhuai Wu, Yuri Burda, Ruslan Salakhutdinov, and Roger Grosse.
\newblock {On the Quantitative Analysis of Decoder-based Generative Models}.
\newblock \emph{International Conference on Learning Representations}, 2016.

\end{thebibliography}
\bibliographystyle{plainnat}

\onecolumn
\clearpage
\section*{\Supmat}
\appendix

\section{Hyperparameters and training settings}\label{sec:hyperparameters}
For all our models we used the following settings in the UCI experiments:

\textbf{Kernels}. All kernels were the RBF, using a lengthscale parameter per input dimension, initialized to the squareroot of the dimension. 

\textbf{Inducing points}. For data with more than 128 training data points the inducing point locations were chosen using the \texttt{kmeans2} from the \texttt{scipy} package, with 128 points. Otherwise they were set to the data. For latent variable layers, the GP layer above has extra dimensions. The inducing points for the extra dimensions were padded with random draws from a standard normal variable. 

\textbf{Linear projections between layers}. We implemented the linear mean functions and multioutput structure using a linear projection of 5 independent GPs concatenated with their inputs. We initialized the projection matrix to the first 5 principle components of the data concatenated with the identity matrix. %We optimized the components of the projection matrix. 

\textbf{Amortization of the variational parameters}. We used three layer fully connected network with skip connections between each layer. We used the $\tanh$ non-linearity with 10 units for the inner layers. We used the weight initialization from \citep{glorot2010understanding} and the exponential function to enforce positivity of the standard deviation parameters. We added a bias of -5 in the final layer to ensure the standard deviations were small at the start of optimization. 

\textbf{Likelihood}. The likelihood variance was intialized 0.01.

%\textbf{Latent variables}. The prior for the latent variables was $\mathcal{N}(0, 1). %initialized to 1 and optimized (the prior mean was fixed to zero). There is a degeneracy between optimizing the latent variable prior standard deviation and the lengthscale of the GP layer above, but we optimized both as the infinite/zero limits are more readily attainable. 

\textbf{Parameterizations}. All positive model parameters were constrained to be positive using the softplus function, clipped to $10^{-6}$. The variational parameters for the sparse GP layers were parameterized by the mean and the square root of the covariance. 

\textbf{Optimization}. The final GP layers were optimized using natural gradients on the natural parameters, with an initial step size of 0.01. All other parameters were optimized using the Adam optimizer \citep{kingma2014adam} with all parameters set to their tensorflow default values except the initial step size of 0.005. Optimizing just final layer using natural gradients and the inner layers with the Adam optimizer in this way is show by \citet{bayesian2019} to be an effective strategy in practice. We used a batch size of 512 and trained for 100K iterations, annealing the learning rate of both the adam and natural gradient steps by a factor of 0.98 per 1000 iterations. The mean functions and kernel projection matrices (the $P$ matrices that correlate the outputs) were \emph{not} optimized. %We did also ran a parallel set of experiments where we did optimize these matrices, and obtained similar or better results on the larger datasets, but observed overfitting on the small datasets. The overall conclusion of the pape

\textbf{Importance weights}. We used 5 importance weights for all the latent variables models with importance-weighted variational inference. % For the other models we used 5 samples also to equalize computation. 

\textbf{Predictions}. We used 2000 samples for prediction, sampling from the prior for $w$ for the latent variables models. We then used a kernel density estimator (with Silverman's rule to set the bandwidth) to estimate the density. 

\textbf{Splits and preprocessing}. We used 90\% splits on the shuffled data, and rescaled the inputs and outputs to unit standard deviation and zero mean. For dataset with more than 10K test points we used a random sample of 10K test points (the same sample for each split). NB we have reported the test {\loglikelihood}s without restoring the scaling, to facilitate comparisons between data. The splits were the same for all the models. We implemented our models using gpflow \citep{matthews2016gpflow}. 

\textbf{Comparison of methods}. The error bars in Table \ref{tab:1D_results} are standard errors over the five splits. These error bars are likely to \emph{overestimate} the uncertainty for comparisons between methods, however, as there may be correlations between the performance of methods over the splits (as the splits are the same for each method). The average ranks were computed for each split separately, and then averaged over the splits and method. To mitigate the effect of correlations between splits we report the median difference in test log-likelihood from the single layer GP. 

\textbf{VAE models} The conditional VAE models used $\tanh$ activations and weight initializations from \citep{glorot2010understanding}. Optimization was performed used the adam optimizer, with the same batch size and learning rate schedule as the GP models. 

\clearpage
\section{Variational posterior derivation}\label{sec:var_posterior_derivation}
Here we derive the sparse GP posterior $q(f)$ and its $\KL$ divergence from the prior, following \citet{hensman2013}. This derivation applies to all the GP layers in the DGP model.

We begin with the approach of \citet{titsias2009} and form a variational distribution over the function $f$ by conditioning the prior on a set of inducing points $\{f(\tilde{x}_i)\}_{i=1}^{M}$. We write $\tilde{\mathbf f}$ for the vector with $i$th component $f(\tilde{x}_i)$. In \citet{titsias2009}, $q(f)$ is defined as 
\begin{align}\label{eq:prior_conditional}
    q(f) = p(f|\tilde{\mathbf f}) q(\tilde{\mathbf f})\,,
\end{align}
where $q(\tilde{\mathbf f})$ is unspecified. In the single layer case, analytic results can be used to show that $q(\tilde{\mathbf f})$ has an optimal form that is Gaussian. Instead, we follow \citet{hensman2013} and \emph{assume} that 
\begin{align}
    q(\tilde{\mathbf f})=\mathcal N(\tilde{\mathbf f}|\tilde{\mathbf m}, \tilde{\mathbf S})\,.
\end{align} 
As this choice is conjugate to the prior conditional $p(f|\tilde{\mathbf f})$, we can use standard Gaussians identities to show that 
\begin{align}
q(f)=\mathcal{GP}(\mu, \Sigma)\,,    
\end{align}
where 
\begin{align}
    \mu(x) =& \mathbf k(x)^{\top} \mathbf {\tilde K}^{-1} \mathbf m\\
    \Sigma(x, x') = & k(x, x') - \mathbf k(x)^{\top} \mathbf {\tilde K}^{-1}\left(\mathbf{\tilde K} - \mathbf S\right)\mathbf {\tilde K}^{-1} \mathbf k(x)
\end{align}
with $[\mathbf k(x)]_i = k(x, \tilde{x}_i)$ and $[\mathbf {\tilde K}]_{ij} = k(\tilde{x_i}, \tilde{x_j})$
The $\KL$ divergence is given by
\begin{align}
    \KL(q(f)||p(f)) =&- \mathbb E_{q(f)} \log \frac{p(f)}{q(f)}\\
    =& -\mathbb E_{q(f)} \log \frac{p(f|\tilde{\mathbf f})p(\tilde{\mathbf f})}{p(f|\tilde{\mathbf f})q(\tilde{\mathbf f})}\\
    =&- \mathbb E_{q(\tilde{\mathbf f})} \log \frac{p(\tilde{\mathbf f})}{q(\tilde{\mathbf f})}\\
    =& \KL(q(\tilde{\mathbf f})||p(\tilde{\mathbf f}))\,,
\end{align}
which is finite. 

The parameters $\tilde{\mathbf m}$ and $\tilde{\mathbf S}$ together with $\{{\tilde x}_i\}$ are variational parameters. 

For multiple outputs, use the same matrix $P$ for correlating the outputs as the prior. In practice, we implement the correlated output model by multiplying independent GPs by the matrix $P$, so we use independent outputs for the variational distribution. 

% \clearpage
\section{Further tables and figures}\label{sec:tables_and_figs}

% \begin{table}[h]
% \scalebox{0.65}{\input{figs/tables/regression.tex}}
% \caption{Full test \loglikelihood results for the UCI datasets}
% \label{tab:1D_results_full}
% \end{table}

% \begin{table}[h]
% \scalebox{0.65}{\input{figs/tables/regression_elbo.tex}}
% \caption{ELBO results for the UCI datasets}
% \label{tab:1D_results_full_ELBO}
% \end{table}

% \clearpage

\begin{figure*}[h]
\centering
\subfloat[][\LG, VI]{
  \includegraphics[width=0.24\linewidth]{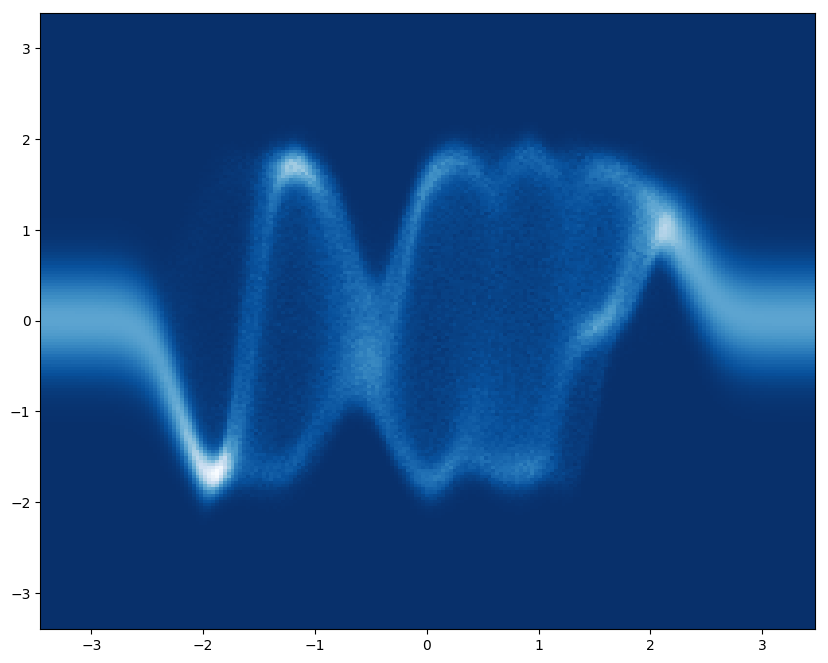}
 }
 \subfloat[][\LGG, VI]{
  \includegraphics[width=0.24\linewidth]{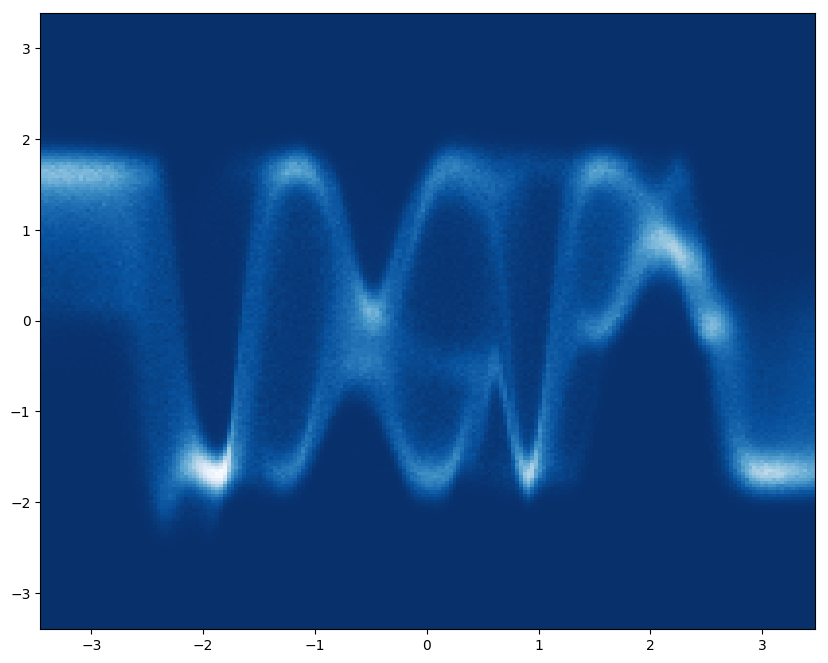}
 }
 \subfloat[][\LGGG, VI]{
  \includegraphics[width=0.24\linewidth]{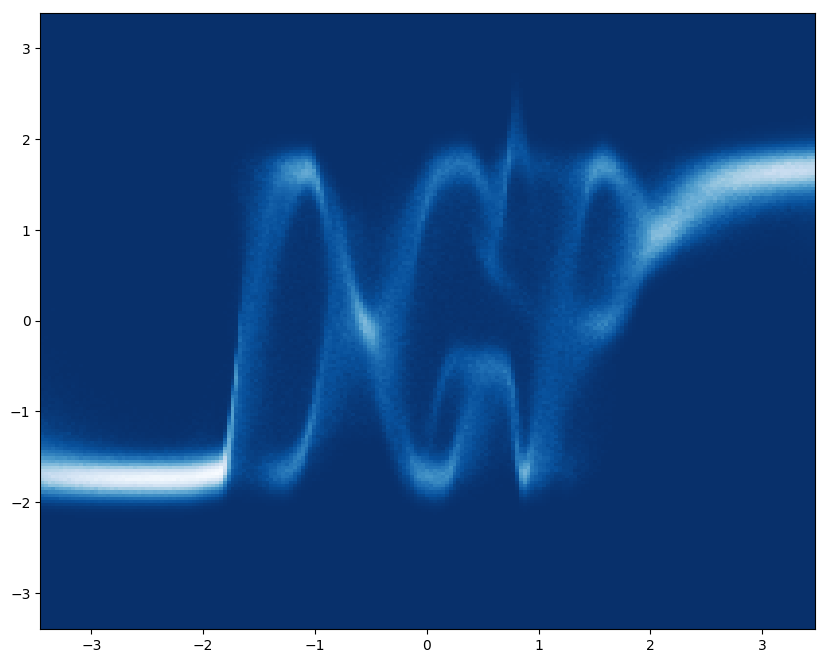}
 }
 \subfloat[][\LGGG, IW]{
  \includegraphics[width=0.24\linewidth]{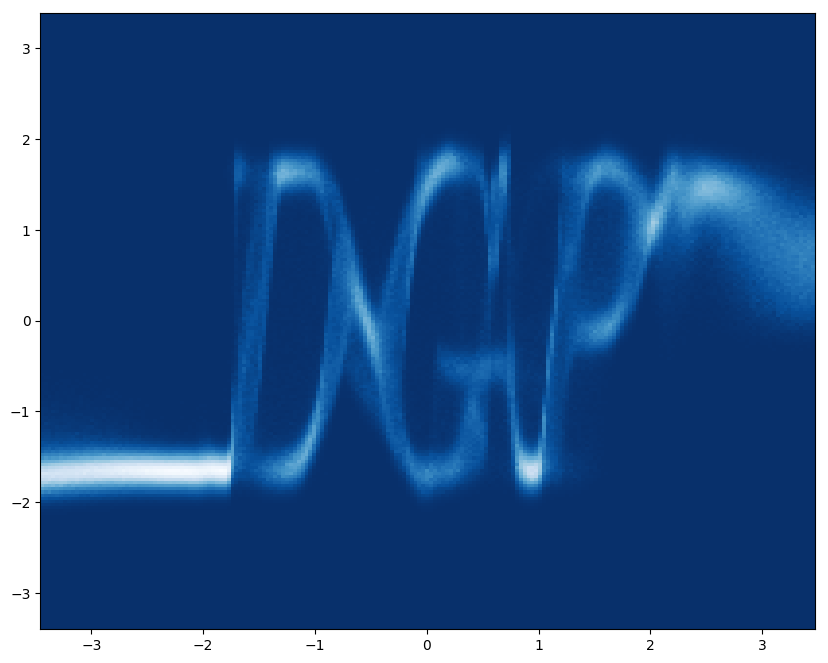}
 }
\caption{Posteriors for further models for the `\textit{DGP}' data}
\label{fig:further_DGP_figs}
\end{figure*}

% \clearpage
% \section*{Prior draws}
\begin{figure*}[h]
\centering
\subfloat[][\G]{
  \includegraphics[width=0.19\linewidth]{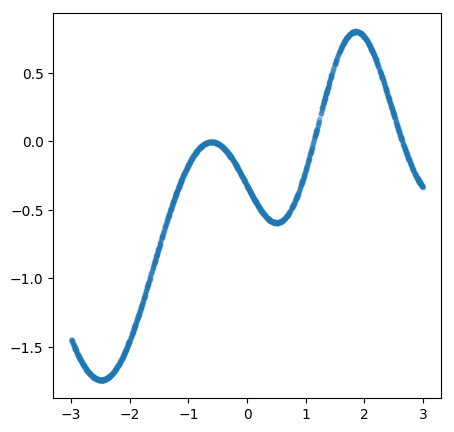}
 }
 \subfloat[][\G]{
  \includegraphics[width=0.19\linewidth]{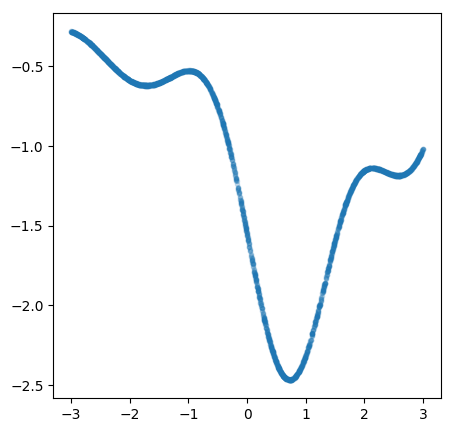}
 }
 \subfloat[][\G]{
  \includegraphics[width=0.19\linewidth]{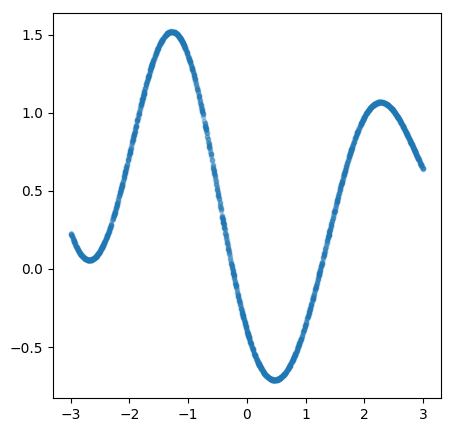}
 }
 \subfloat[][\G]{
  \includegraphics[width=0.19\linewidth]{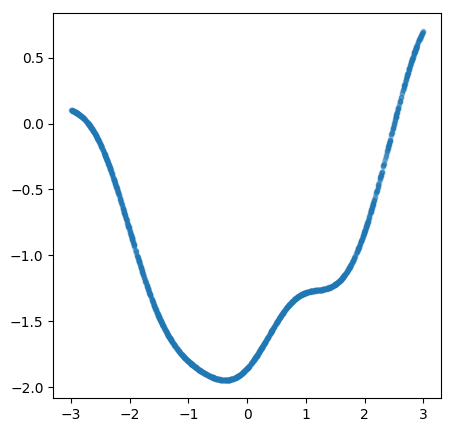}
 }
 \subfloat[][\G]{
  \includegraphics[width=0.19\linewidth]{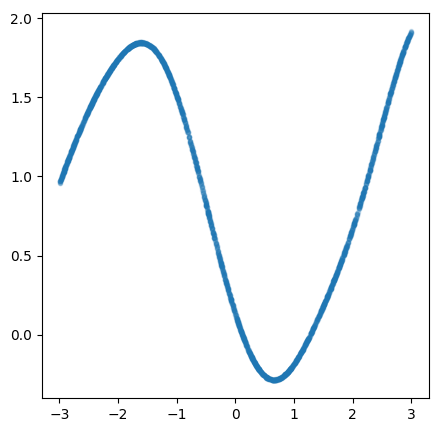}
 }\\
  \subfloat[][\GG]{
  \includegraphics[width=0.19\linewidth]{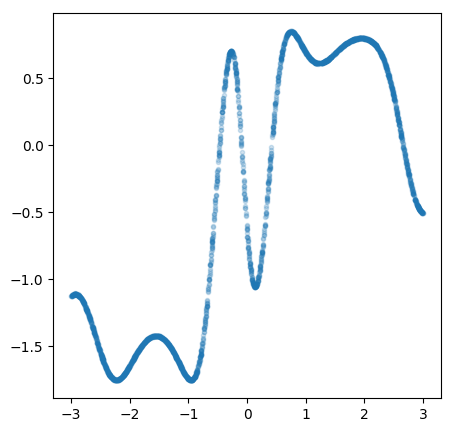}
 }
 \subfloat[][\GG]{
  \includegraphics[width=0.19\linewidth]{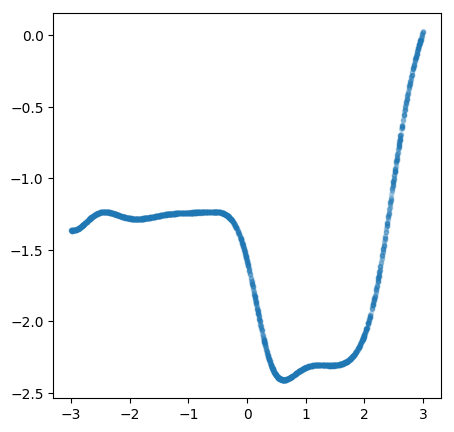}
 }
 \subfloat[][\GG]{
  \includegraphics[width=0.19\linewidth]{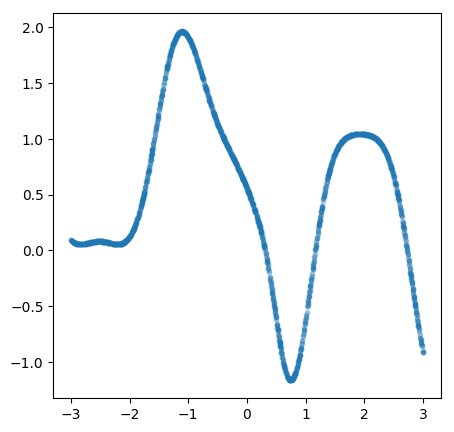}
 }
 \subfloat[][\GG]{
  \includegraphics[width=0.19\linewidth]{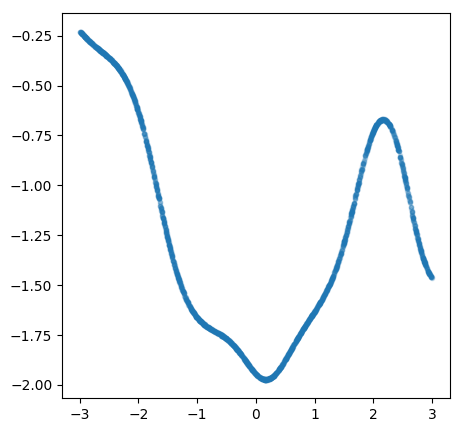}
 }
 \subfloat[][\GG]{
  \includegraphics[width=0.19\linewidth]{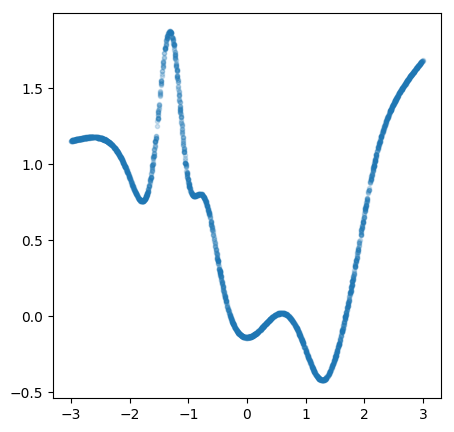}
 }\\
  \subfloat[][\GGG]{
  \includegraphics[width=0.19\linewidth]{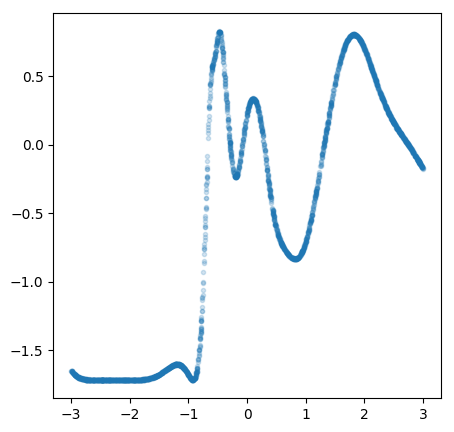}
 }
 \subfloat[][\GGG]{
  \includegraphics[width=0.19\linewidth]{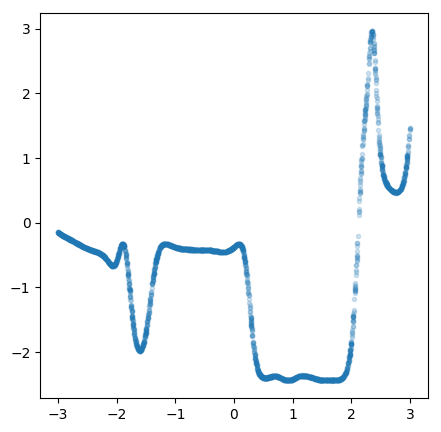}
 }
 \subfloat[][\GGG]{
  \includegraphics[width=0.19\linewidth]{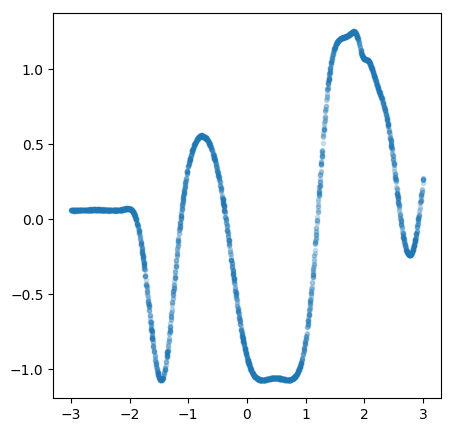}
 }
 \subfloat[][\GGG]{
  \includegraphics[width=0.19\linewidth]{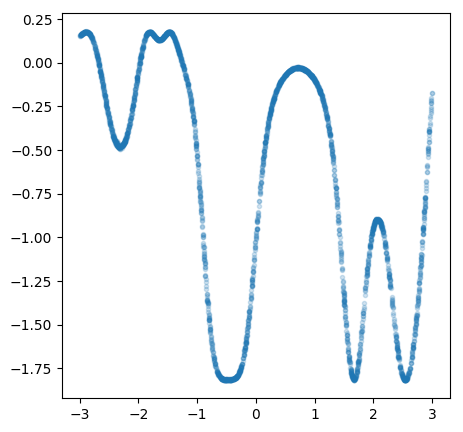}
 }
 \subfloat[][\GGG]{
  \includegraphics[width=0.19\linewidth]{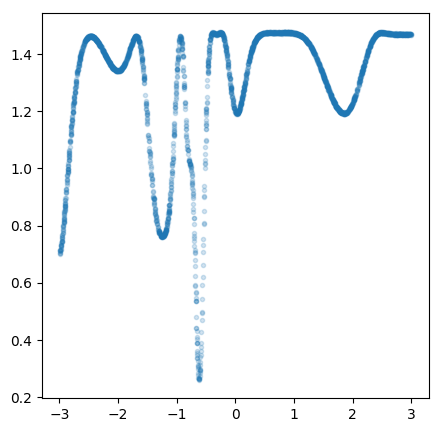}
 }\\
\caption{Prior draws from one, two and three layer models without latent variables}
\label{fig:without_latent_vars}
\end{figure*}

\begin{figure*}[h]
\centering
\subfloat[][\LG]{
  \includegraphics[width=0.19\linewidth]{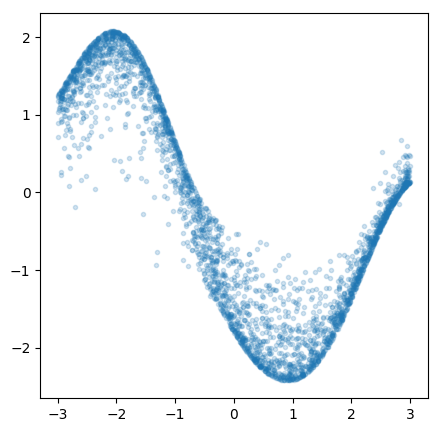}
 }
 \subfloat[][\LG]{
  \includegraphics[width=0.19\linewidth]{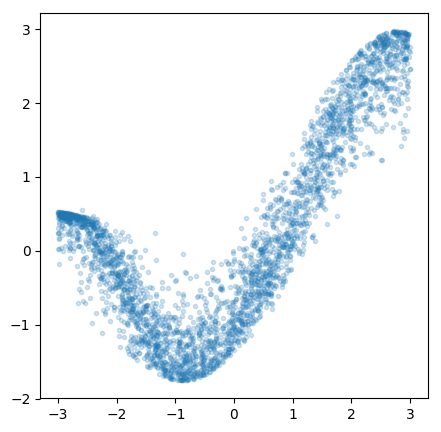}
 }
 \subfloat[][\LG]{
  \includegraphics[width=0.19\linewidth]{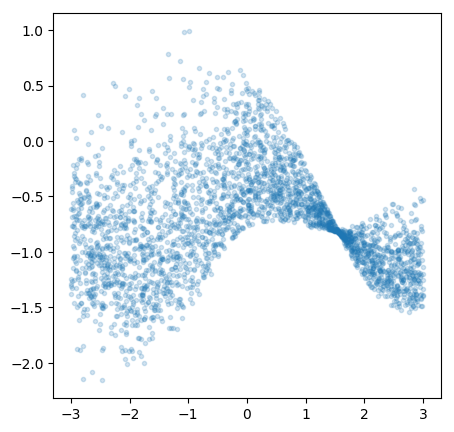}
 }
 \subfloat[][\LG]{
  \includegraphics[width=0.19\linewidth]{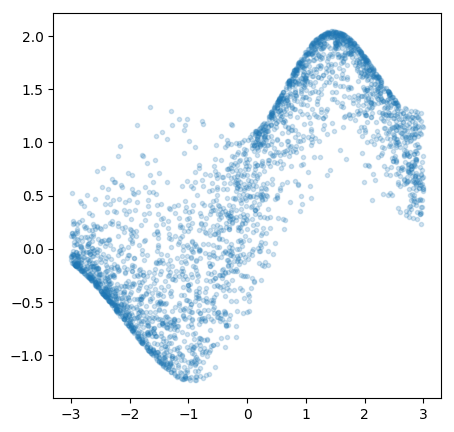}
 }
 \subfloat[][\LG]{
  \includegraphics[width=0.19\linewidth]{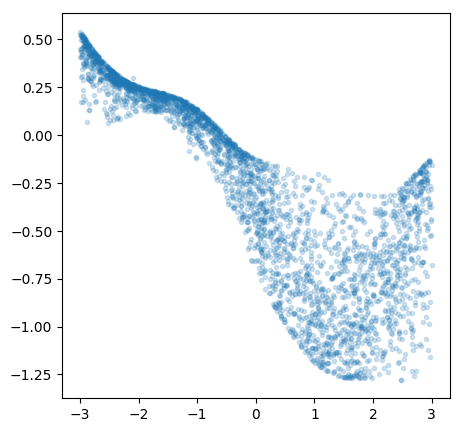}
 }
\caption{Prior draws from the \LG model}
\label{fig:lg}
\end{figure*}

\begin{figure*}
\centering
\subfloat[][\GLG]{
  \includegraphics[width=0.19\linewidth]{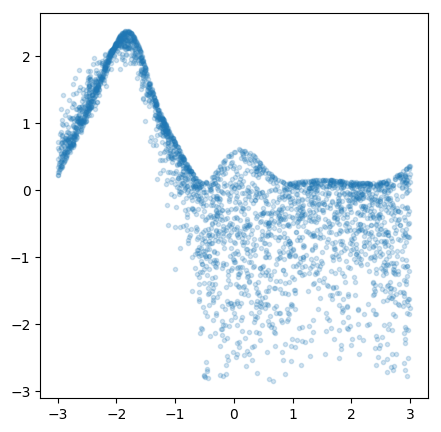}
 }
 \subfloat[][\GLG]{
  \includegraphics[width=0.19\linewidth]{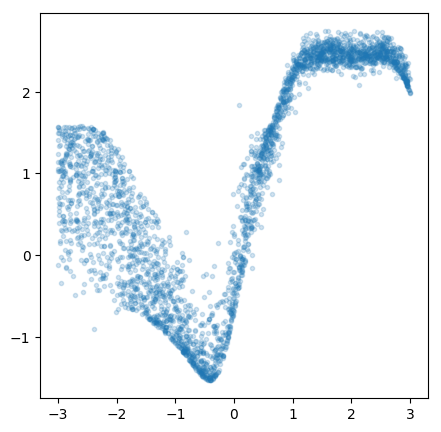}
 }
 \subfloat[][\GLG]{
  \includegraphics[width=0.19\linewidth]{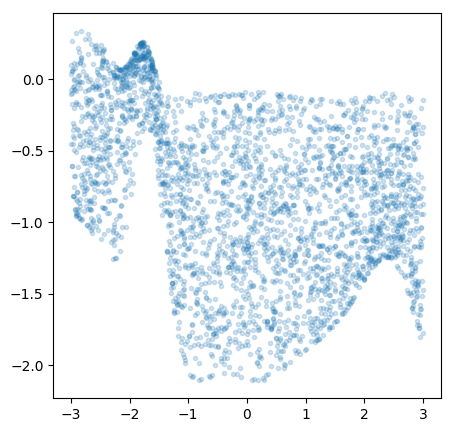}
 }
 \subfloat[][\GLG]{
  \includegraphics[width=0.19\linewidth]{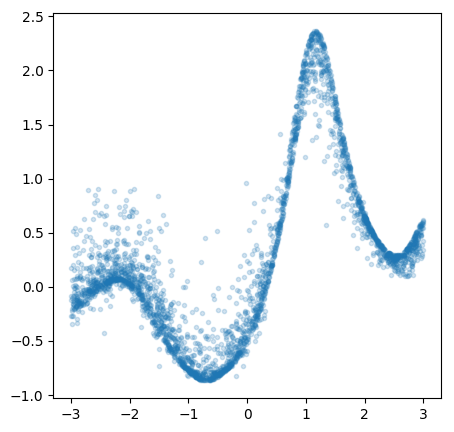}
 }
 \subfloat[][\GLG]{
  \includegraphics[width=0.19\linewidth]{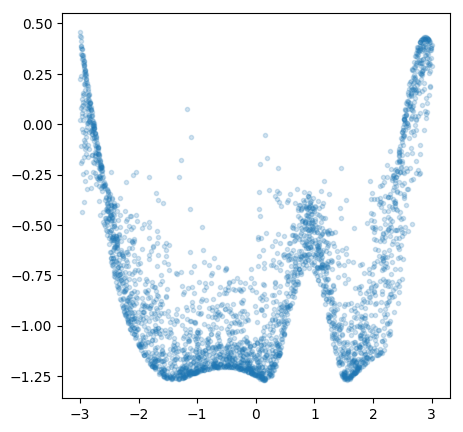}
 }\\
 \subfloat[][\LGG]{
  \includegraphics[width=0.19\linewidth]{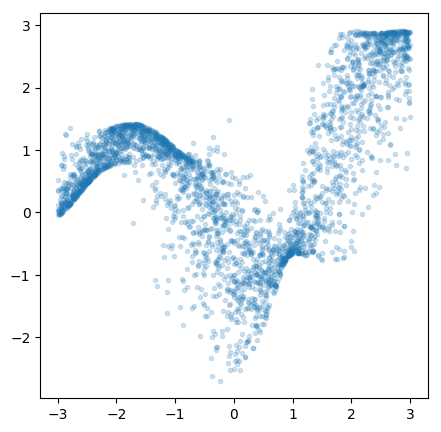}
 }
 \subfloat[][\LGG]{
  \includegraphics[width=0.19\linewidth]{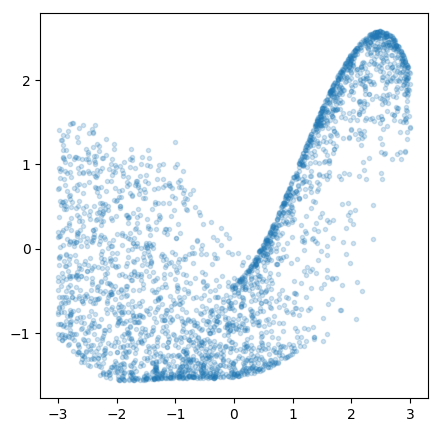}
 }
 \subfloat[][\LGG]{
  \includegraphics[width=0.19\linewidth]{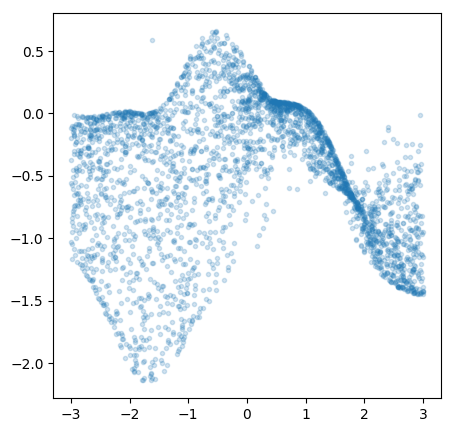}
 }
 \subfloat[][\LGG]{
  \includegraphics[width=0.19\linewidth]{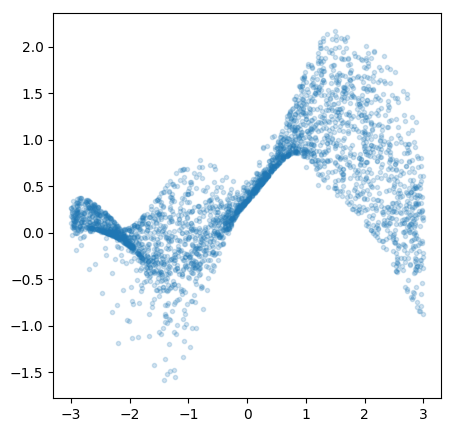}
 }
 \subfloat[][\LGG]{
  \includegraphics[width=0.19\linewidth]{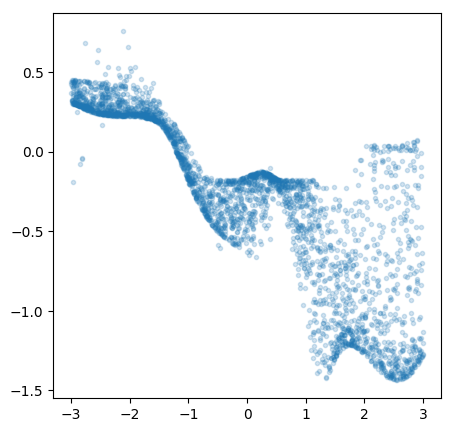}
 }
\caption{Prior draws from the \LGG and \GLG models}
\label{fig:lgg_glg}
\end{figure*}

\begin{figure*}
\centering
\subfloat[][\GGLG]{
  \includegraphics[width=0.19\linewidth]{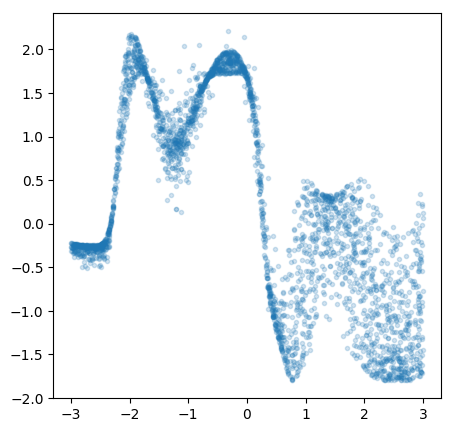}
 }
 \subfloat[][\GGLG]{
  \includegraphics[width=0.19\linewidth]{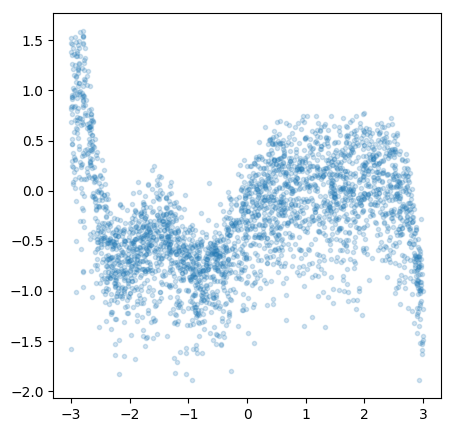}
 }
 \subfloat[][\GGLG]{
  \includegraphics[width=0.19\linewidth]{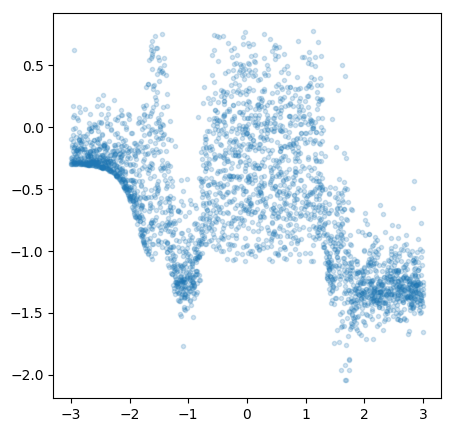}
 }
 \subfloat[][\GGLG]{
  \includegraphics[width=0.19\linewidth]{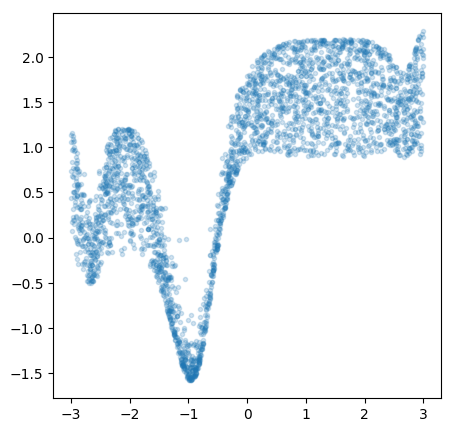}
 }
 \subfloat[][\GGLG]{
  \includegraphics[width=0.19\linewidth]{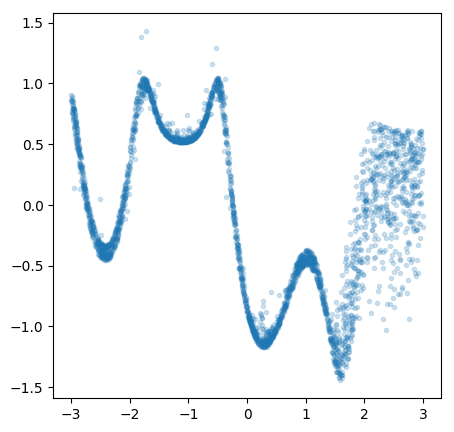}
 }\\
 \subfloat[][\GLGG]{
  \includegraphics[width=0.19\linewidth]{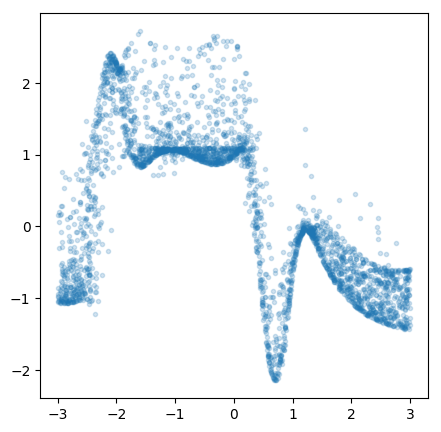}
 }
 \subfloat[][\GLGG]{
  \includegraphics[width=0.19\linewidth]{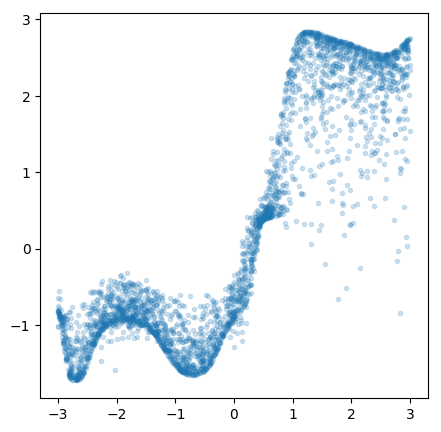}
 }
 \subfloat[][\GLGG]{
  \includegraphics[width=0.19\linewidth]{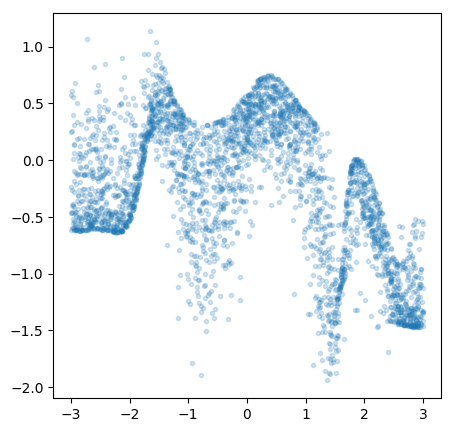}
 }
 \subfloat[][\GLGG]{
  \includegraphics[width=0.19\linewidth]{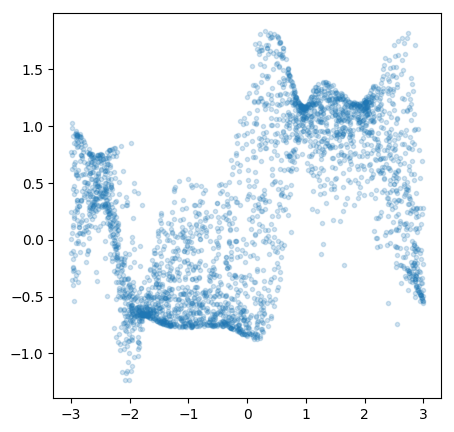}
 }
 \subfloat[][\GLGG]{
  \includegraphics[width=0.19\linewidth]{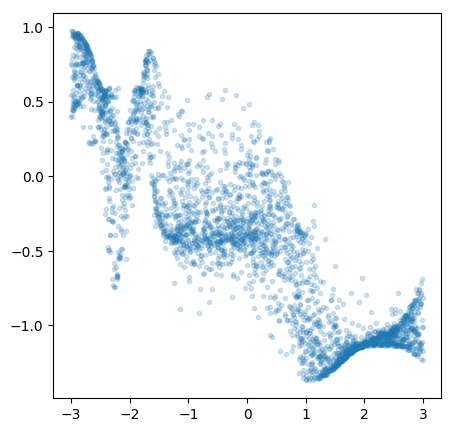}
 }\\
 \subfloat[][\LGGG]{
  \includegraphics[width=0.19\linewidth]{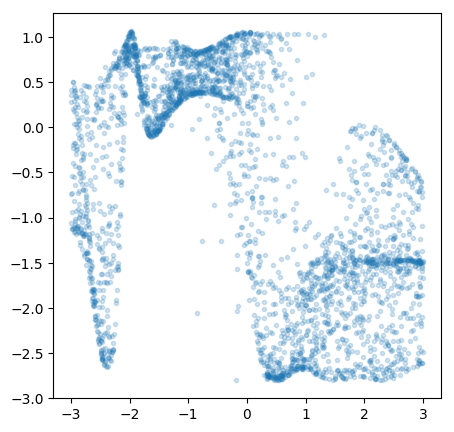}
 }
 \subfloat[][\LGGG]{
  \includegraphics[width=0.19\linewidth]{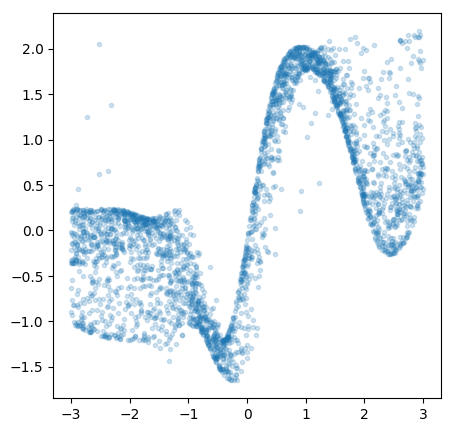}
 }
 \subfloat[][\LGGG]{
  \includegraphics[width=0.19\linewidth]{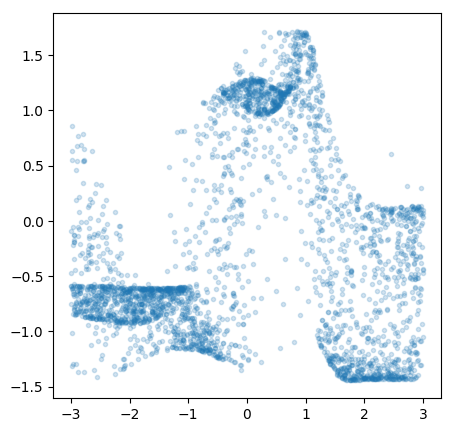}
 }
 \subfloat[][\LGGG]{
  \includegraphics[width=0.19\linewidth]{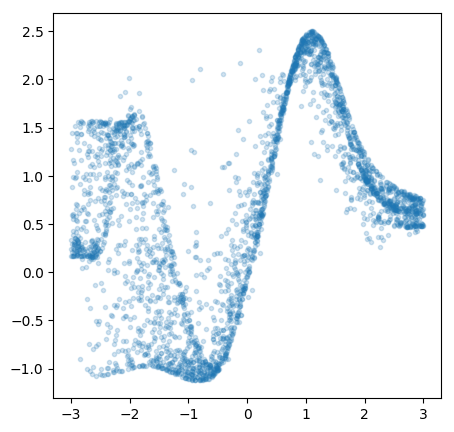}
 }
 \subfloat[][\LGGG]{
  \includegraphics[width=0.19\linewidth]{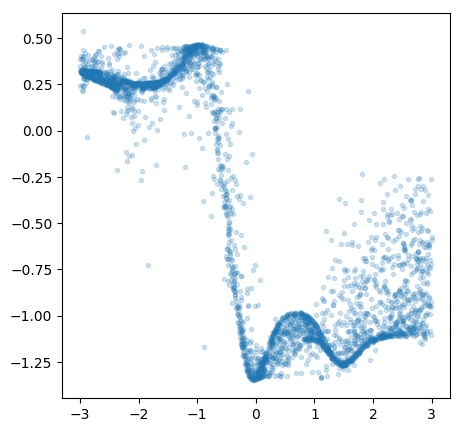}
 }
%  \subfloat[][\GG]{
%   \includegraphics[width=0.25\linewidth]{figs/priors/prior_G1_VI.png}
%  }
% \subfloat[][\LG]{
%   \includegraphics[width=0.25\linewidth]{figs/priors/prior_L1_IWAE.png}
%  }
%  \subfloat[][\LGG]{
%   \includegraphics[width=0.25\linewidth]{figs/priors/prior_L1_G1_IWAE.png}
%  }
\caption{Prior draws from \LGGG, \GLGG and \GGLG models}
\label{fig:three_layer_latent_vars}
\end{figure*}

\clearpage

\begin{landscape}
\begin{table}[]
    \centering
    \scalebox{0.61}{
    \begin{tabular}{llllllllllllllllll}
% \toprule
% {} &        N &    D &        Linear &                         CVAE 50 &                       CVAE $100-100$ &                     G &                     GG &                             GGG &                              LG &                         LG (IW) &                             LGG &                        LGG (IW) &                            LGGG &                       LGGG (IW) \\
% \midrule
\toprule
\multicolumn{3}{l}{}   
& \multicolumn{1}{c}{Linear}
& \multicolumn{1}{c}{CVAE 50}
& \multicolumn{1}{c}{CVAE $100,100$}
& \multicolumn{1}{c}{GP} 
&  \multicolumn{1}{c}{GP-GP}
& \multicolumn{1}{c}{GP-GP-GP} 
& \multicolumn{1}{c}{GP} 
&  \multicolumn{1}{c}{GP-GP}
& \multicolumn{1}{c}{GP-GP-GP} 
&   \multicolumn{2}{c}{LV-GP}
&   \multicolumn{2}{c}{LV-GP-GP} & 
 \multicolumn{2}{c}{LV-GP-GP-GP}  
\\
% \multicolumn{3}{l}{Dataset} 
Dataset & N & D 
& \multicolumn{1}{c}{VI}
& \multicolumn{1}{c}{VI}
& \multicolumn{1}{c}{VI}
& \multicolumn{1}{c}{VI} & \multicolumn{1}{c}{VI}  & \multicolumn{1}{c}{VI}  &
\multicolumn{1}{c}{SGHMC} & \multicolumn{1}{c}{SGHMC}  & \multicolumn{1}{c}{SGHMC}  &
\multicolumn{1}{c}{VI} &  \multicolumn{1}{c}{IWVI} &
\multicolumn{1}{c}{VI} &  \multicolumn{1}{c}{IWVI} &
\multicolumn{1}{c}{VI} &  \multicolumn{1}{c}{IWVI} 
\\
\midrule
challenger                                                &       23 &    4 &           -1.12 (0.02) &  \textit{\textbf{-0.59 (0.03)}} &                       $-\infty$ &          -3.22 (0.07) &  \textit{-2.04 (0.02)} &  \textit{-2.10 (0.03)} &             $-\infty$ &              $-\infty$ &    \textit{-654 (337)} &  \textit{-0.67 (0.01)} &           \textit{-0.68 (0.00)} &           \textit{-9.99 (3.60)} &           \textit{-9.47 (2.19)} &                    -1.21 (0.01) &           \textit{-2.45 (0.04)} \\
fertility                                                 &      100 &    9 &  \textbf{-1.32 (0.01)} &                       $-\infty$ &                       $-\infty$ &          -2.51 (0.13) &           -1.43 (0.01) &           -1.40 (0.01) &          -2.19 (0.02) &  \textit{-1.67 (0.03)} &  \textit{-1.35 (0.02)} &           -1.41 (0.01) &                    -1.40 (0.01) &                    -1.42 (0.01) &                    -1.69 (0.01) &           \textbf{-1.32 (0.00)} &           \textbf{-1.31 (0.01)} \\
concreteslump                                             &      103 &    7 &            0.71 (0.01) &                       $-\infty$ &                       $-\infty$ &           1.91 (0.01) &            1.55 (0.00) &            1.45 (0.00) &           0.59 (0.04) &            0.10 (0.03) &           -0.14 (0.03) &            1.93 (0.00) &            \textbf{1.94 (0.00)} &                     1.53 (0.00) &                     1.75 (0.00) &                     1.43 (0.00) &                     1.62 (0.00) \\
autos                                                     &      159 &   25 &           -0.43 (0.00) &                       $-\infty$ &                       $-\infty$ &          -0.36 (0.00) &           -0.14 (0.01) &           -0.14 (0.03) &               -10 (1) &                -15 (2) &       \textit{-23 (4)} &           -0.36 (0.01) &                    -0.35 (0.00) &           \textbf{-0.06 (0.04)} &                    -0.34 (0.01) &                    -0.25 (0.05) &                    -0.33 (0.02) \\
servo                                                     &      167 &    4 &           -0.86 (0.00) &             \textit{-385 (220)} &                       $-\infty$ &          -0.17 (0.00) &           -0.19 (0.01) &           -0.17 (0.01) &          -1.22 (0.21) &               -27 (10) &              -326 (44) &           -0.07 (0.01) &                    -0.08 (0.01) &                    -0.19 (0.01) &                    -0.07 (0.01) &                    -0.25 (0.01) &            \textbf{0.03 (0.03)} \\
breastcancer                                              &      194 &   33 &           -1.49 (0.01) &                       $-\infty$ &                       $-\infty$ &          -1.32 (0.00) &           -1.36 (0.00) &           -1.34 (0.00) &               -10 (1) &      \textit{-28 (10)} &    \textit{-462 (100)} &  \textbf{-1.31 (0.00)} &           \textbf{-1.31 (0.00)} &                    -1.43 (0.01) &                    -1.59 (0.10) &                    -1.38 (0.00) &                    -1.80 (0.19) \\
machine                                                   &      209 &    7 &           -0.71 (0.02) &              \textit{-114 (78)} &                       $-\infty$ &          -0.70 (0.01) &           -0.65 (0.02) &           -0.61 (0.01) &          -2.21 (0.28) &                -14 (4) &               -74 (13) &           -0.75 (0.02) &                    -0.71 (0.02) &                    -0.63 (0.01) &           \textbf{-0.51 (0.01)} &                    -0.59 (0.01) &           \textbf{-0.52 (0.03)} \\
yacht                                                     &      308 &    6 &           -0.14 (0.10) &                    -7.91 (4.34) &                       $-\infty$ &           1.32 (0.02) &            1.84 (0.01) &   \textbf{2.00 (0.06)} &          -0.26 (0.79) &           -9.70 (1.35) &                -16 (2) &            1.64 (0.00) &                     1.65 (0.00) &                     1.69 (0.09) &                     1.71 (0.01) &                     1.77 (0.07) &            \textbf{2.05 (0.02)} \\
autompg                                                   &      392 &    7 &           -0.48 (0.00) &                \textit{-71 (9)} &                       $-\infty$ &          -0.44 (0.01) &           -0.57 (0.02) &           -0.48 (0.03) &          -2.35 (0.22) &                -43 (4) &              -131 (26) &           -0.32 (0.01) &                    -0.33 (0.01) &                    -0.38 (0.04) &           \textbf{-0.24 (0.01)} &                    -0.65 (0.11) &                    -0.28 (0.03) \\
boston                                                    &      506 &   13 &           -0.56 (0.00) &                       $-\infty$ &                       $-\infty$ &           0.02 (0.00) &            0.07 (0.00) &            0.03 (0.01) &           0.14 (0.00) &           -2.70 (0.70) &                -21 (5) &           -0.04 (0.00) &                    -0.07 (0.00) &                     0.06 (0.00) &                     0.14 (0.00) &                    -0.04 (0.01) &            \textbf{0.17 (0.00)} \\
{\color{myAcolor} \textbf{forest}\myAcolormarker}         &      517 &   12 &           -1.36 (0.00) &               \textit{-47 (15)} &                       $-\infty$ &          -1.38 (0.00) &           -1.37 (0.00) &           -1.37 (0.00) &          -1.75 (0.02) &           -3.24 (0.48) &                -43 (5) &  \textit{-0.99 (0.00)} &           \textit{-1.00 (0.00)} &  \textit{\textbf{-0.91 (0.00)}} &           \textit{-1.02 (0.02)} &  \textit{\textbf{-0.92 (0.01)}} &  \textit{\textbf{-0.91 (0.01)}} \\
stock                                                     &      536 &   11 &  \textbf{-0.16 (0.00)} &                       $-\infty$ &                       $-\infty$ &          -0.22 (0.00) &           -0.29 (0.00) &           -0.26 (0.00) &          -0.22 (0.01) &           -2.46 (0.41) &               -97 (13) &           -0.22 (0.00) &                    -0.22 (0.00) &                    -0.29 (0.00) &                    -0.19 (0.00) &                    -0.23 (0.01) &                    -0.36 (0.02) \\
pendulum                                                  &      630 &    9 &           -1.16 (0.00) &                       $-\infty$ &                       $-\infty$ &          -0.14 (0.00) &            0.22 (0.01) &            0.12 (0.08) &          -0.11 (0.01) &   \textbf{0.33 (0.06)} &           -5.43 (1.52) &           -0.14 (0.00) &                    -0.14 (0.00) &                     0.22 (0.01) &            \textbf{0.33 (0.08)} &            \textbf{0.25 (0.00)} &                     0.20 (0.03) \\
energy                                                    &      768 &    8 &           -0.06 (0.00) &                    -0.44 (0.31) &                       $-\infty$ &           1.71 (0.00) &            1.85 (0.00) &            2.07 (0.01) &           1.75 (0.00) &            1.97 (0.01) &            1.12 (0.33) &            1.86 (0.01) &                     1.92 (0.01) &                     2.00 (0.04) &                     1.96 (0.01) &                     2.07 (0.01) &            \textbf{2.28 (0.01)} \\
concrete                                                  &     1030 &    8 &           -0.97 (0.00) &                        -26 (10) &                       $-\infty$ &          -0.43 (0.00) &           -0.45 (0.00) &           -0.35 (0.02) &          -0.33 (0.01) &           -0.88 (0.38) &           -1.85 (1.01) &           -0.32 (0.00) &                    -0.32 (0.00) &                    -0.35 (0.00) &                    -0.21 (0.00) &                    -0.22 (0.01) &           \textbf{-0.12 (0.00)} \\
{\color{myAcolor} \textbf{solar}\myAcolormarker}          &     1066 &   10 &           -1.54 (0.03) &   \textit{\textbf{0.71 (0.02)}} &            \textit{0.26 (0.34)} &          -1.75 (0.08) &           -1.21 (0.02) &           -1.20 (0.03) &          -1.56 (0.05) &           -1.58 (0.08) &           -1.65 (0.06) &   \textit{0.04 (0.07)} &            \textit{0.07 (0.01)} &            \textit{0.54 (0.01)} &            \textit{0.22 (0.01)} &            \textit{0.54 (0.01)} &            \textit{0.20 (0.02)} \\
airfoil                                                   &     1503 &    5 &           -1.11 (0.00) &                    -0.64 (0.32) &                       $-\infty$ &          -0.79 (0.05) &            0.08 (0.02) &            0.14 (0.03) &          -0.90 (0.05) &           -0.19 (0.20) &   \textbf{0.32 (0.09)} &           -0.44 (0.03) &                    -0.36 (0.01) &                     0.07 (0.00) &                     0.30 (0.00) &                    -0.02 (0.03) &            \textbf{0.34 (0.01)} \\
winered                                                   &     1599 &   11 &           -1.14 (0.00) &           \textit{-9.64 (1.40)} &           \textit{-1.86 (0.85)} &          -1.09 (0.00) &           -1.11 (0.00) &           -1.08 (0.00) &          -1.11 (0.00) &           -1.09 (0.01) &           -1.63 (0.09) &           -1.07 (0.00) &                    -1.06 (0.00) &                    -1.10 (0.00) &  \textit{\textbf{-0.84 (0.01)}} &           \textit{-1.06 (0.03)} &           \textit{-1.31 (0.40)} \\
gas                                                       &     2565 &  128 &            0.23 (0.00) &                       -241 (77) &                       $-\infty$ &           1.07 (0.00) &   \textbf{1.60 (0.05)} &   \textbf{1.69 (0.05)} &           0.93 (0.01) &            1.20 (0.02) &            1.46 (0.05) &   \textbf{1.69 (0.08)} &                     1.56 (0.06) &                     0.70 (0.10) &            \textbf{1.57 (0.16)} &                     1.30 (0.14) &            \textbf{1.61 (0.12)} \\
skillcraft                                                &     3338 &   19 &           -0.95 (0.00) &                    -5.80 (0.29) &                       $-\infty$ &          -0.94 (0.00) &           -0.94 (0.00) &           -0.94 (0.00) &          -0.95 (0.00) &           -0.99 (0.02) &           -1.10 (0.03) &  \textbf{-0.91 (0.00)} &           \textbf{-0.91 (0.00)} &                    -0.92 (0.00) &                    -0.93 (0.00) &                    -0.92 (0.00) &                    -0.94 (0.00) \\
{\color{myBcolor} \textbf{sml}\myBcolormarker}            &     4137 &   26 &            0.32 (0.00) &                    -3.79 (0.87) &                       $-\infty$ &           1.53 (0.00) &            1.72 (0.01) &            1.83 (0.01) &           1.46 (0.00) &            1.71 (0.04) &   \textbf{1.93 (0.01)} &            1.52 (0.00) &                     1.52 (0.00) &                     1.79 (0.00) &                     1.91 (0.00) &            \textbf{1.92 (0.01)} &            \textbf{1.97 (0.05)} \\
winewhite                                                 &     4898 &   11 &           -1.22 (0.00) &                    -1.37 (0.04) &                       $-\infty$ &          -1.14 (0.00) &           -1.14 (0.00) &           -1.14 (0.00) &          -1.13 (0.00) &           -1.13 (0.00) &           -1.14 (0.01) &           -1.13 (0.00) &                    -1.13 (0.00) &                    -1.13 (0.00) &           \textbf{-1.10 (0.00)} &                    -1.13 (0.00) &           \textbf{-1.09 (0.00)} \\
parkinsons                                                &     5875 &   20 &           -1.30 (0.00) &                    -1.14 (0.56) &                       $-\infty$ &           1.99 (0.00) &            2.61 (0.01) &            2.75 (0.02) &           1.87 (0.02) &            2.38 (0.05) &            2.41 (0.04) &            1.79 (0.02) &                     1.82 (0.02) &                     2.36 (0.02) &                     2.76 (0.02) &                     2.71 (0.07) &            \textbf{3.12 (0.05)} \\
{\color{myBcolor} \textbf{kin8nm}\myBcolormarker}         &     8192 &    8 &           -1.12 (0.00) &                    -0.01 (0.01) &                       $-\infty$ &          -0.29 (0.00) &           -0.01 (0.00) &           -0.00 (0.00) &          -0.30 (0.00) &           -0.03 (0.01) &           -0.00 (0.00) &           -0.29 (0.00) &                    -0.29 (0.00) &                    -0.02 (0.00) &                    -0.00 (0.00) &                    -0.00 (0.00) &            \textbf{0.03 (0.00)} \\
pumadyn32nm                                               &     8192 &   32 &           -1.44 (0.00) &                    -2.10 (0.09) &                       $-\infty$ &           0.08 (0.00) &            0.11 (0.01) &            0.11 (0.00) &  \textbf{0.11 (0.00)} &   \textbf{0.11 (0.00)} &            0.10 (0.00) &           -1.44 (0.00) &                    -1.44 (0.00) &                     0.10 (0.01) &           \textit{-0.62 (0.25)} &                     0.11 (0.00) &                     0.11 (0.00) \\
{\color{myAcolor} \textbf{power}\myAcolormarker}          &     9568 &    4 &           -0.43 (0.01) &                    -0.44 (0.08) &                    -2.14 (0.06) &          -0.65 (0.04) &           -0.75 (0.03) &           -0.80 (0.02) &          -0.58 (0.03) &           -1.41 (0.00) &           -1.41 (0.00) &           -0.39 (0.04) &                    -0.23 (0.02) &                    -0.36 (0.04) &           \textit{-0.28 (0.05)} &                    -0.25 (0.06) &  \textit{\textbf{-0.11 (0.06)}} \\
naval                                                     &    11934 &   14 &           -0.63 (0.00) &                     3.07 (0.08) &                     3.39 (0.12) &  \textbf{4.52 (0.02)} &            4.43 (0.03) &            4.35 (0.03) &           3.17 (0.03) &           -1.41 (0.00) &           -1.41 (0.00) &   \textit{4.19 (0.01)} &            \textit{4.24 (0.01)} &                     4.36 (0.01) &            \textbf{4.52 (0.02)} &                     4.27 (0.02) &                     4.41 (0.02) \\
{\color{myCcolor} \textbf{pol}\myCcolormarker}            &    15000 &   26 &           -1.10 (0.00) &                    -0.05 (0.08) &                       -411 (19) &           0.48 (0.00) &            1.51 (0.01) &            1.45 (0.01) &           0.46 (0.00) &            1.32 (0.03) &            1.31 (0.07) &   \textit{0.37 (0.08)} &           \textit{-0.50 (0.00)} &            \textit{2.34 (0.02)} &   \textit{\textbf{2.63 (0.01)}} &                     1.47 (0.01) &            \textbf{2.72 (0.10)} \\
elevators                                                 &    16599 &   18 &           -0.74 (0.00) &                    -0.28 (0.00) &                         -33 (1) &          -0.44 (0.00) &           -0.41 (0.01) &           -0.40 (0.00) &          -0.46 (0.00) &           -0.40 (0.01) &           -0.39 (0.01) &           -0.37 (0.00) &                    -0.36 (0.00) &                    -0.29 (0.00) &                    -0.27 (0.00) &                    -0.28 (0.00) &           \textbf{-0.27 (0.00)} \\
{\color{myCcolor} \textbf{bike}\myCcolormarker}           &    17379 &   17 &           -0.76 (0.00) &            \textit{3.90 (0.01)} &            \textbf{4.47 (0.02)} &           0.82 (0.01) &            3.49 (0.01) &            3.68 (0.01) &           1.06 (0.01) &            1.21 (0.95) &            2.77 (0.10) &   \textit{2.48 (0.01)} &            \textit{2.66 (0.01)} &                     3.48 (0.00) &                     3.75 (0.01) &                     3.72 (0.01) &                     3.95 (0.01) \\
{\color{myBcolor} \textbf{kin40k}\myBcolormarker}         &    40000 &    8 &           -1.43 (0.00) &                     0.65 (0.01) &            \textbf{1.24 (0.13)} &           0.02 (0.00) &            0.84 (0.00) &            1.17 (0.00) &          -0.02 (0.00) &           -1.43 (0.00) &           -1.43 (0.00) &            0.04 (0.00) &                     0.05 (0.00) &                     0.87 (0.01) &                     0.93 (0.00) &                     1.15 (0.00) &            \textbf{1.27 (0.00)} \\
protein                                                   &    45730 &    9 &           -1.25 (0.00) &           \textit{-0.78 (0.02)} &           \textit{-0.57 (0.00)} &          -1.06 (0.00) &           -0.98 (0.00) &           -0.95 (0.00) &          -1.08 (0.00) &           -1.10 (0.07) &           -0.95 (0.00) &  \textit{-0.85 (0.00)} &           \textit{-0.80 (0.00)} &           \textit{-0.70 (0.00)} &           \textit{-0.61 (0.00)} &           \textit{-0.68 (0.00)} &  \textit{\textbf{-0.57 (0.00)}} \\
tamielectric                                              &    45781 &    3 &           -1.43 (0.00) &           \textit{-1.31 (0.00)} &  \textit{\textbf{-1.31 (0.00)}} &          -1.43 (0.00) &           -1.43 (0.00) &           -1.43 (0.00) &          -1.43 (0.00) &           -1.43 (0.00) &           -1.43 (0.00) &  \textit{-1.31 (0.00)} &           \textit{-1.31 (0.00)} &           \textit{-1.31 (0.00)} &           \textit{-1.31 (0.00)} &           \textit{-1.31 (0.00)} &           \textit{-1.31 (0.00)} \\
{\color{myCcolor} \textbf{keggdirected}\myCcolormarker}   &    48827 &   20 &           -0.34 (0.15) &            \textit{0.65 (0.32)} &            \textit{1.59 (0.14)} &           0.16 (0.01) &            0.21 (0.01) &            0.26 (0.02) &           0.26 (0.02) &            0.45 (0.02) &            0.53 (0.04) &   \textit{1.24 (0.06)} &            \textit{2.03 (0.01)} &            \textit{1.84 (0.05)} &            \textit{2.23 (0.01)} &            \textit{1.92 (0.02)} &   \textit{\textbf{2.26 (0.01)}} \\
{\color{myBcolor} \textbf{slice}\myBcolormarker}          &    53500 &  385 &           -0.47 (0.00) &                         -27 (1) &                       -358 (15) &           0.86 (0.00) &            1.80 (0.00) &            1.86 (0.00) &           1.02 (0.00) &            1.84 (0.02) &            1.93 (0.01) &            0.85 (0.00) &                     0.90 (0.00) &                     1.80 (0.00) &                     1.88 (0.00) &                     1.86 (0.00) &            \textbf{2.02 (0.00)} \\
{\color{myAcolor} \textbf{keggundirected}\myAcolormarker} &    63608 &   27 &            0.28 (0.00) &              \textit{-344 (39)} &                       $-\infty$ &           0.06 (0.01) &            0.06 (0.01) &            0.07 (0.01) &          -0.03 (0.02) &           -0.14 (0.23) &            0.10 (0.01) &       \textit{-75 (4)} &           \textit{-4.21 (0.33)} &                \textit{-64 (6)} &            \textit{2.37 (0.29)} &              \textit{-116 (17)} &   \textit{\textbf{2.98 (0.21)}} \\
3droad                                                    &   434874 &    3 &           -1.41 (0.00) &           \textit{-0.85 (0.00)} &           \textit{-0.53 (0.01)} &          -0.79 (0.00) &           -0.61 (0.01) &           -0.58 (0.01) &          -1.15 (0.06) &           -1.42 (0.00) &           -1.42 (0.00) &  \textit{-0.71 (0.00)} &           \textit{-0.65 (0.00)} &           \textit{-0.69 (0.00)} &           \textit{-0.55 (0.01)} &           \textit{-0.61 (0.00)} &  \textit{\textbf{-0.48 (0.00)}} \\
song                                                      &   515345 &   90 &           -1.28 (0.00) &                    -1.11 (0.00) &                    -1.09 (0.00) &          -1.19 (0.00) &           -1.18 (0.00) &           -1.18 (0.00) &          -1.19 (0.00) &           -1.17 (0.00) &           -1.17 (0.00) &           -1.15 (0.00) &                    -1.14 (0.00) &                    -1.13 (0.00) &           \textit{-1.10 (0.00)} &                    -1.12 (0.00) &  \textit{\textbf{-1.07 (0.00)}} \\
buzz                                                      &   583250 &   77 &           -1.56 (0.04) &                     0.01 (0.00) &            \textbf{0.07 (0.00)} &          -0.24 (0.00) &           -0.15 (0.00) &           -0.15 (0.00) &          -0.92 (0.02) &           -0.48 (0.01) &           -0.43 (0.03) &           -0.22 (0.01) &                    -0.23 (0.01) &                    -0.09 (0.01) &                    -0.04 (0.03) &                    -0.09 (0.01) &            \textit{0.04 (0.00)} \\
nytaxi                                                    &  1420068 &    8 &           -1.53 (0.01) &           \textit{-1.64 (0.04)} &           \textit{-1.90 (0.04)} &          -1.42 (0.01) &           -1.44 (0.01) &           -1.42 (0.01) &          -1.46 (0.01) &           -1.78 (0.00) &           -1.77 (0.00) &  \textit{-1.09 (0.03)} &  \textit{\textbf{-0.90 (0.04)}} &           \textit{-1.57 (0.04)} &           \textit{-1.03 (0.04)} &           \textit{-1.74 (0.05)} &  \textit{\textbf{-0.95 (0.07)}} \\
{\color{myCcolor} \textbf{houseelectric}\myCcolormarker}  &  2049280 &   11 &           -0.52 (0.00) &                     1.41 (0.00) &                     1.48 (0.01) &           1.37 (0.00) &            1.41 (0.00) &            1.47 (0.01) &           0.01 (0.04) &           -1.43 (0.00) &           -1.43 (0.00) &   \textit{1.65 (0.03)} &            \textit{1.82 (0.01)} &                     1.66 (0.05) &            \textit{2.01 (0.00)} &                     1.77 (0.06) &   \textit{\textbf{2.03 (0.00)}} \\
\midrule
\multicolumn{3}{l}{Median difference from GP baseline} &
 -0.38 &                           -2.06 &                        -6016.67 &                  0 &                   0.06 &                   0.09 &                 -0.02 &                  -0.02 &                  -0.28 &                   0.04 &                            0.05 &                            0.12 &                            0.26 &                            0.18 &                            0.32 \\
\multicolumn{3}{l}{Average ranks} &    4.99 (0.25) &                     5.32 (0.31) &                     4.35 (0.38) &           6.80 (0.21) &            8.65 (0.20) &            9.63 (0.20) &           5.57 (0.19) &            5.57 (0.24) &            5.21 (0.28) &            8.94 (0.22) &                     9.36 (0.22) &                     9.82 (0.17) &                    12.04 (0.18) &                    10.71 (0.19) &                    13.03 (0.20) \\
\bottomrule
\end{tabular}
    }
    \caption{Test log likelihood values over 5 splits (standard errors), including the CVAE models and models with Stochastic Gradient HMC \citet{havasi2018inference}. Results less than $-1000$ are reported as $-\infty$. The numbers after the CVAE models are the number of hidden units. The SGHMC results were run with identical models as the corresponding VI methods, Hyperparameter optimization was performed using a random sample from the last 100 iterations, as described in \cite{havasi2018inference}. Posterior sample were take with 2000 samples, with a thinning factor of 5 (i.e. a chain of length 10000). We included also a single layer GP using the SGHMC approach for comparison. The discrepancy between the variational approach and the SGHMC is attributable to several factors (note that the variational approach is optimal in this case): different batch sizes (512 in this work compared to 10000 in  \citet{havasi2018inference}); different hyperparameter initializations (for example, a much longer initial lengthscales is used in \citet{havasi2018inference}); different learning rate schedules (decaying learning rate in this work compared to fixed in \citet{havasi2018inference}); the use of natural gradients in this work for the final layer, as opposed to optimizing with Adam optimizer, as done in  \citet{havasi2018inference, salimbeni2017doubly}; insufficient mixing in the posterior chain. 
}
    \label{tab:all_methods}
\end{table}
\end{landscape}

\clearpage
\begin{landscape}
\begin{table}[]
    \centering
    \scalebox{0.75}{
    % \input{figs/tables/SW_full.tex}
    % \begin{tabular}{llllllllllllllllll}
\begin{tabular}{lllccccccccccccccc}
% \toprule
% {} &        N &    D &        Linear &                         CVAE 50 &                       CVAE $100-100$ &                     G &                     GG &                             GGG &                              LG &                         LG (IW) &                             LGG &                        LGG (IW) &                            LGGG &                       LGGG (IW) \\
% \midrule
\toprule
\multicolumn{3}{l}{}   
& \multicolumn{1}{c}{Linear}
& \multicolumn{1}{c}{CVAE 50}
& \multicolumn{1}{c}{CVAE $100,100$}
& \multicolumn{1}{c}{GP} 
&  \multicolumn{1}{c}{GP-GP}
& \multicolumn{1}{c}{GP-GP-GP} 
& \multicolumn{1}{c}{GP} 
&  \multicolumn{1}{c}{GP-GP}
& \multicolumn{1}{c}{GP-GP-GP} 
&   \multicolumn{2}{c}{LV-GP}
&   \multicolumn{2}{c}{LV-GP-GP} & 
 \multicolumn{2}{c}{LV-GP-GP-GP}  
\\
% \multicolumn{3}{l}{Dataset} 
Dataset & N & D 
& \multicolumn{1}{c}{VI}
& \multicolumn{1}{c}{VI}
& \multicolumn{1}{c}{VI}
& \multicolumn{1}{c}{VI} & \multicolumn{1}{c}{VI}  & \multicolumn{1}{c}{VI}  &
\multicolumn{1}{c}{SGHMC} & \multicolumn{1}{c}{SGHMC}  & \multicolumn{1}{c}{SGHMC}  &
\multicolumn{1}{c}{VI} &  \multicolumn{1}{c}{IWVI} &
\multicolumn{1}{c}{VI} &  \multicolumn{1}{c}{IWVI} &
\multicolumn{1}{c}{VI} &  \multicolumn{1}{c}{IWVI} 
\\
\midrule% \begin{tabular}{lllllllllllllll}
challenger                                                &       23 &    4 &               &       0.5191 &                &        0.9990 &        0.8870 &        0.7739 &               &       0.9823 &       0.9099 &       0.7001 &       0.7115 &       0.8466 &       0.8484 &              &       0.7187 \\
fertility                                                 &      100 &    9 &               &       0.9408 &         0.8353 &               &               &               &               &       0.9598 &       0.9341 &              &              &              &       0.9928 &              &       0.9969 \\
concreteslump                                             &      103 &    7 &               &              &                &               &               &               &               &       0.9970 &       0.9922 &              &              &              &              &              &              \\
autos                                                     &      159 &   25 &               &       0.9823 &         0.8798 &               &               &               &               &       0.9903 &       0.9874 &              &              &              &       0.9990 &              &       0.9990 \\
servo                                                     &      167 &    4 &               &       0.8358 &                &               &               &               &               &       0.9989 &       0.9975 &       0.9990 &       0.9990 &              &              &              &       0.9988 \\
breastcancer                                              &      194 &   33 &               &              &                &               &               &               &               &       0.9787 &       0.9814 &              &              &              &       0.9911 &              &       0.9943 \\
machine                                                   &      209 &    7 &               &       0.9181 &         0.8967 &               &               &               &               &       0.9974 &       0.9989 &              &              &              &              &              &              \\
yacht                                                     &      308 &    6 &               &       0.9987 &                &               &               &               &               &              &              &              &              &              &       0.9989 &              &       0.9990 \\
autompg                                                   &      392 &    7 &               &       0.9097 &         0.9948 &               &               &               &               &       0.9989 &       0.9987 &       0.9990 &       0.9988 &       0.9987 &       0.9985 &              &       0.9988 \\
boston                                                    &      506 &   13 &               &       0.9916 &                &               &               &               &               &              &       0.9989 &       0.9989 &       0.9979 &       0.9989 &       0.9980 &              &       0.9981 \\
{\color{myAcolor} \textbf{forest}\myAcolormarker}         &      517 &   12 &               &       0.8453 &         0.7145 &               &               &               &               &       0.9981 &       0.9964 &       0.7034 &       0.7074 &       0.7470 &       0.7398 &       0.7608 &       0.7642 \\
stock                                                     &      536 &   11 &               &              &                &               &               &               &               &              &              &              &              &              &              &              &              \\
pendulum                                                  &      630 &    9 &               &              &                &               &               &               &               &              &              &              &              &              &       0.9989 &              &       0.9988 \\
energy                                                    &      768 &    8 &               &              &                &               &               &               &               &              &              &              &       0.9987 &              &              &              &       0.9990 \\
concrete                                                  &     1030 &    8 &               &       0.9986 &         0.9782 &               &               &               &               &              &              &              &              &              &       0.9990 &              &       0.9989 \\
{\color{myAcolor} \textbf{solar}\myAcolormarker}          &     1066 &   10 &               &       0.3028 &         0.3228 &               &               &               &               &              &              &       0.3380 &       0.3509 &       0.2728 &       0.2945 &       0.2557 &       0.2883 \\
airfoil                                                   &     1503 &    5 &               &       0.9989 &         0.9986 &               &               &               &               &              &              &              &              &              &       0.9988 &              &       0.9988 \\
winered                                                   &     1599 &   11 &               &       0.9790 &         0.6980 &               &               &               &               &              &              &              &              &              &       0.9234 &       0.9751 &       0.8473 \\
gas                                                       &     2565 &  128 &               &       0.9984 &                &               &               &               &               &              &              &       0.9983 &       0.9935 &       0.9989 &       0.9983 &       0.9990 &       0.9986 \\
skillcraft                                                &     3338 &   19 &               &              &                &               &               &               &               &              &              &       0.9989 &       0.9983 &       0.9989 &       0.9988 &       0.9990 &       0.9989 \\
{\color{myBcolor} \textbf{sml}\myBcolormarker}            &     4137 &   26 &               &              &                &               &               &               &               &              &              &              &              &              &              &              &              \\
winewhite                                                 &     4898 &   11 &               &              &                &               &               &               &               &              &              &              &              &              &       0.9990 &              &       0.9990 \\
parkinsons                                                &     5875 &   20 &               &       0.9989 &                &               &               &               &               &              &              &              &              &              &              &              &              \\
{\color{myBcolor} \textbf{kin8nm}\myBcolormarker}         &     8192 &    8 &               &              &                &               &               &               &               &              &              &              &              &              &              &              &              \\
pumadyn32nm                                               &     8192 &   32 &               &              &                &               &               &               &               &              &              &              &              &              &       0.9770 &              &              \\
{\color{myAcolor} \textbf{power}\myAcolormarker}          &     9568 &    4 &               &       0.9964 &                &               &               &               &               &              &              &       0.9963 &       0.9926 &       0.9964 &       0.9852 &       0.9963 &       0.9855 \\
naval                                                     &    11934 &   14 &               &              &                &               &               &               &               &              &              &       0.9649 &       0.9332 &              &              &              &              \\
{\color{myCcolor} \textbf{pol}\myCcolormarker}            &    15000 &   26 &               &       0.9970 &                &               &               &               &               &              &              &       0.8482 &       0.7214 &       0.9839 &       0.9508 &              &       0.9980 \\
elevators                                                 &    16599 &   18 &               &       0.9979 &                &               &               &               &               &              &              &       0.9986 &       0.9977 &       0.9987 &       0.9976 &       0.9988 &       0.9979 \\
{\color{myCcolor} \textbf{bike}\myCcolormarker}           &    17379 &   17 &               &       0.9815 &                &               &               &               &               &              &              &       0.9657 &       0.9337 &       0.9990 &       0.9990 &              &              \\
{\color{myBcolor} \textbf{kin40k}\myBcolormarker}         &    40000 &    8 &               &       0.9990 &                &               &               &               &               &              &              &              &              &              &              &              &              \\
protein                                                   &    45730 &    9 &               &       0.8839 &         0.8800 &               &               &               &               &              &              &       0.9117 &       0.9041 &       0.8904 &       0.8761 &       0.8840 &       0.8691 \\
tamielectric                                              &    45781 &    3 &               &       0.9551 &         0.9502 &               &               &               &               &              &              &       0.9558 &       0.9562 &       0.9550 &       0.9549 &       0.9551 &       0.9549 \\
{\color{myCcolor} \textbf{keggdirected}\myCcolormarker}   &    48827 &   20 &               &       0.8279 &         0.5350 &               &               &               &               &              &              &       0.7498 &       0.5445 &       0.6839 &       0.5187 &       0.7039 &       0.5001 \\
{\color{myBcolor} \textbf{slice}\myBcolormarker}          &    53500 &  385 &               &              &                &               &               &               &               &              &              &       0.9990 &       0.9983 &              &       0.9988 &              &       0.9988 \\
{\color{myAcolor} \textbf{keggundirected}\myAcolormarker} &    63608 &   27 &               &       0.4804 &         0.9882 &               &               &               &               &              &              &       0.5287 &       0.3281 &       0.5511 &       0.2049 &       0.9295 &       0.2322 \\
3droad                                                    &   434874 &    3 &               &       0.9421 &         0.9479 &               &               &               &               &              &              &       0.9432 &       0.9271 &       0.9650 &       0.9467 &       0.9744 &       0.9464 \\
song                                                      &   515345 &   90 &               &       0.9936 &         0.9957 &               &               &               &               &              &              &       0.9983 &       0.9932 &       0.9969 &       0.9893 &       0.9963 &       0.9835 \\
buzz                                                      &   583250 &   77 &               &       0.9952 &         0.9909 &               &               &               &               &              &              &              &              &       0.9988 &       0.9938 &       0.9984 &       0.9838 \\
nytaxi                                                    &  1420068 &    8 &               &       0.9608 &         0.9422 &               &               &               &               &              &              &       0.9698 &       0.9274 &       0.9515 &       0.9019 &       0.9467 &       0.8981 \\
{\color{myCcolor} \textbf{houseelectric}\myCcolormarker}  &  2049280 &   11 &               &              &                &               &               &               &               &              &              &       0.9830 &       0.9676 &       0.9962 &       0.9671 &       0.9933 &       0.9743 \\
\bottomrule
\end{tabular}
    }
    \caption{Median Shapiro–Wilk test statistic for the test points. Blanks have values of one (perfectly Gaussian). Smaller values indicate more evidence for non-Gaussian marginal distributions.}
    \label{tab:shapiro_wilk}
\end{table}
\end{landscape}

\begin{figure}
\begin{tabular}{lccccc}
& CVAE $50-50$ & CVAE $100-100-100$ & \LG (IW) & \LG (IW) & \LGG (IW) \\
\rotatebox{90}{\hspace{0.5cm}forest}
&\includegraphics[width=0.15\linewidth]{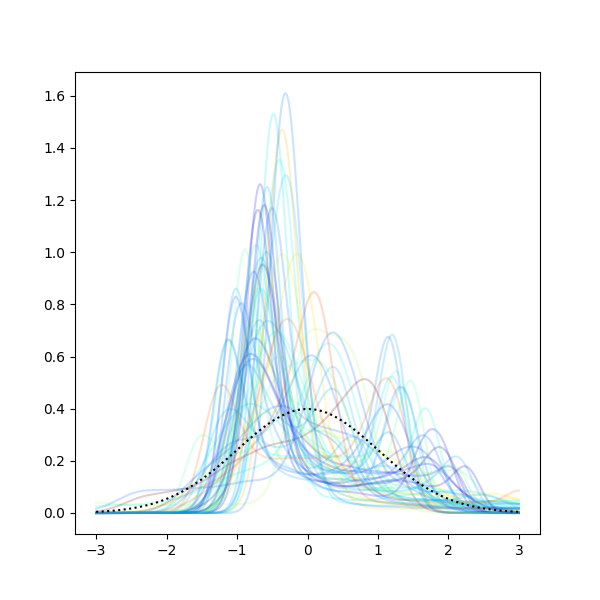}
&\includegraphics[width=0.15\linewidth]{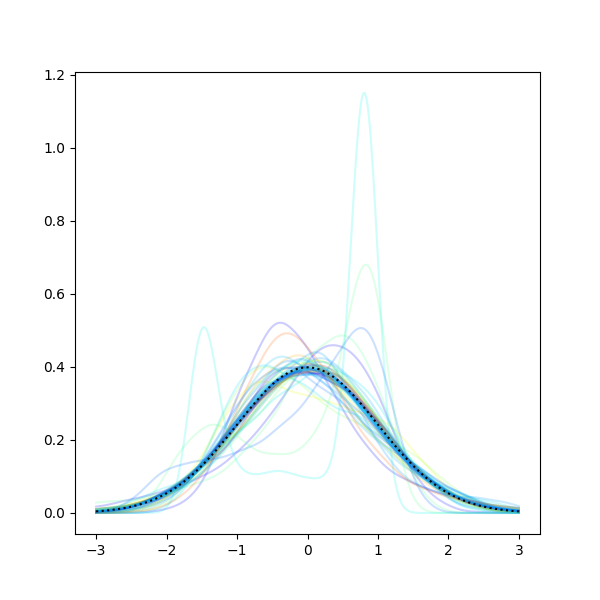}
&\includegraphics[width=0.15\linewidth]{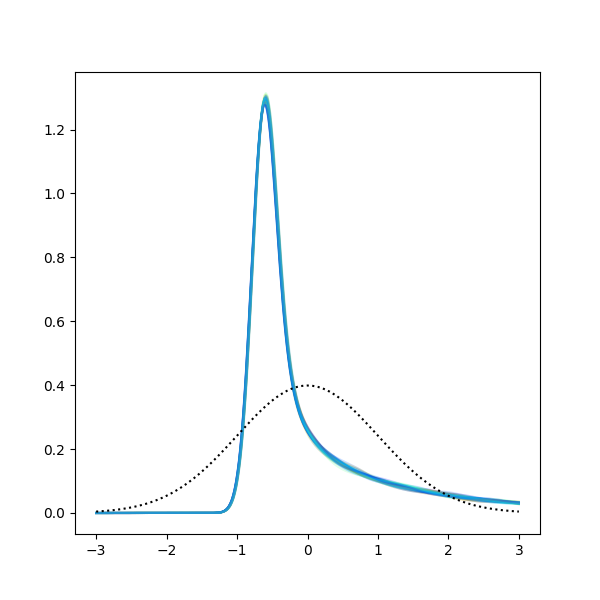}
&\includegraphics[width=0.15\linewidth]{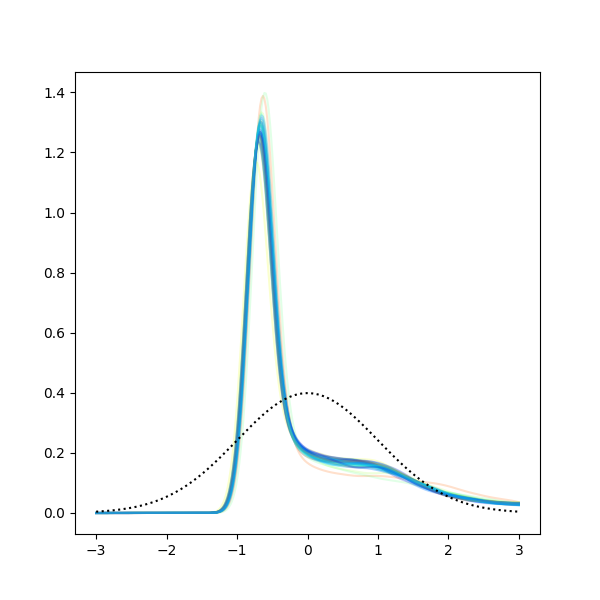}
&\includegraphics[width=0.15\linewidth]{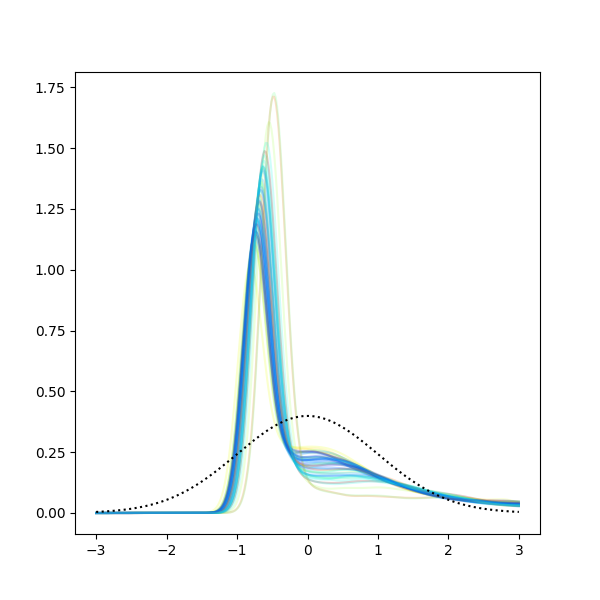}
\\
\rotatebox{90}{\hspace{0.5cm}solar}
&\includegraphics[width=0.15\linewidth]{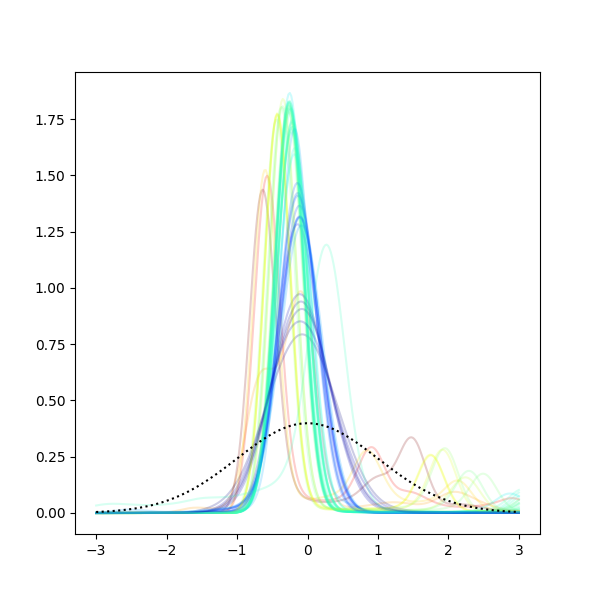}
&\includegraphics[width=0.15\linewidth]{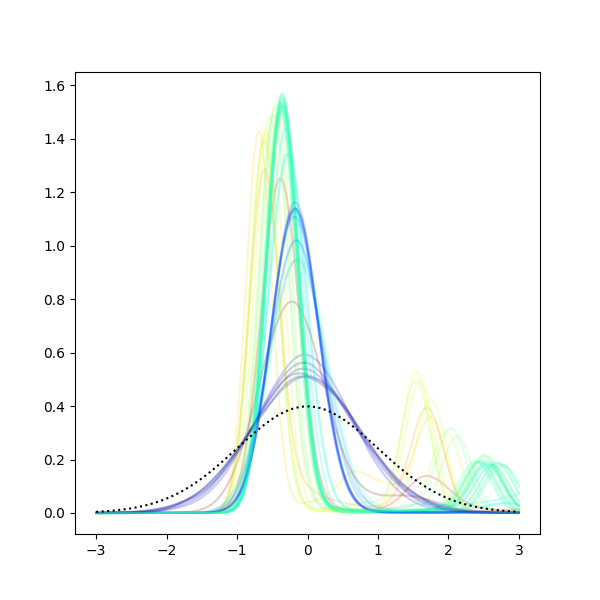}
&\includegraphics[width=0.15\linewidth]{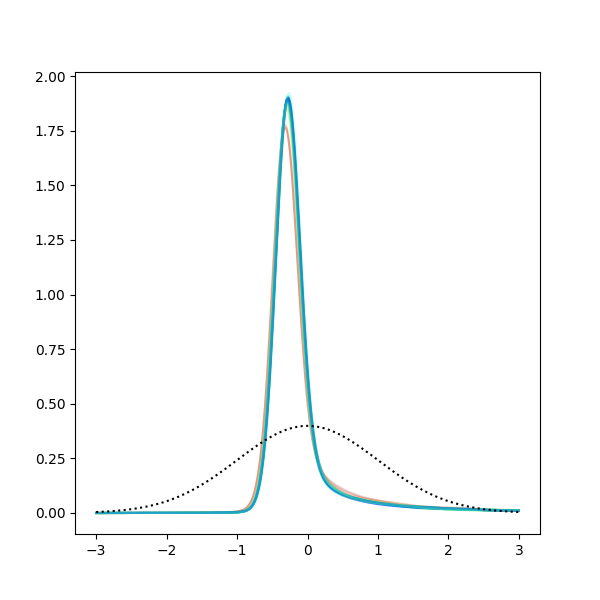}
&\includegraphics[width=0.15\linewidth]{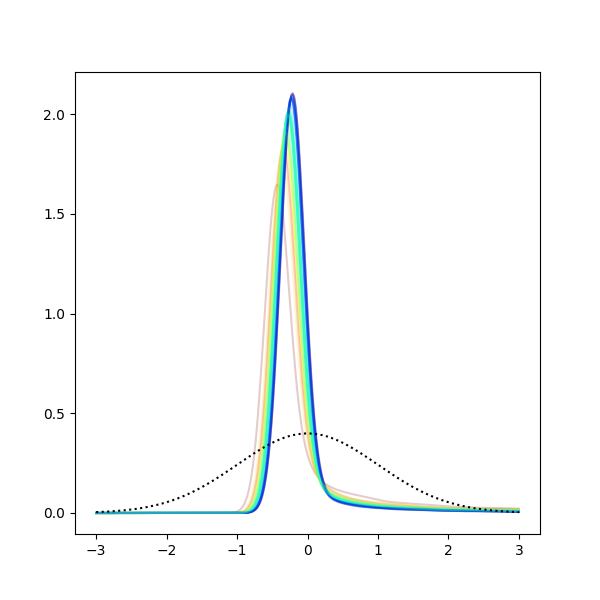}
&\includegraphics[width=0.15\linewidth]{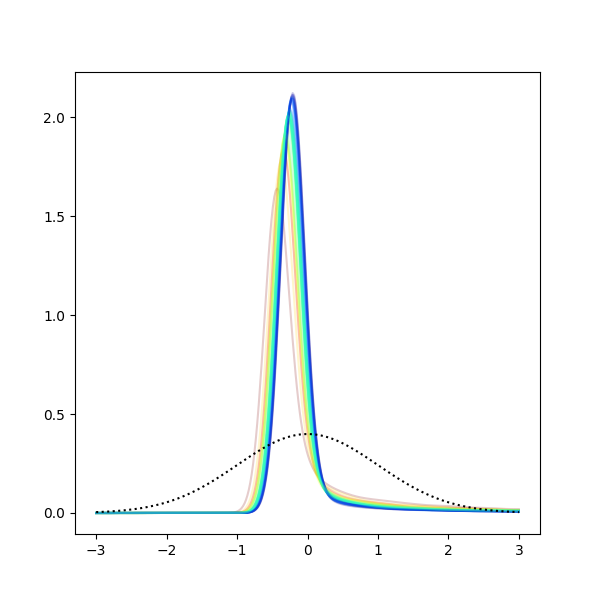}
\\
\rotatebox{90}{\hspace{0.5cm}power}
&\includegraphics[width=0.15\linewidth]{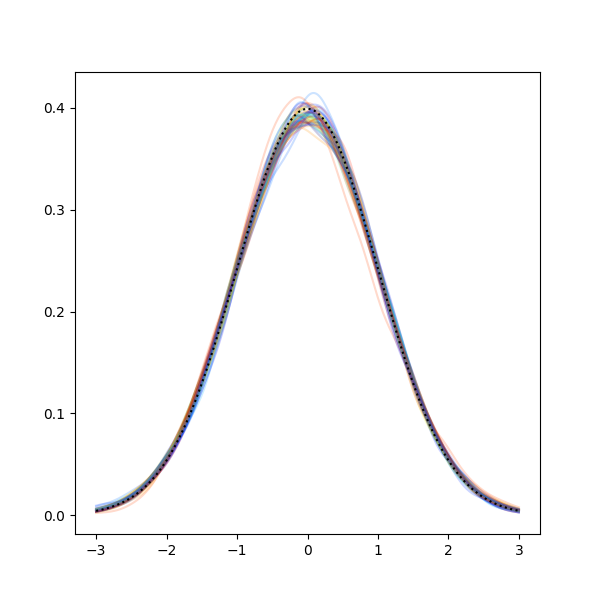}
&\includegraphics[width=0.15\linewidth]{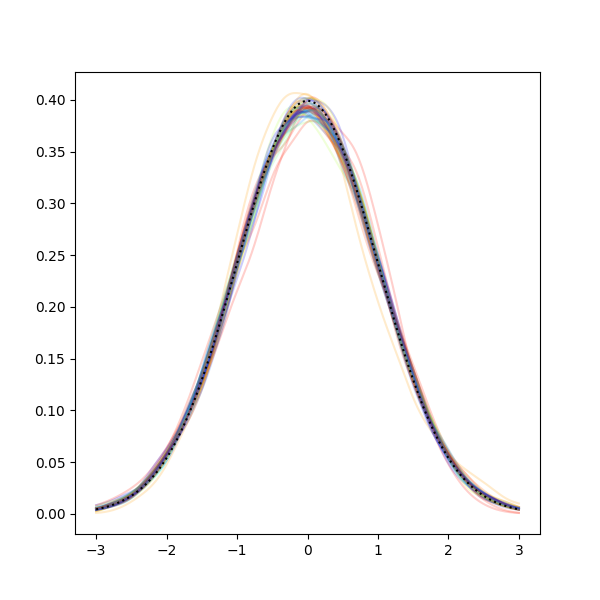}
&\includegraphics[width=0.15\linewidth]{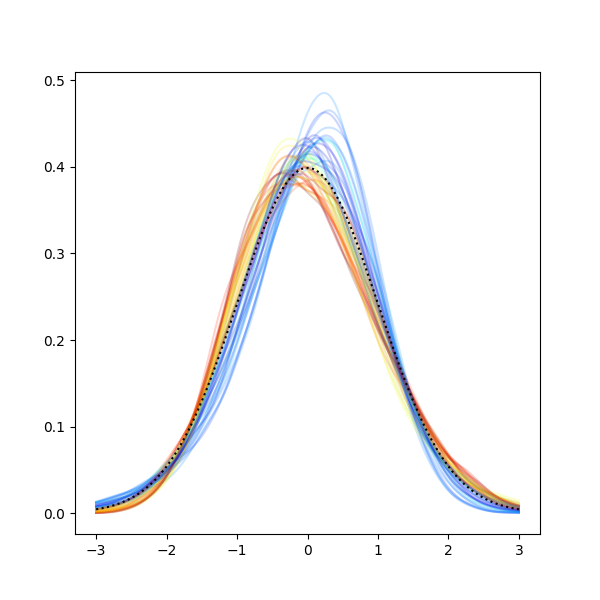}
&\includegraphics[width=0.15\linewidth]{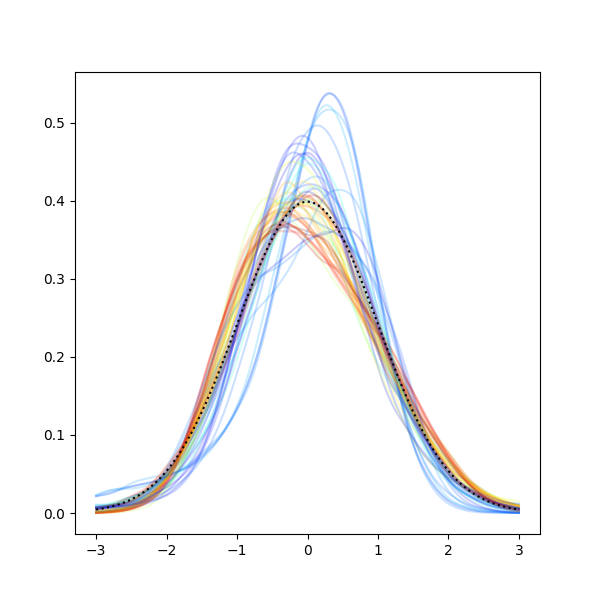}
&\includegraphics[width=0.15\linewidth]{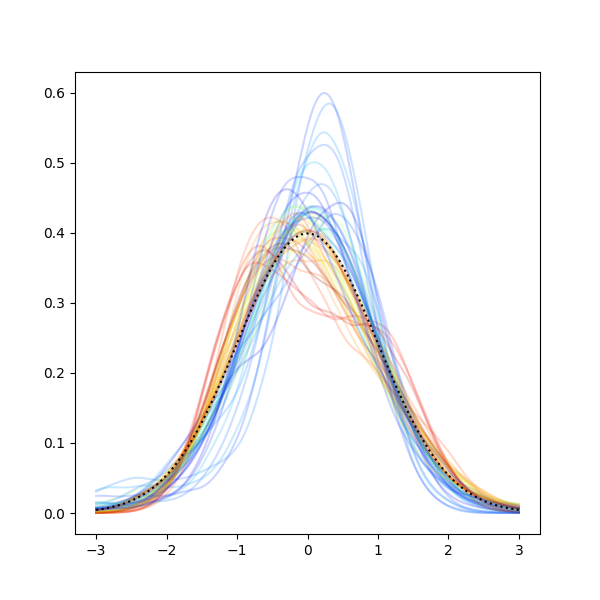}
\\
\rotatebox{90}{\hspace{0.5cm}keggdirected}
&\includegraphics[width=0.15\linewidth]{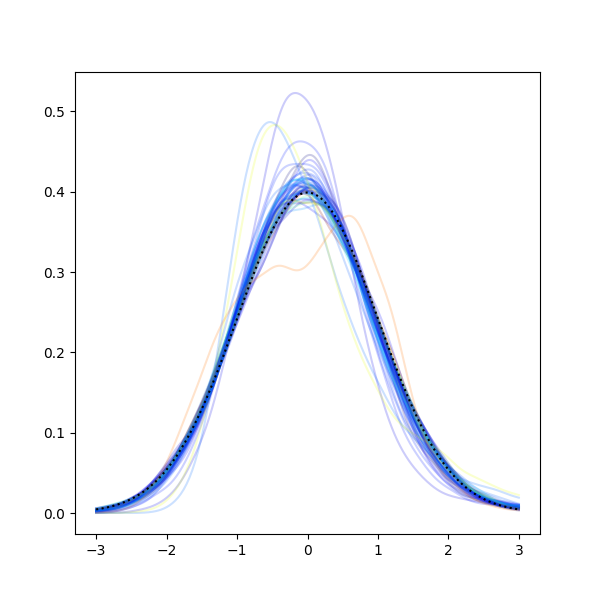}
&\includegraphics[width=0.15\linewidth]{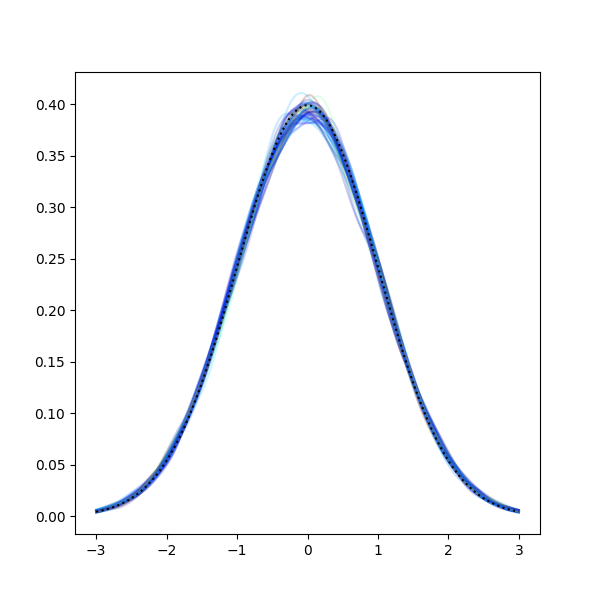}
&\includegraphics[width=0.15\linewidth]{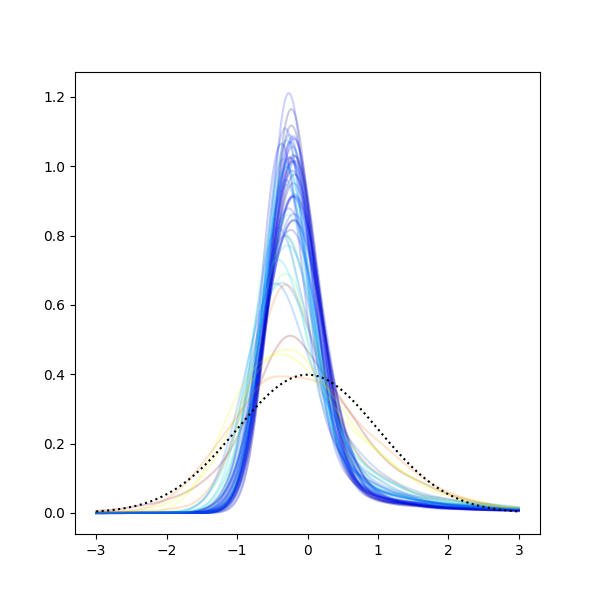}
&\includegraphics[width=0.15\linewidth]{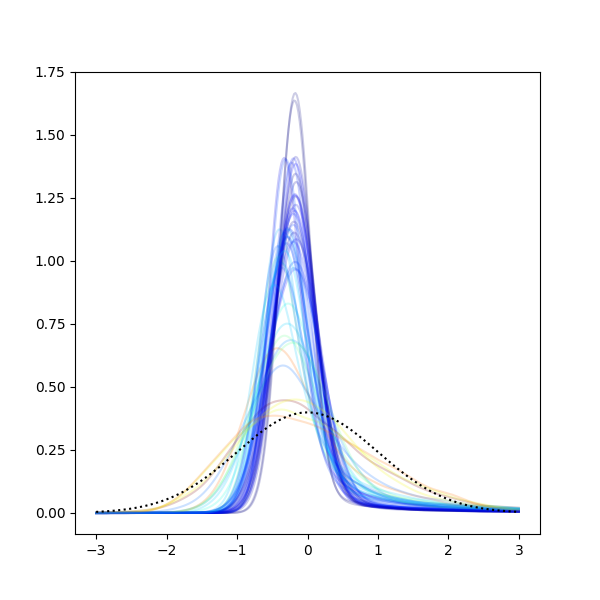}
&\includegraphics[width=0.15\linewidth]{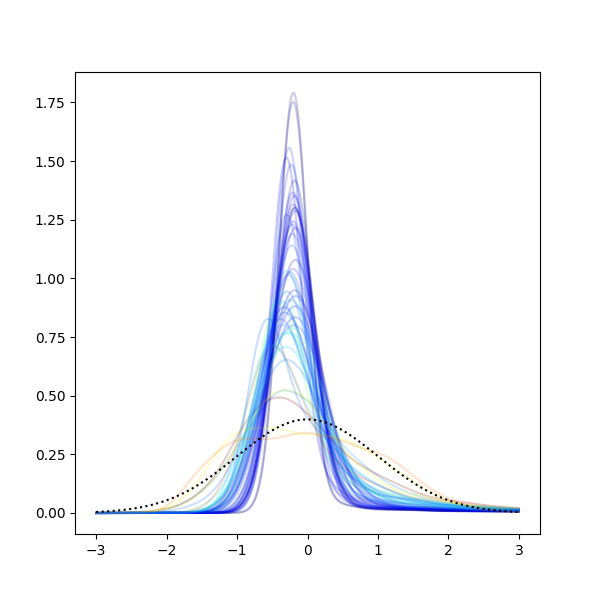}
\end{tabular}
\caption{Posterior marginals, re-scaled to zero mean and unit standard deviation, with the Gaussian density marked as a dotted curve. Colors are according to the principal component of the inputs. These are the four datasets highlighted in the main text for being prominent examples benefiting from latent variables. In these cases the additional depth of model leads to more non-Gaussian marginals.} \label{fig:non-gaussian_marginals}
\end{figure}

\begin{figure}
\begin{tabular}{lccccc}
& CVAE $50,50$ & CVAE $100,100,100$ & \LG (IW) & \LG (IW) & \LGG (IW) \\
\rotatebox{90}{\hspace{0.5cm}keggundirected}
&\includegraphics[width=0.15\linewidth]{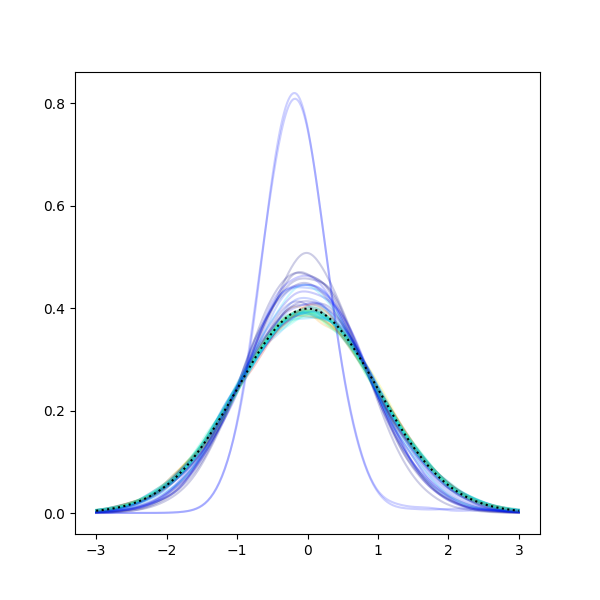}
&\includegraphics[width=0.15\linewidth]{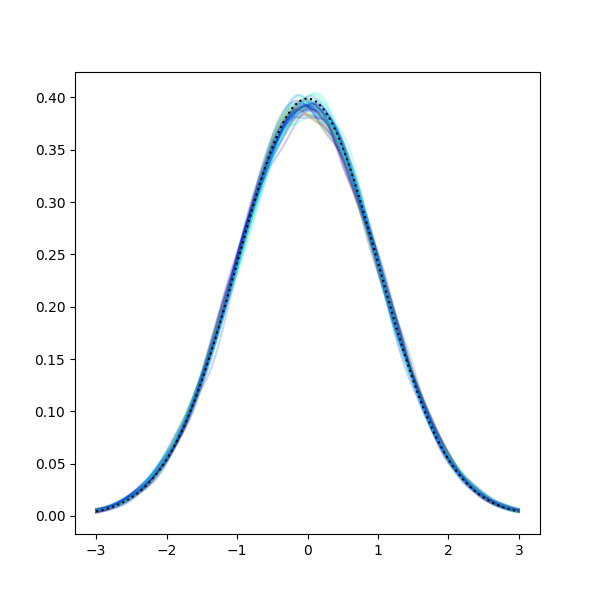}
&\includegraphics[width=0.15\linewidth]{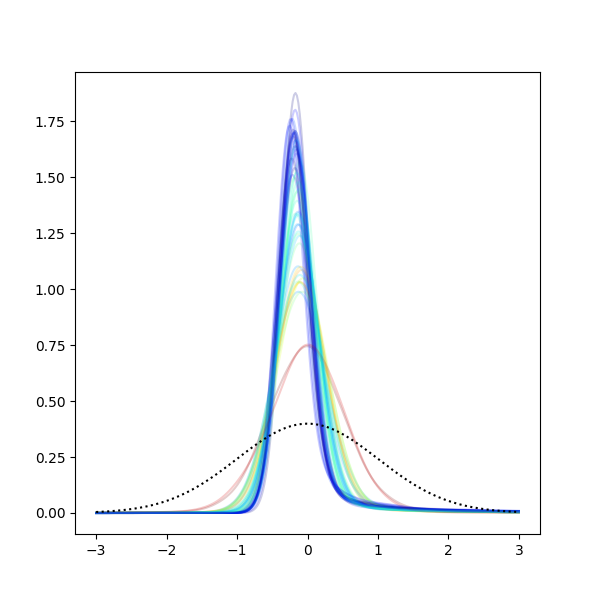}
&\includegraphics[width=0.15\linewidth]{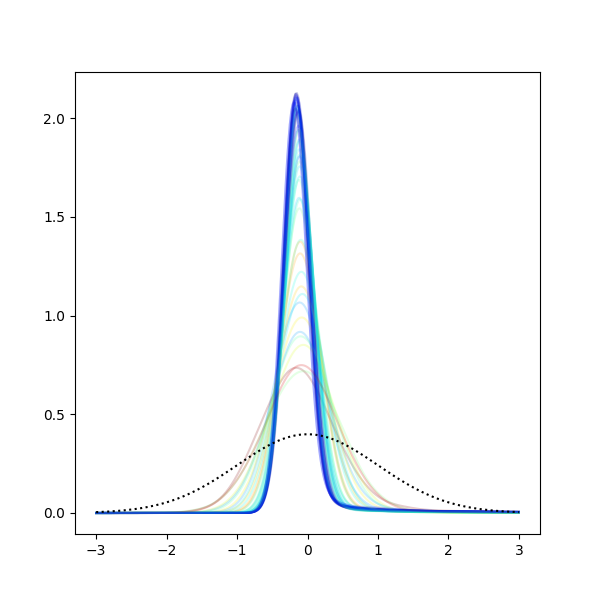}
&\includegraphics[width=0.15\linewidth]{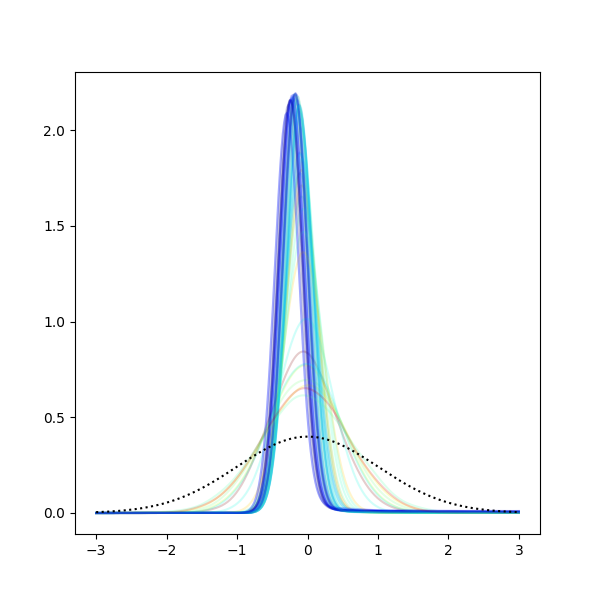}
\\
\rotatebox{90}{\hspace{0.5cm}houseelectric}
&\includegraphics[width=0.15\linewidth]{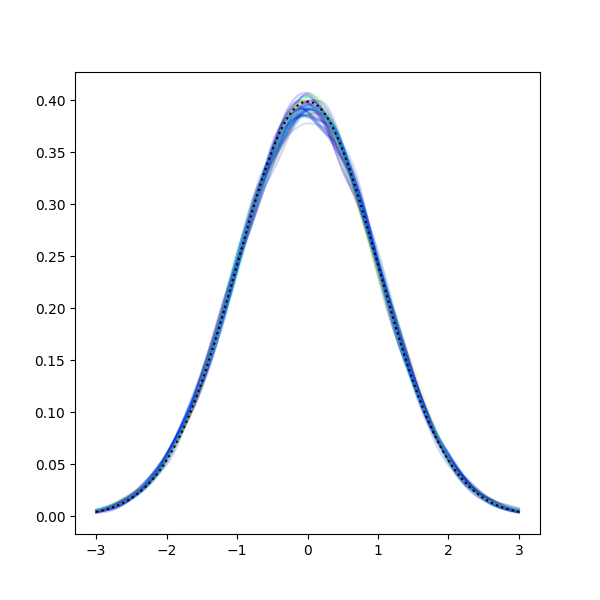}
&\includegraphics[width=0.15\linewidth]{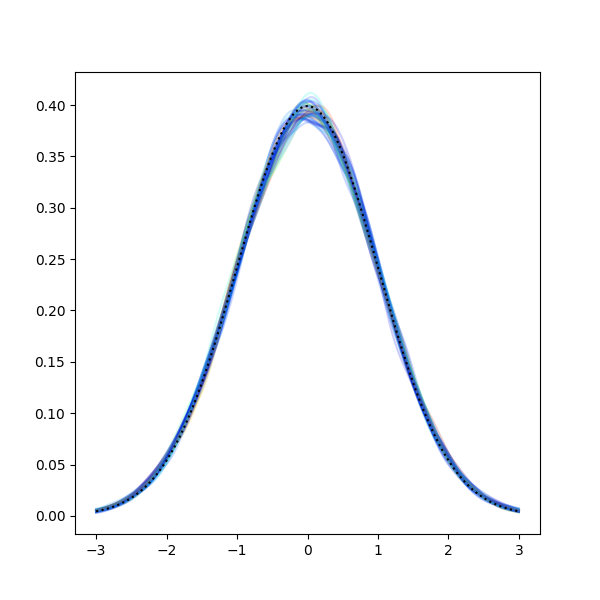}
&\includegraphics[width=0.15\linewidth]{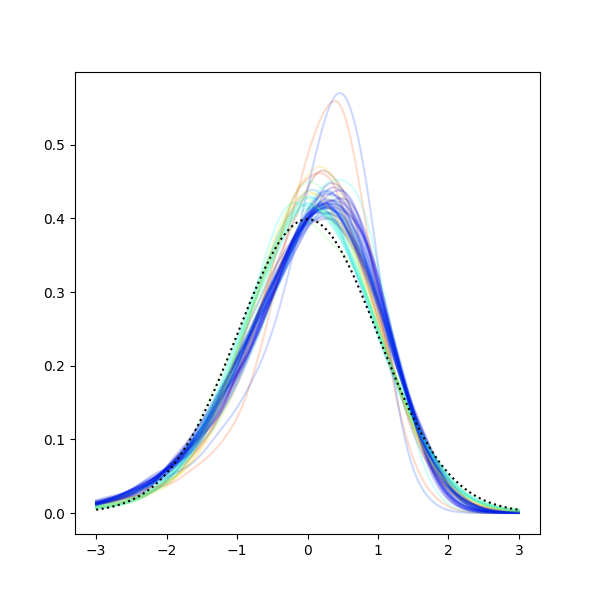}
&\includegraphics[width=0.15\linewidth]{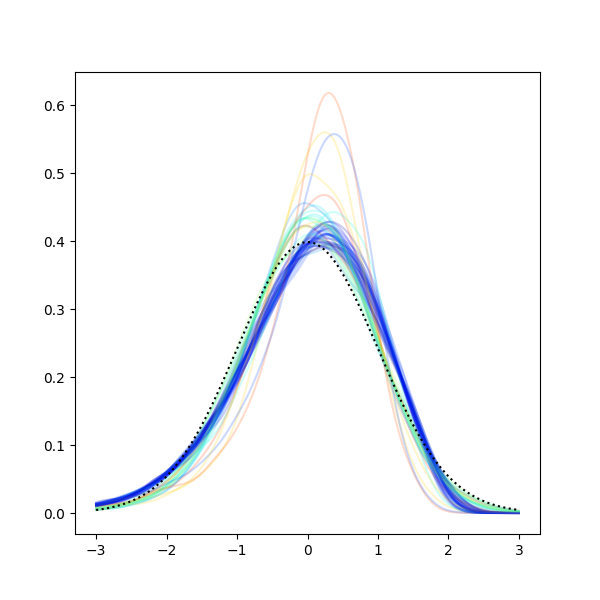}
&\includegraphics[width=0.15\linewidth]{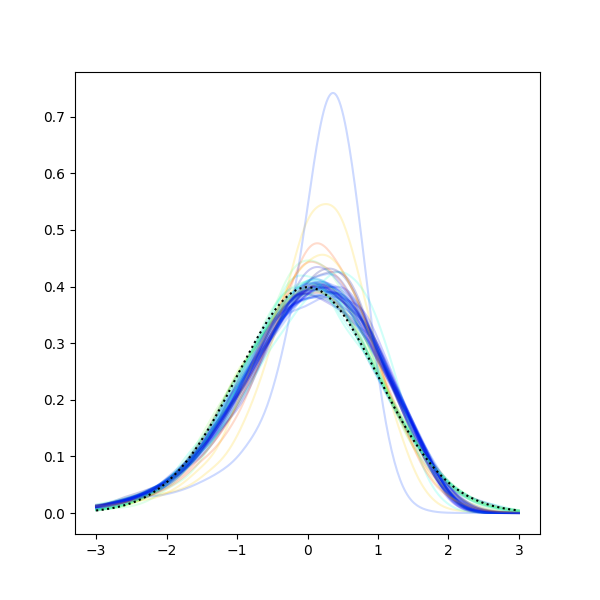}
\\
\rotatebox{90}{\hspace{0.5cm}pol}
&\includegraphics[width=0.15\linewidth]{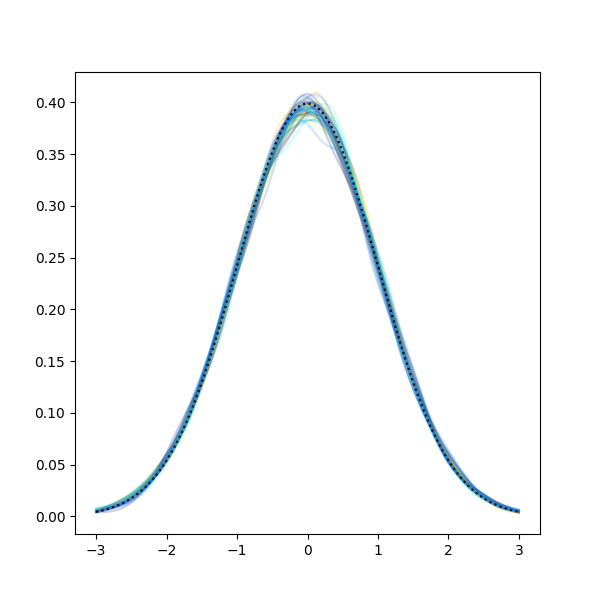}
&\includegraphics[width=0.15\linewidth]{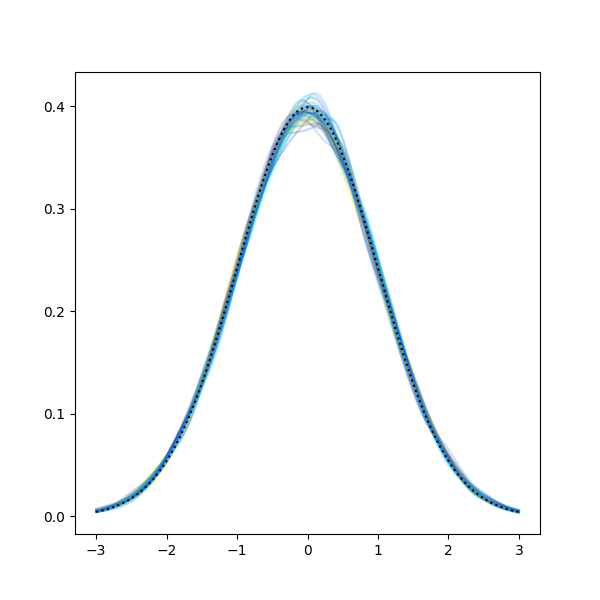}
&\includegraphics[width=0.15\linewidth]{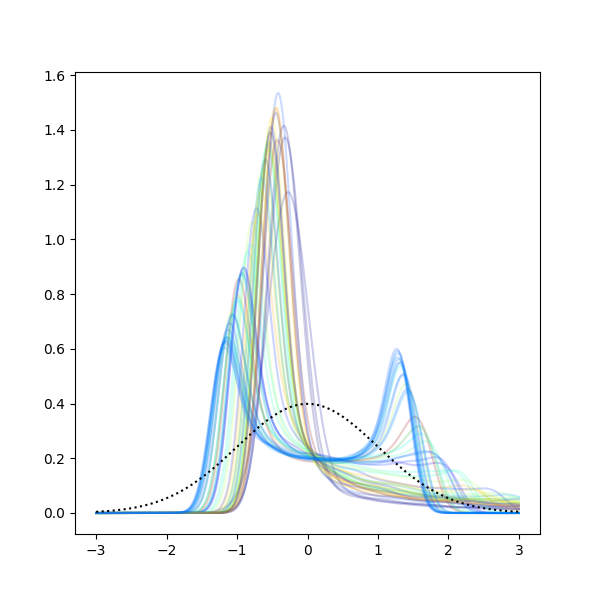}
&\includegraphics[width=0.15\linewidth]{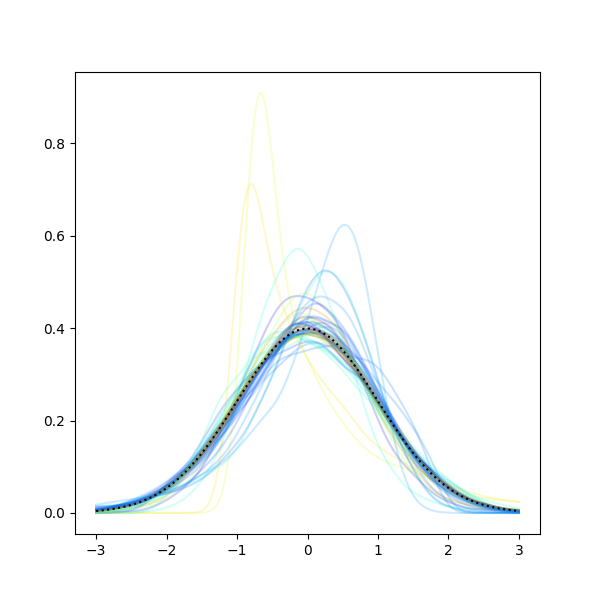}
&\includegraphics[width=0.15\linewidth]{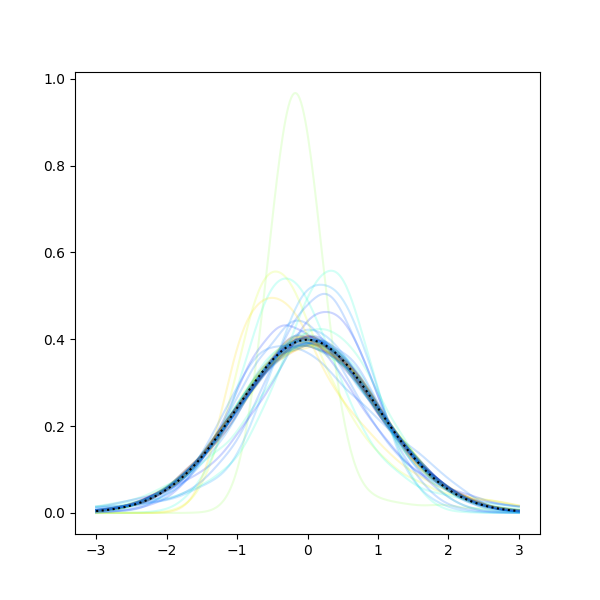}
\\
\rotatebox{90}{\hspace{0.5cm}bike}
&\includegraphics[width=0.15\linewidth]{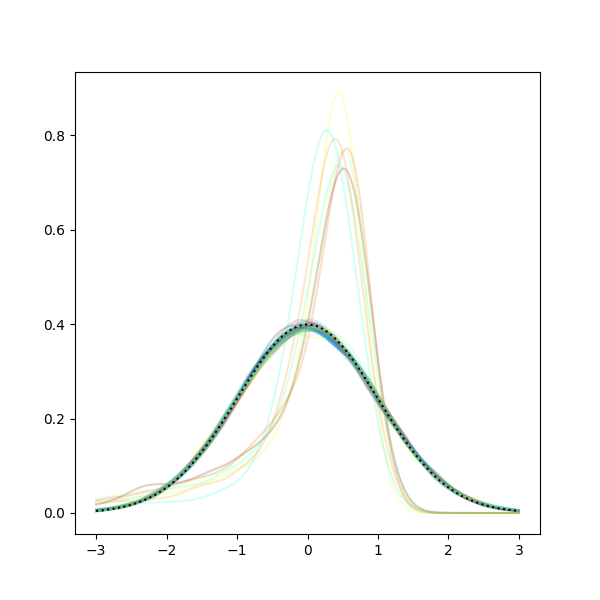}
&\includegraphics[width=0.15\linewidth]{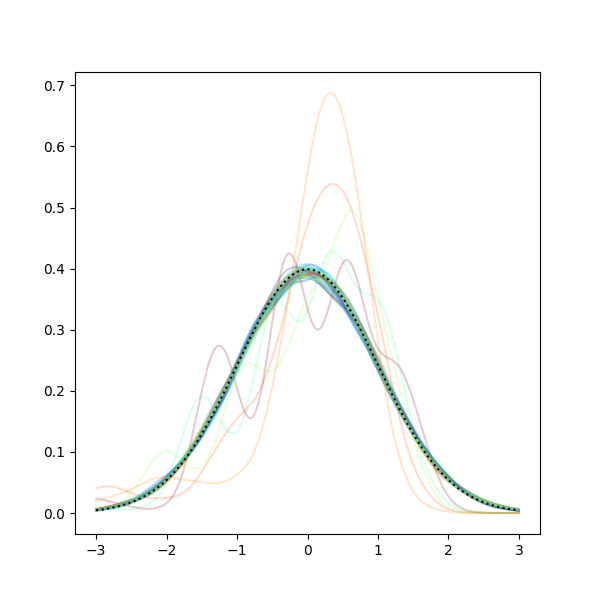}
&\includegraphics[width=0.15\linewidth]{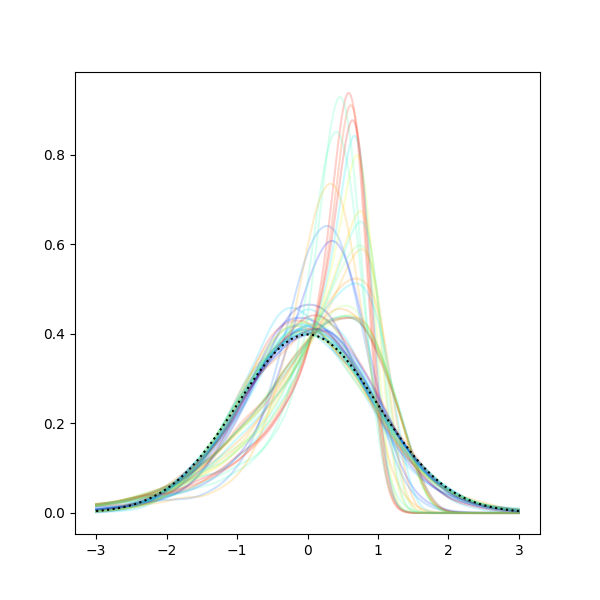}
&\includegraphics[width=0.15\linewidth]{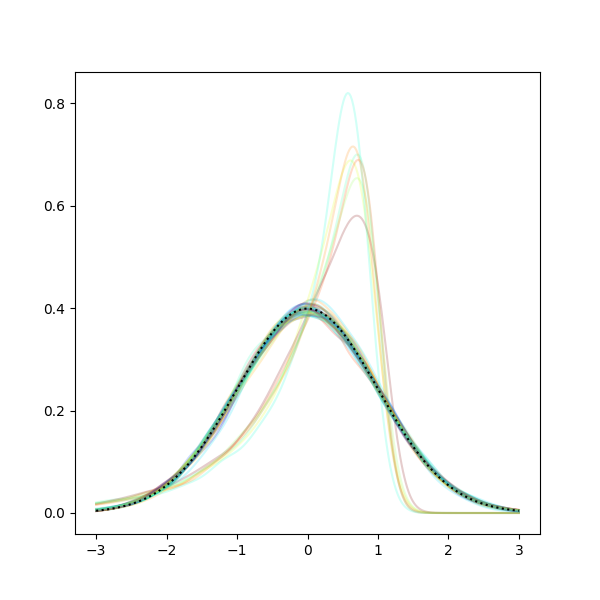}
&\includegraphics[width=0.15\linewidth]{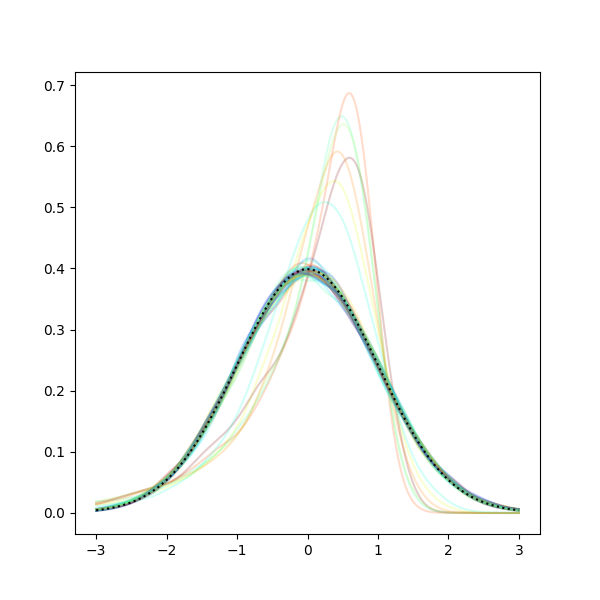}
\end{tabular}
\caption{As Figure \ref{fig:non-gaussian_marginals}, but for the four datasets highlighted in the main text for benefiting from both latent variables and depth. Note that, unlike in Figure \ref{fig:non-gaussian_marginals}, for the `pol' and `bike' data the deeper models actually have \emph{simpler} marginals. This is because the deep model has more complexity in the mapping, so can more closely fit the data, whereas the simple model must explain the data with a complex noise distribution.}
\label{fig:both}
\end{figure}

% % \section{Negative results}
% % We briefly report here the results of some additional experiments that we ran but did not include in the main text. This section is not intended to be thorough, but we hope may provide some additional extra information to future researchers. 

% % \begin{itemize}
% %     \item To calculate posterior likelihoods we found we had to use a large number of samples for some datasets. This is because the likelihood variance in some cases went to its smallest possible value (all positive parameters were clipped at $1e^{-6}$) and so the components of the Gaussian mixture posterior approximation are nearly delta functions. We use $10^4$ samples in all cases, which we found was sufficient. We experimented with using the auxiliary model (known as the `encoder' in the VAE literature) as an importance proposal. To do this, we had to fill the $y$ values for the test points (of course, not using the real ones!), which we did from sampling from $N(0, 1)$. In very preliminary work we found this had high variance so opted instead for the simpler approach of sampling straight from the prior. We expect that some linear combination of the prior and the variational distribution might lead to lower variance results. 
% %     \item We considered using alternatives for random generators to reduce the variance of the Monte Carlo sampling. We tried using low-discrepancy sequences for the $w$ sampling, but noticed extreme artifacts for anything anything greater than 2D so we abandoned this idea. 
% % \end{itemize}[]

\end{document}